\documentclass[letterpaper]{article} 
\usepackage{aaai25}  
\usepackage{times}  
\usepackage{helvet}  
\usepackage{courier}  
\usepackage[hyphens]{url}  
\usepackage{graphicx} 
\usepackage{natbib}  
\usepackage{caption} 
\usepackage{algorithm}
\usepackage{algorithmic}

\usepackage{times}
\usepackage{soul}
\usepackage{url}
\usepackage[utf8]{inputenc}
\usepackage{graphicx}
\usepackage{amsmath}
\usepackage{amsthm}
\usepackage{booktabs}
\usepackage{algorithm}
\usepackage{algorithmic}
\usepackage[switch]{lineno}

\usepackage{natbib}

\usepackage{amsmath}

\usepackage{algorithm}
\usepackage{algorithmic}

\RequirePackage{ascmac}
\RequirePackage{float}
\RequirePackage{perpage}
\MakeSorted{figure}
\MakeSorted{table}

\RequirePackage{comment}

\newtheorem{theorem}{Theorem}
\newtheorem{proposition}{Proposition}
\newtheorem{lemma}{Lemma}

\newtheorem{definition}{Definition}
\newtheorem{assumption}{Assumption}

\RequirePackage{url}
\RequirePackage{natbib}

\RequirePackage{color}
\RequirePackage{tikz}
\tikzset{%
mynode/.style={circle,minimum width=.5ex, fill=none,draw}, 
myfillnode/.style={circle,minimum width=.5ex, fill=lightgray,draw}, 
}
\RequirePackage{amssymb}

\newcommand{\indep}{\perp \!\!\! \perp}
\RequirePackage{amsmath}               
\RequirePackage{lscape}
\RequirePackage{color}
\RequirePackage{tikz}
\usepackage{times}
\usepackage{bm}

\usepackage{stackengine}
\def\defeq{\mathrel{\ensurestackMath{\stackon[1pt]{=}{\scriptscriptstyle\Delta}}}}

\usepackage{soul}

\usepackage{newfloat}
\usepackage{listings}
\DeclareCaptionStyle{ruled}{labelfont=normalfont,labelsep=colon,strut=off} 
\floatstyle{ruled}
\newfloat{listing}{tb}{lst}{}
\floatname{listing}{Listing}
\title{Mediation Analysis for Probabilities of Causation}

\title{Mediation Analysis for Probabilities of Causation}
\author {
    Yuta Kawakami, 
    Jin Tian
}
\affiliations {
    Mohamed bin Zayed University of Artificial Intelligence, UAE\\
    \{Yuta.Kawakami, Jin.Tian\}@mbzuai.ac.ae
}

\usepackage{bibentry}

\begin{document}

\maketitle

\begin{abstract}
\begin{quote}
Probabilities of causation (PoC) offer valuable insights for informed decision-making. This paper introduces novel variants of PoC-controlled direct, natural direct, and natural indirect probability of necessity and sufficiency (PNS). 
These metrics quantify the necessity and sufficiency of a treatment for producing an outcome, accounting for different causal pathways. 
We develop identification theorems for these new PoC measures, allowing for their estimation from observational data. We demonstrate the practical application of our results through an analysis of a real-world psychology dataset.
\end{quote}
\end{abstract}

\section{Introduction}


\citet{Pearl1999} introduced three types of {\it probabilities of causation} (PoC), that is, the probability of necessity and sufficiency (PNS), the probability of necessity (PN), and the probability of sufficiency (PS). 
PoC quantify whether one event was the real cause of another in a given scenario \citep{Robins1989,Tian2000,Pearl09,Kuroki2011,Dawid2014,Murtas2017,Shingaki2021,Kawakami2023b}.  
PoC are valuable for decision-making \citep{Hannart2018,Li2019,Li2022b} and for explaining AI-based decision-making systems \citep{Galhotra2021,Watson2021}. 

Various variants of PoC have been studied, including for multi-valued discrete variables \citep{Li2022,Li2022c} and for continuous and vector variables \citep{Kawakami2024}. 
\citet{Rubinstein2024} introduced direct and indirect mediated PoC to decompose total PoC when there exists a mediator between the treatment and outcome.

Causal mediation analysis is a key method for uncovering the influence of different pathways between the treatment and outcome through mediators \citep{Wright1921,Wright1934,Baron1986,Robins1992,Imai2010a,Imai2010b,TchetgenTchetgen2012}.
Notably, \citet{Pearl2001} formally defined direct and indirect effects for general nonlinear models. 
Causal mediation analysis is also a valuable technique for explainable artificial intelligence (XAI) \citep{Shin2021}. In this paper, we aim to provide causal mediation analysis for PoC, to reveal the necessity and sufficiency of the treatment through different pathways.
Once a treatment is revealed to be necessary and sufficient to induce a particular event via PNS, other causal questions would arise:
\begin{center}
\vspace{0.1cm}
({\bf Q1}). {\it Would the treatment still be necessary and sufficient had the value of the mediator been fixed to a certain value?}\\\vspace{0.1cm}
({\bf Q2}). {\it Would the treatment still be necessary and sufficient had there been no influence via the mediator?}\\\vspace{0.1cm}
({\bf Q3}). {\it Would the treatment still be necessary and sufficient had 
the influence only existed via the mediator?}\vspace{0.1cm}
\end{center}
We introduce new variants of PoC - controlled direct, natural direct, and natural indirect PNS (CD-PNS, ND-PNS, and NI-PNS) 
to answer these questions. 
We further define direct and indirect PoC with evidence to capture more sophisticated counterfactual information useful for decision-making. 
These quantities can retrospectively answer questions (Q1), (Q2), and (Q3) for a specific subpopulation. 
We provide identification results for each type of PoC we introduce.  
Finally, we apply our results to a real-world psychology dataset.

\section{Notations and Background}

We represent a single or vector variable with a capital letter $(X)$ and its realized value with a small letter $(x)$.
Let $\mathbb{I}(\cdot)$ be an indicator function that takes $1$ if the statement in $(\cdot)$ is true and $0$ otherwise.
Denote $\Omega_Y$ be the domain of variable $Y$,
$\mathbb{E}[Y]$ be the expectation of $Y$, 
$\mathbb{P}(Y< y)$ be the cumulative distribution function (CDF) of continuous variable $Y$, and $\mathfrak{p}(Y= y)$ be the probability density function (PDF) of continuous variable $Y$.
We use  $X \indep Y|C$ to denote that $X$ and $Y$ are conditionally independent given $C$.
We use $\preceq$ to denote a total order. A formal definition of total order is given in Appendix A.

\subsubsection{Structural causal models (SCM).}
We use the language of SCMs as our basic 
framework and follow the standard definition in the following \citep{Pearl09}. 
An SCM ${\cal M}$ is a tuple $\left<{\boldsymbol V},{\boldsymbol U}, {\cal F}, \mathbb{P}_{\boldsymbol U} \right>$, where ${\boldsymbol U}$ is a set of exogenous (unobserved) variables following a distribution $\mathbb{P}_{\boldsymbol U}$, and ${\boldsymbol V}$ is a set of endogenous (observable) variables whose values are determined by structural functions ${\cal F}=\{f_{V_i}\}_{V_i \in {\boldsymbol V}}$ such that $v_i:= f_{V_i}({\mathbf{pa}}_{V_i},{\boldsymbol u}_{V_i})$ where ${\mathbf{PA}}_{V_i} \subseteq {\boldsymbol V}$ and $\boldsymbol{U}_{V_i} \subseteq {\boldsymbol U}$. 
Each SCM ${\cal M}$ induces an observational distribution $\mathbb{P}_{\boldsymbol V}$ over ${\boldsymbol V}$, and a causal graph $G({\cal M})$ 
in which there exists a directed edge from every variable in ${\mathbf{PA}}_{V_i}$ and $\boldsymbol{U}_{V_i}$ to $V_i$. 
An intervention of setting a set of endogenous variables ${\boldsymbol X}$ to constants ${\boldsymbol x}$, denoted by $do({\boldsymbol x})$, replaces the original equations of ${\boldsymbol X}$
 by the constants ${\boldsymbol x}$ and induces a \textit{sub-model}  ${\cal M}_{{\boldsymbol x}}$.
We denote the potential outcome $Y$ under intervention $do({\boldsymbol x})$ by $Y_{{\boldsymbol x}}({\boldsymbol u})$, which is the solution of $Y$ in the sub-model ${\cal M}_{{\boldsymbol x}}$ given ${\boldsymbol U}={\boldsymbol u}$.

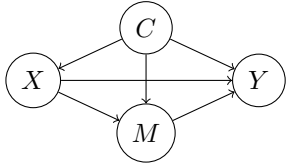
\begin{figure}[tb]
    \centering
    \scalebox{1}{
\begin{tikzpicture}
    \node[mynode] (x) at (0,0) {$X$};
    \node[mynode] (y) at (3,0) {$Y$};
    \node[mynode] (u) at (1.5,0.7) {$C$};
    \node[mynode] (m) at (1.5,-0.7) {$M$};

    \path (x) edge[->] (y);
    \path (u) edge[->] (y);
    \path (u) edge[->]  (x);

\path (x) edge[->] (m);
\path (m) edge[->] (y);
\path (u) edge[->] (m);
\end{tikzpicture}
}
\vspace{0cm}
    \caption{A causal graph representing SCM ${\cal M}$.}
    \label{DAG1}
    \end{figure}

\subsubsection{Probabilities of causation (PoC).}

\citet{Kawakami2024} defined the (multivariate  conditional) PoC for vectors of continuous or discrete variables as follows:
\begin{definition}[PoC] \citep{Kawakami2024}
\label{def41}
The (multivariate  conditional) PoC are defined by 
\begin{equation}
\text{\normalfont PNS}(y;x',x,c)\defeq\mathbb{P}(Y_{x'} \prec y \preceq Y_{x}|C=c),
\end{equation}
\begin{equation}
\text{\normalfont PN}(y;x',x,c)\defeq\mathbb{P}(Y_{x'} \prec y |y \preceq Y,X=x,C=c),
\end{equation}
\begin{equation}
\text{\normalfont PS}(y;x',x,c)\defeq\mathbb{P}(y \preceq Y_{x} |Y \prec y,X=x',C=c).
\end{equation}
\end{definition}
\noindent $\text{\normalfont PNS}(y;x',x,c)$ provides a measure of the necessity and sufficiency   of $x$ w.r.t. $x'$ to produce $Y\succeq y$ given $C=c$.
$\text{\normalfont PN}(y;x',x,c)$ and $\text{\normalfont PS}(y;x',x,c)$ provide a measure of the necessity and sufficiency, respectively, of $x$ w.r.t. $x'$ to produce $Y\succeq y$ given $C=c$.

We will often call $\text{\normalfont PNS}$ \textit{total PNS (T-PNS)} and denote it by $\text{\normalfont T-PNS}(y;x',x,c)$  for convenience. 
When treatment $X$ and outcome $Y$ are binary, 
PNS, PS, and PS become (setting $y=1$) 
$\text{\normalfont PNS}(c)=\mathbb{P}(Y_{0}=0,Y_{1}=1|C=c)$,
$\text{\normalfont PN}(c)=\mathbb{P}(Y_{0}=0|Y=1,X=1,C=c)$, and 
$\text{\normalfont PS}(c)=\mathbb{P}(Y_{1}=1|Y=0,X=0,C=c)$
for any $c \in \Omega_C$, which reduce to Pearl's (1999) original definition when $C=\emptyset$.

\subsubsection{Causal mediation analysis.}
Causal mediation analysis reveals the strength of different pathways between treatment and outcome through a mediator.
Researchers often consider the following SCM ${\cal M}$:
\begin{equation}
\vspace{0.1cm}
\begin{gathered}
Y:=f_Y(X,M,C,U_Y), M:=f_M(X,C,U_M), \\
X:=f_X(C,U_X), C:=f_C(U_C),
\end{gathered}
\end{equation}
where all variables can be vectors, 
and $U_X$, $U_C$, $U_Y$, and $U_M$ are latent exogenous variables.
{Assume that the domains $\Omega_Y$ and $\Omega_{U_Y} \times \Omega_{U_M}$ are totally ordered sets with $\preceq$.} 
Figure \ref{DAG1} shows the causal graph of SCM ${\cal M}$ (with latent variables dropped).

One widely used model in the mediation analysis is a linear SCM ${\cal M}^L$ 
\citep{Baron1986} consisting of 
$Y:=\beta_0+\beta_1 X+\beta_2 M+\beta_3 C+U_Y$ and $M:=\alpha_0+\alpha_1 X+\alpha_2 C+U_M$, where $U_Y\sim {\cal N}(\mu_Y,\sigma^2_Y)$ and $ U_M \sim {\cal N}(\mu_M,\sigma^2_M)$ are independent normal distribution.
Under SCM ${\cal M}^L$, the total effect of $X$ on $Y$ is $\beta_1+\alpha_1\beta_2$, the indirect effect is $\alpha_1\beta_2$, and the direct effect is $\beta_1$.

\citet{Pearl2001} defined the total, controlled direct, natural direct, and natural indirect effects for general (nonlinear and nonparametric) SCM ${\cal M}$. 
\begin{definition}[TE, CDE, NDE, and NIE] \citep{Pearl2001}
\label{def1}
The total, controlled direct, natural direct, and natural indirect effects are defined by:
\begin{enumerate}
\setlength{\itemsep}{1pt}
\setlength{\parskip}{1pt}
    \item Total Effect (TE): $\text{\normalfont TE}(y;x',x)\defeq\mathbb{E}[Y_{x}]-\mathbb{E}[Y_{x'}]$
    \item Controlled Direct Effect (CDE): $\text{\normalfont CDE}(y;x',x,m)\defeq\mathbb{E}[Y_{x,{m}}]-\mathbb{E}[Y_{x',{m}}]$
    \item Natural Direct Effect (NDE): $\text{\normalfont NDE}(y;x',x)\defeq\mathbb{E}[Y_{x,{M_{x'}}}]-\mathbb{E}[Y_{x'}]$
    \item Natural Indirect Effect (NIE): $\text{\normalfont NIE}(y;x',x)\defeq\mathbb{E}[Y_{x',{M_{x}}}]-\mathbb{E}[Y_{x'}]$
\end{enumerate}
\end{definition}
\noindent 
CDE represents the causal effect of changing the treatment from $x'$ to $x$ had the value of the mediator been fixed at a certain value.
NDE represents the causal effect of changing the treatment from $x'$ to $x$  had the value of the mediator been kept to the same value $M_{x'}$ that $M$ attains under $x'$. 
NIE represents the causal effect of changing the mediator from $M_{x'}$ to $M_{x}$ had the value of the treatment been fixed to $x'$.
TE can be decomposed into NDE and NIE by $\text{\normalfont TE}(y;x',x)=\text{\normalfont NDE}(y;x',x)-\text{\normalfont NIE}(y;x,x')=\text{\normalfont NIE}(y;x',x)-\text{\normalfont NDE}(y;x,x')$. 

These direct and indirect effects may be identified from observational distributions under various settings \citep{Pearl2001,Avin2005,Shpitser2008,Shpitser2013,Malinsky2019}. 
A widely used assumption for identifying causal mediation effects is  the following sequential ignorability assumption \citep{Imai2010a}:
\begin{assumption}[Sequential ignorability]
\label{SCAS2}
The following two conditional independence statements hold:
\begin{center}
\vspace{0.1cm}
{\normalfont (1)} $\{Y_{x,m},M_x\} \indep X|C=c$\  and\  
{\normalfont (2)} $M_x \indep Y_{x,m}|C=c$\\
\vspace{0.1cm}
\end{center}
for any $m \in \Omega_M$ and $x \in \Omega_X$, where $\mathfrak{p}(X=x|C=c)>0$ and $\mathfrak{p}(M=m|X=x,C=c)>0$ for any $m \in \Omega_M$, $x \in \Omega_X$, and $c \in \Omega_C$.
\end{assumption}
\noindent 
We have
\begin{proposition}[Identification of $\mathbb{P}(Y_{x',M_{x}}\prec y|C=c)$] \citep{Imai2010a,VanderWeele2014}
\label{prop2}
Under SCM ${\cal M}$  and Assumption \ref{SCAS2}, the  counterfactual  $\mathbb{P}(Y_{x',M_{x}}\prec y|C=c)$ is identifiable by
\begin{equation}
\begin{aligned}
&\mathbb{P}(Y_{x',M_{x}}\prec y|C=c)\\
&=\int_{\Omega_{M}}\mathbb{P}(Y\prec y|X=x',M=m,C=c)\\
&\hspace{2.5cm}\times\mathfrak{p}(M=m|X=x,C=c) dm
\end{aligned}
\end{equation}
for any $x', x \in \Omega_X$, $y \in \Omega_Y$, and $c \in \Omega_C$.
\end{proposition}

\section{Direct and Indirect PNS}

In this section, 
we introduce new concepts of direct and indirect PNS and provide corresponding identification results.
{We will focus our attention on PNS, and show in the next section that direct and indirect PN and PS can be derived as special cases of direct and indirect PNS with evidence.}

\subsection{Definition of CD-PNS, ND-PNS, and NI-PNS}
We define controlled direct, natural direct, and natural indirect probabilities of necessity and sufficiency.
\begin{definition}[CD-PNS, ND-PNS, and NI-PNS]
\label{def3}
The controlled direct, natural direct, and natural indirect 
PNS (CD-PNS, ND-PNS, and NI-PNS) are defined by
\begin{equation}
\begin{aligned}
&\text{\normalfont CD-PNS}(y;x',x,m,c)\defeq\mathbb{P}(Y_{x',m} \prec y \preceq Y_{x,m}|C=c),
\end{aligned}
\end{equation}
\begin{equation}
\begin{aligned}
&\text{\normalfont ND-PNS}(y;x',x,c)\defeq\mathbb{P}(Y_{x'} \prec y \preceq Y_{x}, Y_{x',M_{x}} \prec y|C=c),
\end{aligned}
\end{equation}
\begin{equation}
\begin{aligned}
&\text{\normalfont NI-PNS}(y;x',x,c)\defeq\mathbb{P}(Y_{x'} \prec y \preceq Y_{x},y \preceq Y_{x',M_{x}}|C=c).
\end{aligned}
\end{equation}
\end{definition}

First, the controlled direct PNS (CD-PNS) 
{provides a measure of the necessity and sufficiency   of $x$ w.r.t. $x'$ to produce $Y\succeq y$ given $C=c$ when the mediator is fixed to a value $M=m$.} 
CD-PNS can be used to answer the causal question (Q1).
CD-PNS consists of two counterfactual conditions:
\begin{center}
\vspace{0.1cm}
({\bf A1}). {\it ``had the treatment and the mediator been $(x',m)$, the outcome would be $Y\prec y$"} $(Y_{x',m} \prec y)$;  and\\\vspace{0.1cm}
({\bf A2}). {\it ``had the treatment and the mediator been $(x,m)$, the outcome would be $y \preceq Y$"} $(y \preceq Y_{x,m})$.\vspace{0.1cm}
\end{center}
Conditions (A1) and (A2) have different values of treatment and the same values of mediator.
The relative values of the potential outcomes $Y_{x,m}$  are shown in Figure \ref{fig2} (b). 
For comparison, Figure \ref{fig2} (a) shows the situation for T-PNS.

Second, ND-PNS has three counterfactual conditions: 
\begin{center}
\vspace{0.1cm}
({\bf B1}). {\it ``had the treatment been $x'$, the outcome would be $Y\prec y$"} $(Y_{x'}=Y_{x',M_{x'}} \prec y)$,\\\vspace{0.1cm}
({\bf B2}). {\it ``had the treatment been $x'$ but the mediator was kept at the same value $M_{x}$ when the treatment is $x$, the outcome would be {$Y \prec y$"}} $(Y_{x',M_{x}} \prec y)$, and \\\vspace{0.1cm}
({\bf B3}). {\it ``had the treatment been $x$, the outcome would be $y \preceq Y$"} $(y \preceq Y_{x}=Y_{x,M_{x}})$,\vspace{0.1cm}
\end{center} 
The relative values of the potential outcomes are shown in Figure \ref{fig2} (c).
Conditions (B1) and (B3) mean $Y_{x'} \prec y \preceq Y_{x}$, which is the same condition in T-PNS and represents that the treatment $x$ is necessary and sufficient w.r.t. $x'$ to provoke the event $y \preceq Y$  given $C=c$. 
Conditions (B2) and (B3) mean $ Y_{x',M_{x}} \prec y \preceq Y_{x,M_{x}}$, which represents the necessity and sufficiency of $x$ w.r.t. $x'$ to produce $Y\succeq y$ given $C=c$ when keeping the values of the mediator by the same as $M_{x}$.
In other words, they mean that the treatment would be necessary and sufficient even if there were no influences via the mediator.
Therefore, ND-PNS can answer the causal question (Q2).

Third, NI-PNS has three counterfactual conditions:
\begin{center}
\vspace{0.1cm}
({\bf C1}). {\it ``had the treatment been $x'$, the outcome would be $Y\prec y$"} $(Y_{x'}=Y_{x',M_{x'}} \prec y)$,\\\vspace{0.1cm}
({\bf C2}). {\it ``had the treatment been $x'$ but the mediator was kept at the same value $M_{x}$ when the treatment is $x$, the outcome would be $y \preceq  Y$"} $(y \preceq Y_{x',M_{x}})$, and \\\vspace{0.1cm}
({\bf C3}). {\it ``had the treatment been $x$, the outcome would be $y \preceq Y$"} $(y \preceq  Y_{x}=Y_{x,M_{x}})$,\vspace{0.1cm}
\end{center} 
The relative values of the potential outcomes are shown in Figure \ref{fig2} (d).
Conditions (C1) and (C3) mean $Y_{x'} \prec y \preceq Y_{x}$, which is the same condition in T-PNS and states that the treatment $x$ is necessary and sufficient w.r.t. $x'$ to provoke the event $y \preceq Y$  given $C=c$. 
Conditions (C1) and (C2) mean $ Y_{x',M_{x'}} \prec y \preceq Y_{x',M_{x}}$, which represents the necessity and sufficiency of $M_{x}$ w.r.t. $M_{x'}$ to produce $Y\succeq y$ given $C=c$ when setting the treatment to $x'$. 
In other words, they mean that the treatment would be necessary and sufficient if 
the influence is only via the mediator.
Therefore, NI-PNS can answer the causal question (Q3).


\begin{figure}[tb]
    \centering
    \scalebox{1}{
\begin{tikzpicture}
    \path (0.5,0) edge[->] (7,0); 
    \coordinate[label = above:$Y_{x'}$] (A) at (1.5,0);
    \node at (A)[circle,fill,inner sep=1pt]{};
    \coordinate[label = above:$y$] (B) at (4,0); 
    \node at (B)[circle,fill,inner sep=1pt]{};
    \coordinate[label = above:$Y_{x}$] (D) at (6,0);
    \node at (D)[circle,fill,inner sep=1pt]{};
\end{tikzpicture}
}
\caption*{(a) Order of potential outcomes in T-PNS.}
    \scalebox{1}{
\begin{tikzpicture}
    \path (0.5,0) edge[->] (7,0); 
    \coordinate[label = above:$Y_{x',m}$] (A) at (1.5,0);
    \node at (A)[circle,fill,inner sep=1pt]{};
    \coordinate[label = above:$y$] (B) at (4,0); 
    \node at (B)[circle,fill,inner sep=1pt]{};
    \coordinate[label = above:$Y_{x,m}$] (D) at (6,0);
    \node at (D)[circle,fill,inner sep=1pt]{};
\end{tikzpicture}
}
\caption*{(b) Order of potential outcomes in CD-PNS.}
    \scalebox{1}{
\begin{tikzpicture}
    \path (0.5,0) edge[->] (7,0); 
    \coordinate[label = above:$Y_{x'}$] (A) at (1.5,0);
    \node at (A)[circle,fill,inner sep=1pt]{};
    \coordinate[label = above:$Y_{x',M_{x}}$] (B) at (3,0); 
    \node at (B)[circle,fill,inner sep=1pt]{};
    \coordinate[label = above:$y$] (C) at (4.5,0);
    \node at (C)[circle,fill,inner sep=1pt]{};
    \coordinate[label = above:$Y_{x}$] (D) at (6,0);
    \node at (D)[circle,fill,inner sep=1pt]{};
\end{tikzpicture}
}
\caption*{(c) Order of potential outcomes in ND-PNS.}
    \scalebox{1}{
\begin{tikzpicture}
    \path (0.5,0) edge[->] (7,0);
    \coordinate[label = above:$Y_{x'}$] (A) at (1.5,0);
    \node at (A)[circle,fill,inner sep=1pt]{};
    \coordinate[label = above:$y$] (B) at (3,0);  
    \node at (B)[circle,fill,inner sep=1pt]{};
    \coordinate[label = above:$Y_{x',M_{x}}$] (C) at (4.5,0);
    \node at (C)[circle,fill,inner sep=1pt]{};
    \coordinate[label = above:$Y_{x}$] (D) at (6,0);
    \node at (D)[circle,fill,inner sep=1pt]{};
\end{tikzpicture}
}
\caption*{(d) Order of potential outcomes in NI-PNS.}
\vspace{0cm}
\caption{Order of potential outcomes in each PNS.}
\label{fig2}
\end{figure}

Then, the following proposition holds.
\begin{proposition}
\label{prop}
We have
\begin{equation}\label{eq:pns-decom}
\begin{aligned}
&\text{\normalfont T-PNS}(y;x',x,c)\\
&=\text{\normalfont ND-PNS}(y;x',x,c)+\text{\normalfont NI-PNS}(y;x',x,c).
\end{aligned}
\end{equation}
\end{proposition}
\noindent 
Eq.~(\ref{eq:pns-decom}) states that the total PNS can be decomposed into a summation of the natural direct and natural indirect PNS, a desired property of causal mediation analysis.

\subsubsection{Remark 1.}
Researchers have considered the proportion of direct or indirect influence in the total influence, which captures how important each pathway is in explaining the total influence \citep{VanderWeele2013}. 
However, the proportions of direct and indirect effects in the total effects under linear SCM ${\cal M}^L$ or the proportions of NDE and NIE in TE may not always make sense since these quantities may take negative values. 
In contrast, the proportions of ND-PNS and NI-PNS in T-PNS are given by 
${\text{\normalfont ND-PNS}(y;x',x,c)}/{\text{\normalfont T-PNS}(y;x',x,c)}=\mathbb{P}(Y_{x',M_{x}} \prec y|Y_{x'} \prec y \preceq Y_{x},C=c)$ and
${\text{\normalfont NI-PNS}(y;x',x,c)}/{\text{\normalfont T-PNS}(y;x',x,c)}=\mathbb{P}(y \preceq Y_{x',M_{x}}|Y_{x'} \prec y \preceq Y_{x},C=c)$,
respectively, which do not take negative values. 
Additionally, the sum of the proportions of ND-PNS and NI-PNS is always equal to $1$.

\subsubsection{Remark 2.}
\citet{Rubinstein2024} defined, for binary treatment, outcome, and mediator, the total mediated PoC by
$\delta(c)\defeq\mathbb{P}(Y_0=0|Y_1=1,M_1=1,C=c)$,
the direct mediated PoC by
$\psi(c)\defeq\mathbb{P}(Y_{1,M_0}=0,Y_{0,M_0}=0|Y_{1,M_1}=1,M_1=1,C=c)$,
and the indirect mediated PoC by
$\zeta(c)\defeq\mathbb{P}(Y_{1,M_0}=1,Y_{0,M_0}=0|Y_{1,M_1}=1,M_1=1,C=c)$. 
While we focus on the necessity and sufficiency of the treatment to provoke an event, their definitions of mediated PoC differ from ours and are aimed at answering different questions. 
For example, their total 
mediated PoC is motivated by the question: 
``Given that subjects would experience events $Y=1$ and $M=1$ had they taken a treatment $X=1$, 
what is the probability that they would not have experienced the event $Y=1$ in the absence of the treatment?". 
We note that their mediated PoC satisfy the property $\delta(c)=\psi(c)+\zeta(c)$.
We provide a detailed comparison in Appendix E.


\subsection{Identification of CD-PNS, ND-PNS, and NI-PNS}

Next, we provide identification theorems for the direct and indirect PNSs we have introduced. 

\subsubsection{Assumptions}
{The identification of PoC relies on monotonicity assumptions in the literature \citep{Tian2000}. 
We will make similar assumptions, specifically similar to those in \citep{Kawakami2024}.} 
\begin{assumption}
\label{SUP2}
Potential outcome $Y_{x,m}$ has conditional PDF $\mathfrak{p}_{Y_{x,m}|C=c}$ for each $x \in \Omega_{X}$, $m \in \Omega_M$, and $c \in \Omega_C$, and its support $\{y \in \Omega_Y: \mathfrak{p}_{Y_{x,m}|C=c}(y) \ne0 \}$ is
the same
for each $x \in \Omega_{X}$, $m \in \Omega_M$, and $c \in \Omega_C$.
\end{assumption}
\begin{assumption}
\label{SUP1}
Potential outcome $Y_{x',M_{x}}$ has conditional PDF $\mathfrak{p}_{Y_{x',M_{x}}|C=c}$ for each $x', x \in \Omega_{X}$ and $c \in \Omega_C$, and its support $\{y \in \Omega_Y: \mathfrak{p}_{Y_{x',M_{x}}|C=c}(y) \ne0 \}$ is
the same
for each $x', x \in \Omega_{X}$ and $c \in \Omega_C$.
\end{assumption}
\noindent Assumptions~\ref{SUP2} and \ref{SUP1} are reasonable for continuous variables. For example, 
potential outcomes $Y_{x,m}$, $Y_{x',M_{x}}$ often has $[-\infty,\infty]$ support, 
such as in linear SCM ${\cal M}^L$.

We assume the following monotonicity condition 
for identifying CD-PNS:
\begin{assumption}[Monotonicity over $f_Y$]
\label{AS2}
The function $f_Y(x,m,c,U_Y)$ is either monotonic increasing on $U_Y$ for all $x \in \Omega_X$, $m \in \Omega_M$, and $c \in \Omega_C$, 
or monotonic decreasing on $U_Y$ for all $x \in \Omega_X$, $m \in \Omega_M$, and $c \in \Omega_C$, almost surely w.r.t. $\mathbb{P}_{U_Y}$.
\end{assumption}
\noindent Alternatively, one may assume  monotonicity over potential outcomes:

\noindent{\bf Assumption 4'}
(Conditional monotonicity over $Y_{x,m}$)
{\it 
The potential outcomes $Y_{x,m}$ satisfy:  for any $x', x \in \Omega_X$, $m \in \Omega_M$, $y \in \Omega_Y$, and $c \in \Omega_C$, 
either $\mathbb{P}(Y_{x',m}\prec y \preceq Y_{x,m}|C=c)=0$ or $\mathbb{P}(Y_{x,m}\prec y \preceq Y_{x',m}|C=c)=0$.
}
\vspace{0.1in}

\noindent 
Assumptions~\ref{AS2} and \ref{AS2}' are equivalent under Assumption~\ref{SUP2} (a straightforward extension of Theorem 4.1 in \citep{Kawakami2024}). 
We note that the widely used linear SCM ${\cal M}^L$ satisfies  Assumption \ref{AS2}. 
Furthermore, another popular model, a nonlinear SCM with normal distribution ${\cal M}^N$, consisting of $Y:=f_Y(X,M,C)+U_Y$ and 
$M:=f_M(X,C)+U_M$, where $U_Y\sim {\cal N}(\mu_Y,\sigma^2_Y)$ and $ U_M \sim {\cal N}(\mu_M,\sigma^2_M)$, 
also satisfies Assumptions \ref{SUP2}-\ref{AS2}. 

Let the compound function $f_Y\circ f_M$ represent $(f_Y\circ f_M)(x',x,c,\tilde{U})=f_Y(x',f_M(x,c,U_M),c,U_Y)$ for all $x', x \in \Omega_X$ and $c \in \Omega_C$, where $\tilde{U}=(U_Y,U_M)$. 
We assume the following 
for identifying ND-PNS and NI-PNS: 
\begin{assumption}[Monotonicity over $f_Y \circ f_M$]
\label{AS1}
The function $(f_Y\circ f_M)(x',x,c,\tilde{U})$ is either monotonic increasing on $\tilde{U}$ for all $x', x \in \Omega_X$ and $c \in \Omega_C$, or monotonic decreasing on $\tilde{U}$ for all $x', x \in \Omega_X$ and $c \in \Omega_C$, almost surely w.r.t. $\mathbb{P}_{\tilde{U}}$.
\end{assumption}
\noindent Or, alternatively, 

\noindent{\bf Assumption 5'}
(Conditional monotonicity over $Y_{x',M_{x}}$)
{\it 
The potential outcomes $Y_{x',M_{x}}$ satisfy:  for any $x,x',x'',x''' \in \Omega_X$, $y \in \Omega_Y$, and $c \in \Omega_C$, 
either $\mathbb{P}(Y_{x',M_{x}}\prec y \preceq Y_{x''',M_{x''}}|C=c)=0$ or $\mathbb{P}(Y_{x''',M_{x''}} \prec y \preceq Y_{x',M_{x}}|C=c)=0$.
}
\vspace{0.1in}

\noindent 
Similarly, Assumptions~\ref{AS1} and \ref{AS1}' are equivalent under Assumption~\ref{SUP1}. 
We note that both the linear SCM ${\cal M}^L$ and the nonlinear SCM with normal distribution ${\cal M}^N$ satisfy  Assumption \ref{AS1} with $\tilde{U}=U_Y+U_M$. 

\subsubsection{Lemmas.}
Then, we obtain the following results.
\begin{lemma}
\label{LEM1}
Under SCM ${\cal M}$, and Assumptions \ref{SUP2} and \ref{AS2}, 
\begin{equation}
\label{eq17}
\begin{aligned}
&\text{\normalfont CD-PNS}(y;x',x,m,c)\\
&\hspace{0.2cm}=\max\Big\{\mathbb{P}(Y_{x',m} \prec y|C=c)-\mathbb{P}(Y_{x,m} \prec y|C=c),0\Big\}.
\end{aligned}
\end{equation}
\end{lemma}
\begin{lemma}
\label{LEM2}
Under SCM ${\cal M}$, and Assumptions \ref{SUP1} and \ref{AS1}, 
\begin{equation}
\label{eq7}
\begin{aligned}
&\text{\normalfont ND-PNS}(y;x',x,c)\\
&\hspace{0.2cm}=\max\Big\{\min\{\mathbb{P}(Y_{x'} \prec y|C=c),\mathbb{P}(Y_{x',M_{x}} \prec y|C=c)\}\\
&\hspace{3cm}-\mathbb{P}(Y_{x} \prec y|C=c),0\Big\},
\end{aligned}
\end{equation}
\begin{equation}
\label{eq8}
\begin{aligned}
&\text{\normalfont NI-PNS}(y;x',x,c)=\max\Big\{\mathbb{P}(Y_{x'} \prec y|C=c)\\
&\hspace{0.5cm}-\max\{\mathbb{P}(Y_{x} \prec y|C=c),\mathbb{P}(Y_{x',M_{x}} \prec y|C=c)\},0\Big\}.
\end{aligned}
\end{equation}
\end{lemma}
\noindent The lemmas mean that, under monotonicity, CD-PNS, ND-PNS, and NI-PNS can be computed from the CDF of certain counterfactual outcomes.



\subsubsection{Identification theorems.}
{The CDF of the counterfactual outcomes $\mathbb{P}(Y_{x,M_{x'}}\prec y|C=c)$ is identifiable under the sequential ignorability Assumption~\ref{SCAS2} by Proposition \ref{prop2} as $\mathbb{P}(Y_{x,M_{x'}}\prec y|C=c) = \rho(y;x',x,c)$, where we donote}
\begin{equation}
\begin{aligned}
&\rho(y;x',x,c)\defeq \int_{\Omega_{M}}\mathbb{P}(Y\prec  y|X=x',M=m,C=c)\\
&\hspace{2.5cm}\times\mathfrak{p}(M=m|X=x,C=c) dm.
\end{aligned}
\end{equation}
Then, we obtain the following identification theorems by combining Lemmas \ref{LEM1} and \ref{LEM2} and Proposition \ref{prop2}:
\begin{theorem}[Identification of CD-PNS]
\label{theo1}
Under SCM ${\cal M}$, and Assumptions \ref{SCAS2}, \ref{SUP2}, and \ref{AS2}, 
CD-PNS is identifiable by 
\begin{equation}
\begin{aligned}
&\text{\normalfont CD-PNS}(y;x',x,m,c)=\\
&\hspace{1cm}\min\Big\{\mathbb{P}(Y \prec y|X=x',M=m,C=c)\\
&\hspace{2cm}-\mathbb{P}(Y \prec y|X=x,M=m,C=c),0\Big\}.
\end{aligned}
\end{equation}
\end{theorem}
\begin{theorem}[Identification of ND-PNS and NI-PNS]
\label{theo2}
Under SCM ${\cal M}$, and Assumptions \ref{SCAS2}, \ref{SUP1}, and \ref{AS1},  
ND-PNS and NI-PNS are  identifiable by
\begin{equation}
\begin{aligned}
&\text{\normalfont ND-PNS}(y;x',x,c)=\max\Big\{\min\{\mathbb{P}(Y \prec y|X=x',C=c),\\
&\hspace{1cm}\rho(y;x',x,c)\}-\mathbb{P}(Y \prec y|X=x,C=c),0\Big\},
\end{aligned}
\end{equation}
\begin{equation}
\begin{aligned}
&\text{\normalfont NI-PNS}(y;x',x,c)=\max\Big\{\mathbb{P}(Y \prec y|X=x',C=c)\\
&\hspace{0.5cm}-\max\{\mathbb{P}(Y \prec y|X=x,C=c),\rho(y;x',x,c)\},0\Big\}.
\end{aligned}
\end{equation}
\end{theorem}
\noindent As a consequence, under SCM ${\cal M}$ and Assumptions \ref{SCAS2}, \ref{SUP1}, and \ref{AS1},  the proportions of ND-PNS and NI-PNS in T-PNS are also identifiable. 

\section{Direct and Indirect PNS with Evidence}

In this section, we define CD-PNS, ND-PNS, and NI-PNS with evidence and provide corresponding identification theorems. 
Specifically, we consider 
two types of evidence:
\begin{equation}
{\cal E}\defeq(X=x^*,M=m^*,Y\in {\cal I}_Y),
\end{equation}
\begin{equation}
{\cal E}'\defeq(X=x^*, Y\in {\cal I}_Y),
\end{equation}
where ${\cal I}_Y$ is a half-open interval $[y^l,y^u)$ or a closed interval $[y^l,y^u]$ w.r.t. $\prec$. 
PNS with evidence allows us to examine PNS for a specific subpopulation characterized by the evidence.  

The main distinction between the evidence $\cal{E}$ or $\cal{E}'$ and the subject's covariates $C$ in the definition of CD-PNS, ND-PNS, and NI-PNS (Def.~\ref{def3}) is that $C$ in the SCM $\cal{M}$ are pre-treatment variables but  $\cal{E}$ are post-treatment variables.
Conditioning on post-treatment variables differs from traditional conditioning on pre-treatment variables and has been discussed in the context of PN or PS \citep{Pearl1999} and the posterior causal effects \citep{Lu2022,Li2023B}. 
They have applications in various fields, such as attribution of risk factors in public health and epidemiology, medical diagnosis of diseases, root-cause diagnosis in equipment and production processes, and reference measures for penalties in law.

\subsection{Definitions of CD-PNS, ND-PNS, and NI-PNS with evidence}
First, we define CD-PNS with evidence ${\cal E}$, and T-PNS, ND-PNS, and NI-PNS with evidence ${\cal E}'$ \footnote{\citet{Kawakami2024} have studied T-PNS with evidence $(X=x^*, Y=y^*)$, which is a special case of  ${\cal E}'$.}. 
\begin{definition}[CD-PNS, T-PNS, ND-PNS, and NI-PNS with evidence]
\label{def4}
CD-PNS with evidence ${\cal E}$, and T-PNS, ND-PNS, and NI-PNS with evidence ${\cal E}'$ are defined by
\begin{equation}
\begin{aligned}
&\text{\normalfont CD-PNS}(y;x',x,m,{\cal E},c)\defeq\mathbb{P}(Y_{x',m} \prec y \preceq Y_{x,m}|{\cal E},C=c),
\end{aligned}
\end{equation}
\begin{equation}
\begin{aligned}
&\text{\normalfont T-PNS}(y;x',x,{\cal E}',c)\defeq\mathbb{P}(Y_{x'} \prec y \preceq Y_{x}|{\cal E}',C=c),
\end{aligned}
\end{equation}
\begin{equation}
\begin{aligned}
&\text{\normalfont ND-PNS}(y;x',x,{\cal E}',c)\\
&\hspace{0.2cm}\defeq\mathbb{P}(Y_{x'} \prec y \preceq Y_{x},  Y_{x',M_{x}} \prec y|{\cal E}',C=c),
\end{aligned}
\end{equation}
\begin{equation}
\begin{aligned}
&\text{\normalfont NI-PNS}(y;x',x,{\cal E}',c)\\
&\hspace{0.2cm}\defeq\mathbb{P}(Y_{x'} \prec y \preceq Y_{x},y \preceq Y_{x',M_{x}}|{\cal E}',C=c).
\end{aligned}
\end{equation}
\end{definition}
\noindent CD-PNS with evidence can answer questions: ``What is the probability that the situation in the question (Q1) holds for the subjects, when, in reality, their treatment is $x^*$, their mediator is $m^*$, their outcome is in ${\cal I}_Y$, and their covariates is $c$?\,"
ND-PNS and NI-PNS with evidence can answer questions: ``What is the probability that the situation in the questions (Q2) and (Q3) hold,  when, in reality, their treatment is $x^*$, their outcome is in ${\cal I}_Y$, and their covariates is $c$?\,"
CD-PNS, ND-PNS, and NI-PNS with evidence can retrospectively answer questions for the specific subpopulation characterized by the evidence.

The following desired decomposition property holds:
\begin{proposition}
\begin{equation}
\begin{aligned}
&\text{\normalfont T-PNS}(y;x',x,{\cal E}',c)\\
&=\text{\normalfont ND-PNS}(y;x',x,{\cal E}',c)+\text{\normalfont NI-PNS}(y;x',x,{\cal E}',c).
\end{aligned}
\end{equation}
\end{proposition}

\paragraph{Remark 3.}
We do not use mediator information in evidence for T-PNS, ND-PNS, and NI-PNS because a more strict assumption is required for identification to exploit mediator information. 
In Appendix C, we provide an identification theorem (Theorem 4') of T-PNS, ND-PNS, and NI-PNS with evidence ${\cal E}'=(X=x^*, M\in {\cal I}_M, Y\in {\cal I}_Y)$ with an additional assumption, where ${\cal I}_M$ is a half-open interval $[m^l,m^u)$ w.r.t. the total order on $\Omega_M$.

\subsubsection{ND-PN, NI-PN, ND-PS, and NI-PS.}
So far, we have focused our attention on PNS in the PoC family. It turns out that PN and PS, the other two members of the PoC family defined in Def.~\ref{def41}, can be computed as special cases of T-PNS with evidence.
Specifically, PN is equivalent to T-PNS with the evidence ${\cal E}'=(y \preceq Y,X=x)$, and PS is equivalent to T-PNS with the evidence ${\cal E}'=(Y \prec y,X=x')$ as follows.
\begin{proposition}
\label{prop3}
We have the following:
\begin{equation}
\begin{aligned}
&\text{\normalfont PN}(y;x',x,c)=\mathbb{P}(Y_{x'} \prec y \preceq Y_{x}|y \preceq Y,X=x,C=c),
\end{aligned}
\end{equation}
\begin{equation}
\begin{aligned}
&\text{\normalfont PS}(y;x',x,c)=\mathbb{P}(Y_{x'} \prec y \preceq Y_{x}|Y \prec y,X=x',C=c).
\end{aligned}
\end{equation}
\end{proposition}

Then, direct and indirect PN and PS can be naturally defined by extending the definitions of ND-PNS and NI-PNS with evidence in Def.~\ref{def4}. 
\begin{definition}[ND-PN, NI-PN, ND-PS, and NI-PS]
The natural direct PN (ND-PN), natural indirect PN (NI-PN), natural direct PS (ND-PS), and natural indirect PS (NI-PS) are defined by
\begin{equation}
\begin{aligned}
&\text{\normalfont ND-PN}(y;x',x,c)\\
&\hspace{0.2cm}\defeq\mathbb{P}(Y_{x'} \prec y,  Y_{x',M_{x}} \prec y|y \preceq Y,X=x,C=c),
\end{aligned}
\end{equation}
\begin{equation}
\begin{aligned}
&\text{\normalfont NI-PN}(y;x',x,c)\\
&\hspace{0.2cm}\defeq\mathbb{P}(Y_{x'} \prec y, y \preceq Y_{x',M_{x}}|y \preceq Y,X=x,C=c),
\end{aligned}
\end{equation}
\begin{equation}
\begin{aligned}
&\text{\normalfont ND-PS}(y;x',x,c)\\
&\hspace{0.2cm}\defeq\mathbb{P}(y \preceq Y_{x},  Y_{x',M_{x}} \prec y|Y \prec y,X=x',C=c),
\end{aligned}
\end{equation}
\begin{equation}
\begin{aligned}
&\text{\normalfont NI-PS}(y;x',x,c)\\
&\hspace{0.2cm}\defeq\mathbb{P}(y \preceq Y_{x}, y \preceq Y_{x',M_{x}}|Y \prec y,X=x',C=c).
\end{aligned}
\end{equation}
\end{definition}
ND-PN, NI-PN, ND-PS, and NI-PS provide a measure of the necessity or the sufficiency of the treatment for the outcome through direct or indirect pathways. We have the desirable decomposition property that $\text{\normalfont PN}(y;x',x,c)=\text{\normalfont ND-PN}(y;x',x,c)+\text{\normalfont NI-PN}(y;x',x,c)$ and $\text{\normalfont PS}(y;x',x,c)=\text{\normalfont ND-PS}(y;x',x,c)+\text{\normalfont NI-PS}(y;x',x,c)$.


\subsection{Identification of CD-PNS, T-PNS, ND-PNS, and NI-PNS with evidence}
We obtain the following two identification theorems under the same assumptions for Theorems \ref{theo1} or \ref{theo2}.
\begin{theorem}[Identification of CD-PNS with evidence ${\cal E}$]
\label{theo3}
Let ${\cal I}_Y$ be a half-open interval $[y^l,y^u)$ in evidence ${\cal E}$. Under SCM ${\cal M}$, and Assumptions \ref{SCAS2}, \ref{SUP2}, and \ref{AS2},  for each $x',x \in \Omega_X$, $m \in \Omega_M$, $y \in \Omega_Y$, and $c \in \Omega_C$, 
we have
\vspace{0.1cm}

\noindent {\normalfont (A).}  If $\mathbb{P}(Y \prec y^u|X=x^*,M=m^*,C=c)\ne\mathbb{P}(Y \prec y^l|X=x^*,M=m^*,C=c)$, then 
\begin{equation}\label{eq-cdpnsid}
\begin{aligned}
\text{\normalfont CD-PNS}(y;x',x,m,{\cal E},c)=\max\left\{{\alpha}/{\beta},0\right\},
\end{aligned}
\end{equation}
where 
\begin{equation}
\begin{aligned}
&\alpha=\min\Big\{\mathbb{P}(Y \prec y|X=x',M=m,C=c),\\
&\hspace{2.5cm}\mathbb{P}(Y \prec y^u|X=x^*,M=m^*,C=c)\Big\}\\
&\hspace{1cm}-\max\Big\{\mathbb{P}(Y \prec y|X=x,M=m,C=c),\\
&\hspace{2cm}\mathbb{P}(Y \prec y^l|X=x^*,M=m^*,C=c)\Big\},
\end{aligned}
\end{equation}
\begin{equation}
\begin{aligned}
&\beta=\mathbb{P}(Y \prec y^u|X=x^*,M=m^*,C=c)\\
&\hspace{1cm}-\mathbb{P}(Y \prec y^l|X=x^*,M=m^*,C=c).
\end{aligned}
\end{equation}
\vspace{0.1cm}

\noindent {\normalfont (B).}  If $\mathbb{P}(Y \prec y^u|X=x^*,M=m^*,C=c)=\mathbb{P}(Y \prec y^l|X=x^*,M=m^*,C=c)$, then 
\begin{equation}
\begin{aligned}
&\text{\normalfont CD-PNS}(y;x',x,m,{\cal E},c)=\mathbb{I}\Big(\mathbb{P}(Y \prec y|X=x',C=c) \\
&\hspace{0cm}\leq\mathbb{P}(Y \prec y^l|X=x^*,M=m^*,C=c)< \mathbb{P}(Y_{x} \prec y|C=c)\Big).
\end{aligned}
\end{equation}
\end{theorem}

\begin{theorem}[Identification of T-PNS, ND-PNS, and NI-PNS with evidence ${\cal E}'$]
\label{theo4}
Let ${\cal I}_Y$ be a half-open interval $[y^l,y^u)$ in evidence ${\cal E}'$. 
Under SCM ${\cal M}$, and Assumptions \ref{SCAS2}, \ref{SUP1}, and \ref{AS1}, for each $x',x \in \Omega_X$, $y \in \Omega_Y$, and $c \in \Omega_C$, we have 
\vspace{0.1cm}

\noindent {\normalfont (A).} If $\mathbb{P}(Y \prec y^u|X=x^*,C=c)\ne\mathbb{P}(Y \prec y^l|X=x^*,C=c)$, then 
\begin{equation}
\text{\normalfont T-PNS}(y;x',x,{\cal E}',c)=\max\left\{{\gamma^T}/{\delta},0\right\},
\end{equation}
\begin{equation}
\text{\normalfont ND-PNS}(y;x',x,{\cal E}',c)=\max\left\{{\gamma^D}/{\delta},0\right\},
\end{equation}
\begin{equation}
\text{\normalfont NI-PNS}(y;x',x,{\cal E}',c)=\max\left\{{\gamma^I}/{\delta},0\right\},
\end{equation}
where 
\begin{equation}
\begin{aligned}
&\gamma^T=\min\Big\{\mathbb{P}(Y \prec y|X=x',C=c),\\
&\hspace{2.5cm}\mathbb{P}(Y \prec y^u|X=x^*,C=c)\}\\
&\hspace{1cm}-\max\{\mathbb{P}(Y \prec y|X=x,C=c),\\
&\hspace{3cm}\mathbb{P}(Y \prec y^l|X=x^*,C=c)\Big\},
\end{aligned}
\end{equation}
\begin{equation}
\begin{aligned}
&\gamma^D=\min\Big\{\mathbb{P}(Y \prec y|X=x',C=c),\\
&\hspace{2.5cm}\mathbb{P}(Y \prec y^u|X=x^*,C=c),\rho(y;x',x,c)\}\\
&\hspace{1cm}-\max\{\mathbb{P}(Y \prec y|X=x,C=c),\\
&\hspace{3cm}\mathbb{P}(Y \prec y^l|X=x^*,C=c)\Big\},
\end{aligned}
\end{equation}
\begin{equation}
\begin{aligned}
&\gamma^I=\min\Big\{\mathbb{P}(Y \prec y|X=x',C=c),\\
&\hspace{3cm}\mathbb{P}(Y \prec y^u|X=x^*,C=c)\}\\
&\hspace{1cm}-\max\{\mathbb{P}(Y \prec y|X=x,C=c),\\
&\hspace{2cm}\mathbb{P}(Y \prec y^l|X=x^*,C=c),\rho(y;x',x,c)\Big\},
\end{aligned}
\end{equation}

\begin{equation}
\begin{aligned}
&\delta=\mathbb{P}(Y \prec y^u|X=x^*,C=c)-\mathbb{P}(Y \prec y^l|X=x^*,C=c).
\end{aligned}
\end{equation}
\vspace{0.1cm}

\noindent {\normalfont (B).}  If $\mathbb{P}(Y \prec y^u|X=x^*,C=c)=\mathbb{P}(Y \prec y^l|X=x^*,C=c)$, then 
\begin{equation}
\begin{aligned}
&\text{\normalfont T-PNS}(y;x',x,{\cal E}',c)=\mathbb{I}\Big(\mathbb{P}(Y \prec y|X=x',C=c)\leq\\
&\hspace{0cm}\mathbb{P}(Y \prec y^l|X=x^*,C=c) < \mathbb{P}(Y \prec y|X=x,C=c)\Big),
\end{aligned}
\end{equation}
\begin{equation}
\begin{aligned}
&\text{\normalfont ND-PNS}(y;x',x,{\cal E}',c)=\mathbb{I}\Big(\mathbb{P}(Y \prec y|X=x',C=c)\leq\\
&\hspace{0.5cm}\mathbb{P}(Y \prec y^l|X=x^*,C=c) < \mathbb{P}(Y \prec y|X=x,C=c),\\
&\hspace{1.5cm}\rho(y;x',x,c)\leq \mathbb{P}(Y \prec y^l|X=x^*,C=c)\Big),
\end{aligned}
\end{equation}
\begin{equation}
\begin{aligned}
&\text{\normalfont NI-PNS}(y;x',x,{\cal E}',c)=\mathbb{I}\Big(\mathbb{P}(Y \prec y|X=x',C=c)\leq\\
&\hspace{0.5cm}\mathbb{P}(Y \prec y^l|X=x^*,C=c) < \mathbb{P}(Y \prec y|X=x,C=c),\\
&\hspace{1.5cm} \mathbb{P}(Y \prec y^l|X=x^*,C=c)<\rho(y;x',x,c)\Big).
\end{aligned}
\end{equation}
\end{theorem}
\paragraph{Remark 4.} 
When ${\cal I}_Y$ is a closed intervel $[y^l,y^u]$ in evidence ${\cal E}$ or ${\cal E}'$, the identification results are obtained by changing ``$Y \prec y^u$" to ``$Y \preceq y^u$" in Theorems \ref{theo3} and \ref{theo4}.  

\paragraph{Remark 5.}  
When ${\cal I}_Y$ is a point   $y^l=y^u$,
the identification of T-PNS with evidence $(X=x^*,Y=y^l)$ in Theorem \ref{theo4} reduces to Theorem 5.1 in \citep{Kawakami2024}.
Thus, T-PNS identification in Theorem \ref{theo4} is an extension of Theorem 5.1 in \citep{Kawakami2024}.

\section{Simulated Experiments}
\subsection{Estimation from finite sample size}
We perform numerical experiments to illustrate the properties of the estimators from finite sample size.
Theoretically, the estimators in this paper are consistent and it is expected that the estimates are reliable when the sample size is large.

\subsubsection{Estimation methods.}
All identification theorems in the paper compute all quantities through conditional CDFs. 
Using  dataset $\{x_i,m_i,y_i\}_{i=1}^N$, 
we estimate the conditional CDFs by the empirical conditional CDFs, i.e., 
$\hat{\mathbb{P}}(Y \prec y|X=x,M=m)={\sum\nolimits_{i=1}^N\mathbb{I}(y_i\prec y,x_i=x,m_i=m)}/{\sum\nolimits_{i=1}^N\mathbb{I}(x_i=x,m_i=m)}$,
$\hat{\mathbb{P}}(M=m|X=x)={\sum\nolimits_{i=1}^N\mathbb{I}(m_i=m,x_i=x)}/{\sum\nolimits_{i=1}^N\mathbb{I}(x_i=x)}$,   
and, in addition,
$\hat{\rho}(y;x',x)=\sum_{m \in \Omega_{M}}\hat{\mathbb{P}}(Y\prec y|X=x',M=m)\hat{\mathbb{P}}(M=m|X=x)$. 
We conduct the bootstrapping \citep{Efron1979} to reveal the distribution of the estimators,
and provide the means and 95\% confidential intervals (CI) for each estimator.

\subsubsection{Setting.}
We assume the following SCM:
\begin{gather}
X:=\text{Bern}(0.5), M:=\text{Bern}(\pi(X)),\nonumber\\
Y:=\text{Bern}(\pi(X+M)),
\end{gather}
where $\pi(x)=\text{exp}(1+0.5x)/(1+\text{exp}(1+0.5x))$.
$\text{Bern}(z)$ represents a Bernoulli distribution with probability $z$.
$X$, $M$, and $Y$ are all binary variables.
We simulate 1000 times with the sample size $N=100,1000,10000$, respectively, and assess the means and 95$\%$ confidential intervals (CIs) of the estimators.

\subsubsection{Results.}
The ground truths of T-PNS, ND-PNS, and NI-PNS are $0.074$, $0.066$, and $0.008$.
When $N=100$, the estimates are 
\begin{center}
T-PNS: $0.083$ (CI: [$0.000,0.228$]),\\\vspace{0.1cm}
ND-PNS: $0.074$ (CI: [$0.000,0.220$]),\\\vspace{0.1cm}
NI-PNS: $0.009$ (CI: [$0.000,0.046$]).
\end{center}
When $N=1000$, the estimates are
\begin{center}
T-PNS: $0.075$ (CI: [$0.029,0.125$]),\\\vspace{0.1cm}
ND-PNS: $0.068$ (CI: [$0.021,0.116$]),\\\vspace{0.1cm}
NI-PNS: $0.007$ (CI: [$0.000,0.017$]).
\end{center}
When $N=10000$, the estimates are
\begin{center}
T-PNS: $0.074$ (CI: [$0.060,0.088$]),\\\vspace{0.1cm}
ND-PNS: $0.067$ (CI: [$0.052,0.082$]),\\\vspace{0.1cm}
NI-PNS: $0.008$ (CI: [$0.005,0.011$]).
\end{center}
When the sample size is small ($N=100$), the estimators have relatively wide 95$\%$ CIs.
When the sample size is large enough ($N=1000$ or $N=10000$), the estimators are close to the ground truths and have relatively narrow 95$\%$ CIs.
We perform additional experiments for T-PN, ND-PN, NI-PN, T-PS, ND-PS, and NI-PS and the results are presented in Appendix F.

\subsection{Illustration of the proposed measures}
To illustrate the behavior of the proposed direct and indirect PoC measures, we simulate data from an SCM and plot the measures against the covariate. The results are discussed in Appendix F.

\section{Application to a real-world dataset}

We show an application to a real-world psychology dataset.

\subsubsection{Dataset.}
We take up a dataset from the Job Search Intervention Study (JOBS II) \citep{Vinokur1997}.
This dataset is open through the R package ``mediation" (\url{https://cran.r-project.org/web/packages/mediation/index.html}).
JOBS II was a randomized job training intervention for unemployed subjects aiming at increasing the prospect of reemployment and improving their mental health.
In the experiment, the unemployed workers were randomly assigned to treatment and control groups. 
Those in the treatment group participated in job-skills workshops, and they learned job-search skills and coping strategies for dealing with setbacks in the job-search process. 
Those in the control group received a booklet of job-search tips. 
In follow-up interviews, a measure of depressive symptoms based on the Hopkins Symptom Checklist was assessed. 
The sample size is 899 with no missing values.

\subsubsection{Variables.}
Let the randomly assigned interventions be treatment variable ($X$) \texttt{(treat)},  which takes $0$ for the control group and $1$ for the treatment group.
We choose the measure of depressive symptoms based on the Hopkins Symptom Checklist \texttt{(depress2)} as the outcome ($Y$).
We consider job-search self-efficacy ($M$) \texttt{(job\_seek)} as a discrete mediating variable.
{We set $C=\emptyset$.}
We let the threshold of the depression be $y=3$ in all the definitions of PoC variants, and let $x'=0$ and $x=1$.
We assume Assumptions \ref{SCAS2}-\ref{SUP1}. 
These are reasonable because the interventions are randomly assigned, and the linear model used in the previous study \citep{Vinokur1997}  satisfies these assumptions.
On this dataset, it is reasonable that $X=0$ increases the depression compared to $X=1$ and we assume \ref{AS2}' and \ref{AS1}' for monotonic increasing.
Assumption \ref{AS2}' for monotonic increasing represents $\mathbb{P}(Y_{1,m}\succ y \succeq Y_{0,m}|C=c)=0$, which means that
there do not exist subjects whose potential depression score when setting the value of job-search self-efficiency by $m$ and receiving no intervention is under the given threshold $y$, 
and whose potential depression score when setting the value of job-search self-efficiency by $m$ and receiving an intervention is over the given threshold $y$.
This seems reasonable.
Assumption \ref{AS1}' for monotonic increasing represents $\mathbb{P}(Y_{1,M_{1}}\succ y \succeq Y_{1,M_{0}}|C=c)=0$, $\mathbb{P}(Y_{1,M_{1}}\succ y \succeq Y_{0,M_{1}}|C=c)=0$, $\mathbb{P}(Y_{1,M_{1}}\succ y \succeq Y_{0,M_{0}}|C=c)=0$, $\mathbb{P}(Y_{1,M_{0}}\succ y \succeq Y_{0,M_{0}}|C=c)=0$, and $\mathbb{P}(Y_{0,M_{1}}\succ y \succeq Y_{0,M_{0}}|C=c)=0$.
For example, $\mathbb{P}(Y_{1,M_{1}}\succ y \succeq Y_{1,M_{0}}|C=c)=0$ means that 
there do not exist subjects whose potential depression score when receiving an intervention and keeping the value of job-search self-efficiency by $M_0$ is under the given threshold $y$, 
and whose potential depression score when receiving an intervention and keeping the value of job-search self-efficiency by $M_1$ is over the given threshold $y$.
This also seems reasonable.

\subsubsection{Results.}
The estimated T-PNS is 23.840\% (CI: [19.021\%,29.254\%]). 
Then, we consider the following three questions:
\begin{center}
\vspace{0.1cm}
({\bf Q1'}). {\it Would the intervention be necessary and sufficient to cure the depression had the job-search self-efficacy been fixed to a value $(m=5)$?}\\\vspace{0.1cm}
({\bf Q2'}). {\it Would the intervention still be necessary and sufficient to cure the depression had there been no influence via the job-search self-efficacy?}\\\vspace{0.1cm}
({\bf Q3'}). {\it Would the intervention still be necessary and sufficient to cure the depression had the influence only existed via the job-search self-efficacy?}\vspace{0.1cm}
\end{center}
We evaluate CD-PNS $(m=5)$, ND-PNS, and NI-PNS specified in Def.~\ref{def3} and obtain the following results:
\begin{center}
\vspace{0cm}
CD-PNS: 7.484\% (CI: [0.000\%,41.676\%]),\\\vspace{0.1cm}
ND-PNS: 0.000\% (CI: [0.000\%,0.000\%]),\\\vspace{0.1cm}
NI-PNS: 23.840\% (CI: [19.021\%,29.254\%]).
\vspace{0cm}
\end{center}
CD-PNS, ND-PNS, and NI-PNS answer the questions (Q1'), (Q2'), and (Q3'), respectively.
CD-PNS is less than T-PNS.
T-PNS is equal to NI-PNS, and this means that the necessity and sufficiency of the treatment is entirely due to the indirect influence via the mediator.
The proportions of ND-PNS and NI-PNS in T-PNS are $0$ and $1$.

While \citet{Vinokur1997}  reported both direct and indirect effects as statistically significant,  our results decompose the total influence entirely into the indirect.
However, this does not contradict the observation that treatment has a direct effect on the outcome. 
Our results imply that the treatment would be necessary and sufficient at the same level of T-PNS  had the influence only existed via the mediator, and the treatment would not be necessary and sufficient had there been no influence via the mediator.
Our results do not imply that the treatment has no direct effect on outcome.

Next, we study PoC for a specific subpopulation described by evidence.
We evaluate T-PNS, CD-PNS $(m=5)$, ND-PNS, and NI-PNS with evidence specified in Def.~\ref{def4}.
We consider the evidence of $x^*=0$, ${\cal I}_Y=[y^l,y^u)$ where $y^l=1.5$ and $y^u=2.5$ for ND-PNS and NI-PNS, and additionally $m^*=5$ for CD-PNS. 
We obtain: 
\begin{center}
\vspace{0cm}
T-PNS with evidence: 57.899\%(CI: [39.130\%,76.190\%]),\\\vspace{0.1cm}
CD-PNS with evidence: 0.000\%(CI: [0.000\%,0.000\%])\footnote{{The result of bootstrap CI width $0$ 
is due to the ``max'' function in the estimators. In Eq.~\ref{eq-cdpnsid}, CD-PNS  with evidence are identified using the ``max'' function, i.e., $\max\{\cdot,0\}$. 
If the upper bound of the bootstrap CI of the inside value of the ``max'' function is negative (i.e., 95\% chance the inside value is in a range all negative), the estimated CD-PNS  is  0\% with bootstrap CI width 0. The interpretation is that CD-PNS is 0\% with 95\% confidence.}},\\\vspace{0.1cm}
ND-PNS with evidence: 0.000\%(CI: [0.000\%,0.000\%]),\\\vspace{0.1cm}
NI-PNS with evidence: 57.899\%(CI: [39.130\%,76.190\%]).
\vspace{0.1cm}
\end{center}
\noindent CD-PNS, ND-PNS, and NI-PNS can answer the questions (Q1'), (Q2'), and (Q3'), respectively, for the subpopulation specified by the evidence.
CD-PNS is $0$, and the proportions of ND-PNS and NI-PNS in T-PNS are $0$ and $1$.
T-PNS and NI-PNS for this subpopulation are larger than those of the whole population.

\section{Conclusion}

We consider mediation analysis for PoC and introduce new direct and indirect variants of PoC to represent the necessity and sufficiency of the treatment to produce an outcome event directly or through a mediator. We provide identification theorems for each type of PoC we introduce. The results expand the family of PoC and provide tools for researchers to answer more sophisticated causal questions. 
In addition, we show in Appendix D how these direct and indirect variants of PoC look like for binary treatment, outcome, and mediator variables. 
In settings where the identification assumptions (sequential ignorability, monotonicity) do not hold, bounding \citep{Tian2000,Dawid2017,Dawid2024,Li2022} or sensitivity analysis \citep{Imai2010a,Imai2010b,Vanderweele2016}  is desired.
Also, researchers are often interested in path-specific effects, of which direct and indirect effects are special instances \citep{Daniel2015,Xia2022,Zhou2023}.
Extending our results to these cases will be interesting future work.

\section*{Acknowledgements}
The authors thank the anonymous reviewers for their time and thoughtful comments.

\bibliography{main_arXiv}

\appendix

\section*{Appendix A: Definition of total order}
We show the definition of the total order used in the body of the paper.

\subsubsection{Total orders.}
The definition of the total order is as follows \citep{Harzheim2005}:
\begin{definition}[Total order]
    A total order on a set $\Omega$ is a relation ``$\preceq $'' on $\Omega$ satisfying the following four conditions for all $a_1, a_2, a_3 \in \Omega$:
    \begin{enumerate}
    \vspace{-0cm}
      \setlength{\parskip}{1pt}
  \setlength{\itemsep}{1pt}
        \item $a_1\preceq a_1$;
        \item if $a_1\preceq a_2$ and $a_2\preceq a_3$ then $a_1\preceq a_3$;
        \item if $a_1\preceq a_2$ and $a_2\preceq a_1$ then $a_1= a_2$;
        \item at least one of $a_1\preceq a_2$ and $a_2\preceq a_1$ holds.
    \end{enumerate}
    \vspace{-0cm}
\end{definition}
\noindent In this case we say that the ordered pair $(\Omega,\preceq)$ is a totally ordered set. 
The inequality $a\preceq  b$ of total order means $a\prec b$ or $a=b$, and the relationship $\lnot (a\preceq  b) \Leftrightarrow a\succ b$ holds for a totally ordered set, where $\lnot$ means the negation.

\section*{Appendix B: Proofs}

We provide the proofs of propositions, lemmas, theorems, and statements in the body of the paper.\\

\noindent{\bf Proposition 2.}
{\it  
We have
\begin{equation}
\begin{aligned}
&\text{\normalfont T-PNS}(y;x',x,c)\\
&=\text{\normalfont ND-PNS}(y;x',x,c)+\text{\normalfont NI-PNS}(y;x',x,c).
\end{aligned}
\end{equation}
}

\begin{proof}
We have
\begin{equation}
\begin{aligned}
&\text{\normalfont T-PNS}(y;x',x,c)\\
&=\mathbb{P}(Y_{x'} \prec y \preceq Y_{x}|C=c)\\
&=\mathbb{P}(Y_{x'} \prec y \preceq Y_{x}, y \preceq Y_{x',M_{x}}|C=c)\\
&+\mathbb{P}(Y_{x'} \prec y \preceq Y_{x}, Y_{x',M_{x}} \prec y|C=c)\\
&=\text{\normalfont ND-PNS}(y;x',x,c)+\text{\normalfont NI-PNS}(y;x',x,c)
\end{aligned}
\end{equation}
for any $x', x \in \Omega_X$, $y \in \Omega_Y$, and $c \in \Omega_C$.
\end{proof}\vspace{0.2cm}

\noindent{\bf Lemma 1.}
{\it
Under SCM ${\cal M}$, and Assumptions 2 and 4, for all $x',x \in \Omega_X$, $m \in \Omega_M$, $y \in \Omega_Y$, and $c \in \Omega_C$, we have
\begin{equation}
\begin{aligned}
&\text{\normalfont CD-PNS}(y;x',x,m,c)\\
&\hspace{0.2cm}=\max\Big\{\mathbb{P}(Y_{x',m} \prec y|C=c)-\mathbb{P}(Y_{x,m} \prec y|C=c),0\Big\}.
\end{aligned}
\end{equation}
}

\begin{proof}
Letting $u_{x',m,y}=\sup\{u_Y \in \Omega_{U_Y}; Y_{x',m}(u_Y) \prec y\}$, $u_{x,m,y}=\sup\{u_Y \in \Omega_{U_Y}; Y_{x,m}(u_Y) \prec y\}$ given $C=c$, from Assumptions 2 and 4, then we have 
\begin{equation}
\begin{aligned}
&\text{\normalfont CD-PNS}(y;x',x,m,c)\\
&=\mathbb{P}(Y_{x',m} \prec y \preceq Y_{x,m}|C=c)\\
&=\mathbb{P}_{U_Y}(u_{x',m,y}\preceq U_Y \prec u_{x,m,y}|C=c)\\
&=\max\{\mathbb{P}(Y_{x',m} \prec y|C=c)-\mathbb{P}(Y_{x,m} \prec y|C=c),0\}
\end{aligned}
\end{equation}
for all $x',x \in \Omega_X$, $m \in \Omega_M$, $y \in \Omega_Y$, and $c \in \Omega_C$.
\end{proof}\vspace{0.2cm}

\noindent{\bf Lemma 2.}
{\it
Under SCM ${\cal M}$, and Assumptions 3 and 5, for all $x',x \in \Omega_X$, $y \in \Omega_Y$, and $c \in \Omega_C$, we have
\begin{equation}
\begin{aligned}
&\text{\normalfont ND-PNS}(y;x',x,c)\\
&\hspace{0.2cm}=\max\Big\{\min\{\mathbb{P}(Y_{x'} \prec y|C=c),\mathbb{P}(Y_{x',M_{x}} \prec y|C=c)\}\\
&\hspace{3cm}-\mathbb{P}(Y_{x} \prec y|C=c),0\Big\},
\end{aligned}
\end{equation}
\begin{equation}
\begin{aligned}
&\text{\normalfont NI-PNS}(y;x',x,c)\\
&\hspace{0.2cm}=\max\Big\{\mathbb{P}(Y_{x'} \prec y|C=c)\\
&\hspace{0.5cm}-\max\{\mathbb{P}(Y_{x} \prec y|C=c),\mathbb{P}(Y_{x',M_{x}} \prec y|C=c)\},0\Big\}.
\end{aligned}
\end{equation}
}

\begin{proof}
Letting $\tilde{u}_{x',y}=\sup\{\tilde{u} \in \Omega_{\tilde{U}}; Y_{x'}(\tilde{u}) \prec y\}$, $\tilde{u}_{x,y}=\sup\{\tilde{u} \in \Omega_{\tilde{U}}; Y_{x}(\tilde{u}) \prec y\}$, and $\tilde{u}_{x',M_{x},y}=\sup\{\tilde{u} \in \Omega_{\tilde{U}}; Y_{x',M_{x}}(\tilde{u}) \prec y\}$ given $C=c$, from Assumptions 3 and 5, then we have 
\begin{equation}
\begin{aligned}
&\text{\normalfont ND-PNS}(y;x',x,c)\\
&=\mathbb{P}(Y_{x'} \prec y \preceq Y_{x},  Y_{x',M_{x}} \prec y|C=c)\\
&=\mathbb{P}_{\tilde{U}}(\tilde{u}_{x',y}\preceq \tilde{U} \prec \tilde{u}_{x,y},  \tilde{u}_{x',M_{x},y}\preceq \tilde{U}|C=c)\\
&=\max\{\min\{\mathbb{P}(Y_{x'} \prec y|C=c),\mathbb{P}(Y_{x',M_{x}} \prec y|C=c)\}\\
&\hspace{4cm}-\mathbb{P}(Y_{x} \prec y|C=c),0\},\\
\end{aligned}
\end{equation}
and
\begin{equation}
\begin{aligned}
&\text{\normalfont NI-PNS}(y;x',x,c)\\
&=\mathbb{P}(Y_{x'} \prec y \preceq Y_{x},y \preceq Y_{x',M_{x}}|C=c)\\
&=\mathbb{P}_{\tilde{U}}(\tilde{u}_{x',y}\preceq \tilde{U} \prec \tilde{u}_{x,y}, \tilde{U} \prec \tilde{u}_{x',M_{x},y}|C=c)\\
&=\max\{\mathbb{P}(Y_{x'} \prec y|C=c)\\
&\hspace{0.2cm}-\max\{\mathbb{P}(Y_{x} \prec y|C=c),\mathbb{P}(Y_{x',M_{x}} \prec y|C=c)\},0\}
\end{aligned}
\end{equation}
for all $x',x \in \Omega_X$, $y \in \Omega_Y$, and $c \in \Omega_C$.
\end{proof}\vspace{0.2cm}

\noindent{\bf Theorem 1.} (Identification of CD-PNS).
{\it
Under SCM ${\cal M}$, and Assumptions 1, 2, and 4, for all $x',x \in \Omega_X$, $m \in \Omega_M$, $y \in \Omega_Y$, and $c \in \Omega_C$, 
CD-PNS is identifiable by 
\begin{equation}
\begin{aligned}
&\text{\normalfont CD-PNS}(y;x',x,m,c)=\\
&\hspace{1cm}\min\Big\{\mathbb{P}(Y \prec y|X=x',M=m,C=c)\\
&\hspace{2cm}-\mathbb{P}(Y \prec y|X=x,M=m,C=c),0\Big\}.
\end{aligned}
\end{equation}
}

\begin{proof}
Since all elements in Lemma 1 are identifiable by Proposition 1, CD-PNS is identifiable as above.
\end{proof}\vspace{0.2cm}

\noindent{\bf Theorem 2.} (Identification of ND-PNS and NI-PNS).
{\it
Under SCM ${\cal M}$, and Assumptions 1, 3, and 5, for all $x',x \in \Omega_X$, $y \in \Omega_Y$, and $c \in \Omega_C$, 
ND-PNS and NI-PNS are identifiable by
\begin{equation}
\begin{aligned}
&\text{\normalfont ND-PNS}(y;x',x,c)\\
&\hspace{0.2cm}=\max\Big\{\min\{\mathbb{P}(Y \prec y|X=x',C=c),\rho(y;x',x,c)\}\\
&\hspace{2cm}-\mathbb{P}(Y \prec y|X=x,C=c),0\Big\},
\end{aligned}
\end{equation}
\begin{equation}
\begin{aligned}
&\text{\normalfont NI-PNS}(y;x',x,c)=\max\Big\{\mathbb{P}(Y \prec y|X=x',C=c)\\
&\hspace{0.5cm}-\max\{\mathbb{P}(Y \prec y|X=x,C=c),\rho(y;x',x,c)\},0\Big\}.
\end{aligned}
\end{equation}
}

\begin{proof}
Since all elements in Lemma 2 are identifiable by Proposition 1, CD-PNS is identifiable as above.
\end{proof}\vspace{0.2cm}

\noindent{\bf Proposition 3.}
{\it
\begin{equation}
\begin{aligned}
&\text{\normalfont T-PNS}(y;x',x,{\cal E}',c)\\
&=\text{\normalfont ND-PNS}(y;x',x,{\cal E}',c)+\text{\normalfont NI-PNS}(y;x',x,{\cal E}',c).
\end{aligned}
\end{equation}
}

\begin{proof}
We have
\begin{equation}
\begin{aligned}
&\text{\normalfont ND-PNS}(y;x',x,{\cal E}',c)+\text{\normalfont NI-PNS}(y;x',x,{\cal E}',c)\\
&=\mathbb{P}(Y_{x'} \prec y \preceq Y_{x}, Y_{x',M_{x}} \prec y{\cal E}'|C=c)\\
&\hspace{1cm}+\mathbb{P}(Y_{x'} \prec y \preceq Y_{x},y \preceq Y_{x',M_{x}}|{\cal E}',C=c)\\
&=\mathbb{P}(Y_{x'} \prec y \preceq Y_{x}|C=c)\\
&=\text{\normalfont T-PNS}(y;x',x,{\cal E}',c).
\end{aligned}
\end{equation}

\end{proof}



\noindent{\bf Theorem 3.} (Identification of CD-PNS with evidence ${\cal E}$).
{\it
Let ${\cal I}_Y$ be a half-open interval $[y^l,y^u)$ in evidence ${\cal E}$. Under SCM ${\cal M}$, and Assumptions 1, 2, and 4,  for each $x',x \in \Omega_X$, $m \in \Omega_M$, $y \in \Omega_Y$, and $c \in \Omega_C$, we have
\vspace{0.1cm}

\noindent {\normalfont (A).}  If $\mathbb{P}(Y \prec y^u|X=x^*,M=m^*,C=c)\ne\mathbb{P}(Y \prec y^l|X=x^*,M=m^*,C=c)$, then 
\begin{equation}
\begin{aligned}
\text{\normalfont CD-PNS}(y;x',x,m,{\cal E},c)=\max\left\{{\alpha}/{\beta},0\right\},
\end{aligned}
\end{equation}
where 
\begin{equation}
\begin{aligned}
&\alpha=\min\Big\{\mathbb{P}(Y \prec y|X=x',M=m,C=c),\\
&\hspace{2.5cm}\mathbb{P}(Y \prec y^u|X=x^*,M=m^*,C=c)\Big\}\\
&\hspace{1cm}-\max\Big\{\mathbb{P}(Y \prec y|X=x,M=m,C=c),\\
&\hspace{2cm}\mathbb{P}(Y \prec y^l|X=x^*,M=m^*,C=c)\Big\},
\end{aligned}
\end{equation}
\begin{equation}
\begin{aligned}
&\beta=\mathbb{P}(Y \prec y^u|X=x^*,M=m^*,C=c)\\
&\hspace{1cm}-\mathbb{P}(Y \prec y^l|X=x^*,M=m^*,C=c).
\end{aligned}
\end{equation}
\vspace{0.1cm}

\noindent {\normalfont (B).}  If $\mathbb{P}(Y \prec y^u|X=x^*,M=m^*,C=c)=\mathbb{P}(Y \prec y^l|X=x^*,M=m^*,C=c)$, then 
\begin{equation}
\begin{aligned}
&\text{\normalfont CD-PNS}(y;x',x,m,{\cal E},c)=\mathbb{I}\Big(\mathbb{P}(Y \prec y|X=x',C=c) \\
&\hspace{1cm}\leq\mathbb{P}(Y \prec y^l|X=x^*,M=m^*,C=c)\\
&\hspace{2.5cm}< \mathbb{P}(Y_{x} \prec y|C=c)\Big).
\end{aligned}
\end{equation}
}

\begin{proof}
From Assumptions 1, 2, and 4, letting 
$u_{x^*,m^*,y}=\sup\{u_Y \in \Omega_{U_Y}; Y_{x^*,m^*}(u_Y) \prec y\}$ and $u_{x^*,m^*,y}=\sup\{u_Y \in \Omega_{U_Y}; Y_{x^*,m^*}(u_Y) \prec y\}$
given $C=c$, if $\mathbb{P}(Y \prec y^u|X=x^*,M=m^*,C=c)\ne\mathbb{P}(Y \prec y^l|X=x^*,M=m^*,C=c)$, then
\begin{equation}
\begin{aligned}
&\text{\normalfont CD-PNS}(y;x',x,m,{\cal E},c)\\
&=\mathbb{P}(Y_{x',m} \prec y \preceq Y_{x,m}|X=x^*,M=m^*,Y\in {\cal I}_Y,C=c)\\
&={\mathbb{P}(Y_{x',m} \prec y \preceq Y_{x,m},Y\in {\cal I}_Y|X=x^*,M=m^*,C=c)}\\
&\hspace{2.5cm}\Big/{\mathbb{P}(Y\in {\cal I}_Y|X=x^*,M=m^*,C=c)}\\
&=\mathbb{P}(Y_{x',m} \prec y \preceq Y_{x,m},\\
&\hspace{2.5cm}Y_{x^*,m^*}\in {\cal I}_Y|X=x^*,M=m^*,C=c)\\
&\hspace{2cm}\Big/{\mathbb{P}(Y_{x^*,m^*}\in {\cal I}_Y|X=x^*,M=m^*,C=c)}\\
&=\mathbb{P}_{U_Y}(u_{x',m,y} \preceq u_Y \prec u_{x,m,y},\\
&\hspace{1cm}u_{x^*,m^*,y^l} \preceq u_Y \prec u_{x^*,m^*,y^u}|X=x^*,M=m^*,C=c)\\
&\hspace{0cm}\Big/{\mathbb{P}_{U_Y}(u_{x^*,m^*,y^l} \preceq u_Y \prec u_{x^*,m^*,y^u}|X=x^*,M=m^*,C=c)}\\
&=\max\left\{{\alpha}/{\beta},0\right\},
\end{aligned}
\end{equation}
where $\alpha=\min\{\mathbb{P}(Y \prec y|X=x',M=m,C=c),\mathbb{P}(Y \prec y^u|X=x^*,M=m^*,C=c)\}-\max\{\mathbb{P}(Y \prec y|X=x,M=m,C=c),\mathbb{P}(Y \prec y^l|X=x^*,M=m^*,C=c)\}$ and $\beta=\mathbb{P}(Y \prec y^u|X=x^*,M=m^*,C=c)-\mathbb{P}(Y \prec y^l|X=x^*,M=m^*,C=c)$, for all $x',x \in \Omega_X$, $m \in \Omega_M$, $y \in \Omega_Y$, and $c \in \Omega_C$.

From Assumptions 1, 2, and 4,
when $\mathbb{P}(Y \prec y^u|X=x^*,M=m^*,C=c)=\mathbb{P}(Y \prec y^l|X=x^*,M=m^*,C=c)$, then we have
\begin{equation}
\begin{aligned}
&\text{\normalfont CD-PNS}(y;x',x,m,{\cal E},c)\\
&=\mathbb{P}(Y_{x',m} \prec y \preceq Y_{x,m}|X=x^*,M=m^*,Y\in {\cal I}_Y,C=c)\\
&=\mathbb{P}_{U_Y}(u_{x',m,y} \preceq U_Y \preceq u_{x,m,y}|U_Y=u_{x^*,m^*,y^l}=u_{x^*,m^*,y^u},\\
&\hspace{2cm}X=x^*,M=m^*,C=c)\\
&=\mathbb{I}\Big(\mathbb{P}(Y_{x'} \prec y|C=c)\\
&\hspace{1cm} \leq\mathbb{P}(Y \prec y^l=y^u|X=x^*,M=m^*,C=c)\\
&\hspace{2cm}< \mathbb{P}(Y_{x} \prec y|C=c)\Big)
\end{aligned}
\end{equation}
for all $x',x \in \Omega_X$, $m \in \Omega_M$, $y \in \Omega_Y$, and $c \in \Omega_C$.
\end{proof}\vspace{0.2cm}

\noindent{\bf Theorem 4.} (Identification of T-PNS, ND-PNS, and NI-PNS with evidence ${\cal E}'$).
{\it
Let ${\cal I}_Y$ be a half-open interval $[y^l,y^u)$ in evidence ${\cal E}'$. Under SCM ${\cal M}$, and Assumptions 1, 3, and 5, for each $x',x \in \Omega_X$, $y \in \Omega_Y$, and $c \in \Omega_C$, we have 
\vspace{0.1cm}

\noindent {\normalfont (A).} If $\mathbb{P}(Y \prec y^u|X=x^*,C=c)\ne\mathbb{P}(Y \prec y^l|X=x^*,C=c)$, then 
\begin{equation}
\text{\normalfont T-PNS}(y;x',x,{\cal E}',c)=\max\left\{{\gamma^T}/{\delta},0\right\},
\end{equation}
\begin{equation}
\text{\normalfont ND-PNS}(y;x',x,{\cal E}',c)=\max\left\{{\gamma^D}/{\delta},0\right\},
\end{equation}
\begin{equation}
\text{\normalfont NI-PNS}(y;x',x,{\cal E}',c)=\max\left\{{\gamma^I}/{\delta},0\right\},
\end{equation}
where 
\begin{equation}
\begin{aligned}
&\gamma^T=\min\Big\{\mathbb{P}(Y \prec y|X=x',C=c),\\
&\hspace{2.5cm}\mathbb{P}(Y \prec y^u|X=x^*,C=c)\}\\
&\hspace{1cm}-\max\{\mathbb{P}(Y \prec y|X=x,C=c),\\
&\hspace{3cm}\mathbb{P}(Y \prec y^l|X=x^*,C=c)\Big\},
\end{aligned}
\end{equation}
\begin{equation}
\begin{aligned}
&\gamma^D=\min\Big\{\mathbb{P}(Y \prec y|X=x',C=c),\\
&\hspace{2.5cm}\mathbb{P}(Y \prec y^u|X=x^*,C=c),\rho(y;x',x,c)\}\\
&\hspace{1cm}-\max\{\mathbb{P}(Y \prec y|X=x,C=c),\\
&\hspace{3cm}\mathbb{P}(Y \prec y^l|X=x^*,C=c)\Big\},
\end{aligned}
\end{equation}
\begin{equation}
\begin{aligned}
&\gamma^I=\min\Big\{\mathbb{P}(Y \prec y|X=x',C=c),\\
&\hspace{3cm}\mathbb{P}(Y \prec y^u|X=x^*,C=c)\}\\
&\hspace{1cm}-\max\{\mathbb{P}(Y \prec y|X=x,C=c),\\
&\hspace{2cm}\mathbb{P}(Y \prec y^l|X=x^*,C=c),\rho(y;x',x,c)\Big\},
\end{aligned}
\end{equation}

\begin{equation}
\begin{aligned}
&\delta=\mathbb{P}(Y \prec y^u|X=x^*,C=c)-\mathbb{P}(Y \prec y^l|X=x^*,C=c).
\end{aligned}
\end{equation}
\vspace{0.1cm}

\noindent {\normalfont (B).}  If $\mathbb{P}(Y \prec y^u|X=x^*,C=c)=\mathbb{P}(Y \prec y^l|X=x^*,C=c)$, then 
\begin{equation}
\begin{aligned}
&\text{\normalfont T-PNS}(y;x',x,{\cal E}',c)=\mathbb{I}\Big(\mathbb{P}(Y \prec y|X=x',C=c)\leq\\
&\hspace{0cm}\mathbb{P}(Y \prec y^l|X=x^*,C=c) < \mathbb{P}(Y \prec y|X=x,C=c)\Big),
\end{aligned}
\end{equation}
\begin{equation}
\begin{aligned}
&\text{\normalfont ND-PNS}(y;x',x,{\cal E}',c)=\mathbb{I}\Big(\mathbb{P}(Y \prec y|X=x',C=c)\leq\\
&\hspace{0.5cm}\mathbb{P}(Y \prec y^l|X=x^*,C=c) < \mathbb{P}(Y \prec y|X=x,C=c),\\
&\hspace{1.5cm}\rho(y;x',x,c)\leq \mathbb{P}(Y \prec y^l|X=x^*,C=c)\Big),
\end{aligned}
\end{equation}
\begin{equation}
\begin{aligned}
&\text{\normalfont NI-PNS}(y;x',x,{\cal E}',c)=\mathbb{I}\Big(\mathbb{P}(Y \prec y|X=x',C=c)\leq\\
&\hspace{0.5cm}\mathbb{P}(Y \prec y^l|X=x^*,C=c) < \mathbb{P}(Y \prec y|X=x,C=c),\\
&\hspace{1.5cm} \mathbb{P}(Y \prec y^l|X=x^*,C=c)<\rho(y;x',x,c)\Big).
\end{aligned}
\end{equation}
}

\begin{proof}
Let $\tilde{u}_{x*,y^u}=\sup\{\tilde{u} \in \Omega_{\tilde{U}}; Y_{x^*}(\tilde{u}) \prec y^u\}$ and $\tilde{u}_{x^*,y^l}=\sup\{\tilde{u} \in \Omega_{\tilde{U}}; Y_{x^*}(\tilde{u}) \prec y^l\}$ given $C=c$.

\noindent {\normalfont (A).} 
From Assumptions 1, 3, and 5, if $\mathbb{P}(Y \prec y^u|X=x^*,C=c)\ne\mathbb{P}(Y \prec y^l|X=x^*,C=c)$,
then we have
\begin{equation}
\begin{aligned}
&\text{\normalfont T-PNS}(y;x',x,{\cal E}',c)\\
&=\mathbb{P}(Y_{x'} \prec y \preceq Y_{x}|X=x^*,Y\in {\cal I}_Y,C=c)\\
&={\mathbb{P}(Y_{x'} \prec y \preceq Y_{x},Y_{x^*}\in {\cal I}_Y|X=x^*,C=c)}\\
&\hspace{2.5cm}\Big/{\mathbb{P}(Y_{x^*}\in {\cal I}_Y|X=x^*,C=c)}\\
&={\mathbb{P}_{\tilde{U}}(\tilde{u}_{x',y} \preceq \tilde{U} \prec \tilde{u}_{x,y},\tilde{u}_{x*,y^l} \preceq \tilde{U} \prec \tilde{u}_{x*,y^u}|X=x^*,C=c)}\\
&\hspace{2.5cm}\Big/{\mathbb{P}(\tilde{u}_{x*,y^l} \preceq \tilde{U} \prec \tilde{u}_{x*,y^u}|X=x^*,C=c)}\\
&=\max\left\{{{\gamma}^T}/{\delta},0\right\},
\end{aligned}
\end{equation}
where $\gamma^T=\min\{\mathbb{P}(Y \prec y|X=x',C=c),\mathbb{P}(Y \prec y^u|X=x^*,C=c)\}-\max\{\mathbb{P}(Y \prec y|X=x,C=c),\mathbb{P}(Y \prec y^l|X=x^*,C=c)\}$ and $\delta=\mathbb{P}(Y \prec y^u|X=x^*,C=c)-\mathbb{P}(Y \prec y^l|X=x^*,C=c)$ for each $x',x \in \Omega_X$, $y \in \Omega_Y$, and $c \in \Omega_C$, and
\begin{equation}
\begin{aligned}
&\text{\normalfont ND-PNS}(y;x',x,{\cal E}',c)\\
&=\mathbb{P}(Y_{x'} \prec y \preceq Y_{x},  Y_{x',M_{x}} \prec y|X=x^*,Y\in {\cal I}_Y,C=c)\\
&={\mathbb{P}(Y_{x'} \prec y \preceq Y_{x},  Y_{x',M_{x}} \prec y,Y_{x^*}\in {\cal I}_Y|X=x^*,C=c)}\\
&\hspace{2.5cm}\Big/{\mathbb{P}(Y_{x^*}\in {\cal I}_Y|X=x^*,C=c)}\\
&=\mathbb{P}_{\tilde{U}}(\tilde{u}_{x',y} \preceq \tilde{U} \prec \tilde{u}_{x,y},  \tilde{u}_{x',M_x,y} \prec \tilde{U},\\
&\hspace{2cm}\tilde{u}_{x*,y^l} \preceq \tilde{U} \prec \tilde{u}_{x*,y^u}|X=x^*,C=c)\\
&\hspace{2.5cm}\Big/{\mathbb{P}(\tilde{u}_{x*,y^l} \preceq \tilde{U} \prec \tilde{u}_{x*,y^u}|X=x^*,C=c)}\\
&=\max\left\{{{\gamma}^D}/{\delta},0\right\},
\end{aligned}
\end{equation}
where ${\gamma}^D=\min\{\mathbb{P}(Y \prec y|X=x',C=c),\mathbb{P}(Y \prec y^u|X=x^*,C=c),\rho(y;x',x,c)\}-\max\{\mathbb{P}(Y \prec y|X=x,C=c),\mathbb{P}(Y \prec y^l|X=x^*,C=c)\}$ and $\delta=\mathbb{P}(Y \prec y^u|X=x^*,C=c)-\mathbb{P}(Y \prec y^l|X=x^*,C=c)$ for each $x',x \in \Omega_X$, $y \in \Omega_Y$, and $c \in \Omega_C$,
and
\begin{equation}
\begin{aligned}
&\text{\normalfont NI-PNS}(y;x',x,{\cal E}',c)\\
&=\mathbb{P}(Y_{x'} \prec y \preceq Y_{x}, y\preceq  Y_{x',M_{x}}|X=x^*,Y\in {\cal I}_Y,C=c)\\
&={\mathbb{P}(Y_{x'} \prec y \preceq Y_{x},  Y_{x',M_{x}} \prec y,Y_{x^*}\in {\cal I}_Y|X=x^*,C=c)}\\
&\hspace{2.5cm}\Big/{\mathbb{P}(Y_{x^*}\in {\cal I}_Y|X=x^*,C=c)}\\
&=\mathbb{P}_{\tilde{U}}(\tilde{u}_{x',y} \preceq \tilde{U} \prec \tilde{u}_{x,y}, \tilde{U} \preceq \tilde{u}_{x',M_x,y},\\
&\hspace{2cm}\tilde{u}_{x*,y^l} \preceq \tilde{U} \prec \tilde{u}_{x*,y^u}|X=x^*,C=c)\\
&\hspace{2.5cm}\Big/{\mathbb{P}(\tilde{u}_{x*,y^l} \preceq \tilde{U} \prec \tilde{u}_{x*,y^u}|X=x^*,C=c)}\\
&=\max\left\{{{\gamma}^I}/{\delta},0\right\},
\end{aligned}
\end{equation}
where $\gamma^I=\min\{\mathbb{P}(Y \prec y|X=x',C=c),\mathbb{P}(Y \prec y^u|X=x^*,C=c)\}-\max\{\mathbb{P}(Y \prec y|X=x,C=c),\mathbb{P}(Y \prec y^l|X=x^*,C=c),\rho(y;x',x,c)\}$ and $\delta=\mathbb{P}(Y \prec y^u|X=x^*,C=c)-\mathbb{P}(Y \prec y^l|X=x^*,C=c)$
for each $x',x \in \Omega_X$, $y \in \Omega_Y$, and $c \in \Omega_C$.\\

\noindent {\normalfont (B).} 
If $\mathbb{P}(Y \prec y^u|X=x^*,C=c)\ne\mathbb{P}(Y \prec y^l|X=x^*,C=c)$, then
\begin{equation}
\begin{aligned}
&\text{\normalfont T-PNS}(y;x',x,{\cal E}',c)\\
&=\mathbb{P}(Y_{x'} \prec y \preceq Y_{x}|X=x^*,Y\in {\cal I}_Y,C=c)\\
&=\mathbb{P}_{\tilde{U}}(\tilde{u}_{x',y} \preceq \tilde{U} \prec \tilde{u}_{x,y}|\tilde{U}=\tilde{u}_{x*,y^l},\\
&\hspace{2cm}X=x^*,Y\in {\cal I}_Y,C=c)\\
&=\mathbb{I}\Big(\mathbb{P}(Y \prec y|X=x',C=c)\leq\\
&\hspace{1cm}\mathbb{P}(Y \prec y^l|X=x^*,C=c)\\
&\hspace{2cm} < \mathbb{P}(Y \prec y|X=x,C=c)\Big),
\end{aligned}
\end{equation}
\begin{equation}
\begin{aligned}
&\text{\normalfont ND-PNS}(y;x',x,{\cal E}',c)\\
&=\mathbb{P}(Y_{x'} \prec y \preceq Y_{x},  Y_{x',M_{x}} \prec y|X=x^*,Y\in {\cal I}_Y,C=c)\\
&=\mathbb{P}_{\tilde{U}}(\tilde{u}_{x',y} \preceq \tilde{U} \prec \tilde{u}_{x,y},  \tilde{u}_{x',M_x,y} \prec \tilde{U}|\tilde{U}=\tilde{u}_{x*,y^l},\\
&\hspace{2cm}X=x^*,Y\in {\cal I}_Y,C=c)\\
&=\mathbb{I}\Big(\mathbb{P}(Y \prec y|X=x',C=c)\leq\\
&\hspace{0.5cm}\mathbb{P}(Y \prec y^l|X=x^*,C=c) < \mathbb{P}(Y \prec y|X=x,C=c),\\
&\hspace{1cm}\rho(y;x',x,c)\leq \mathbb{P}(Y \prec y^l|X=x^*,C=c)\Big)
\end{aligned}
\end{equation}
and
\begin{equation}
\begin{aligned}
&\text{\normalfont NI-PNS}(y;x',x,{\cal E}',c)\\
&=\mathbb{P}(Y_{x'} \prec y \preceq Y_{x},  Y_{x',M_{x}} \prec y|X=x^*,Y\in {\cal I}_Y,C=c)\\
&=\mathbb{P}_{\tilde{U}}(\tilde{u}_{x',y} \preceq \tilde{U} \prec \tilde{u}_{x,y}, \tilde{U} \prec \tilde{u}_{x',M_x,y}|\tilde{U}=\tilde{u}_{x*,y^l},\\
&\hspace{2cm}X=x^*,Y\in {\cal I}_Y,C=c)\\
&=\mathbb{I}\Big(\mathbb{P}(Y \prec y|X=x',C=c)\leq\\
&\hspace{0.5cm}\mathbb{P}(Y \prec y^l|X=x^*,C=c) < \mathbb{P}(Y \prec y|X=x,C=c),\\
&\hspace{1cm} \mathbb{P}(Y \prec y^l|X=x^*,C=c)<\rho(y;x',x,c)\Big)
\end{aligned}
\end{equation}
for each $x',x \in \Omega_X$, $y \in \Omega_Y$, and $c \in \Omega_C$.
\end{proof}\vspace{0.2cm}

Finally, we formally prove the two statements described on page 4:
\begin{center}
(i) ``Assumptions 4 and 4’ are equivalent under Assumption 2,"
\end{center}
and
\begin{center}
(ii) ``Assumptions 5 and 5’ are equivalent under Assumption 3,"    
\end{center}
(a straightforward extension of Theorem 4.1 in \citep{Kawakami2024}).
The proofs also follow the proof of Theorem 4.1 in \citep{Kawakami2024}).
\vspace{0.2cm}

\noindent {\bf Proof of Statement (i).}
We prove the statement (i).

\begin{proof}
We show the proof of equivalence of Assumptions 4 and 4’ under Assumption 2. \vspace{0.2cm}

\noindent (Assumption 4 $\Rightarrow$ Assumption 4'.)
For any $c \in \Omega_C$, from Assumption 3, if we have the negation of Assumption 4'
\begin{center}
there exists a set ${\cal U} \subset \Omega_{U_Y}$ such that $0<\mathbb{P}({\cal U})<1$, and 
\begin{equation}
\begin{aligned}
&f_Y(x,m,c,u')\succeq y\succ f_Y(x,m,c,u)\\
& \land f_Y(x',m',c,u')\prec y \preceq f_Y(x',m',c,u)
\end{aligned}
\end{equation} for some $x, x' \in \Omega_X$, $m, m' \in \Omega_M$ and $y \in \Omega_Y$ and for any $u',u \in {\cal U}$ such that $u'\preceq u$,
\end{center}
then we have
\begin{center}
$f_Y(x,m,c,u')\succeq y \succ f_Y(x',m',c,u')$ and $f_Y(x,m,c,u)\prec y \preceq f_Y(x',m',c,u)$ for some $x, x' \in \Omega_X$, $m, m' \in \Omega_M$ and $y \in \Omega_Y$ and\\ for any $u', u \in {\cal U}$ such that $u' \preceq u$,
\end{center}
and we also have 
\begin{center}
$f_Y(x,m,c,u)\succeq y \succ f_Y(x',m',c,u)$ and $f_Y(x,m,c,u)\prec y \preceq f_Y(x',m',c,u)$ for some $x, x' \in \Omega_X$, $m, m' \in \Omega_M$, and $y \in \Omega_Y$ and  for any $u \in {\cal U}$.
\end{center}
This implies the negation of Assumption 4' $\mathbb{P}(Y_{x,m}\prec y \preceq Y_{x',m'}|C=c)\ne 0$ and $\mathbb{P}(Y_{x',m'}\prec y \preceq Y_{x,m}|C=c)\ne 0$ for some $x, x' \in \Omega_Y$, $m, m' \in \Omega_M$ and $y \in \Omega_Y$ since $f_Y(x,m,c,u)\succeq y \succ f_Y(x',m',c,u) \Leftrightarrow Y_{x,m}(c,u)\succeq y \succ Y_{x',m'}(c,u)$ and $f_Y(x,m,c,u)\prec y \preceq f_Y(x',m',c,u) \Leftrightarrow Y_{x',m'}(c,u)\succeq y \succ Y_{x,m}(c,u)$. \vspace{0.2cm}

\noindent (Assumption 4' $\Rightarrow$ Assumption 4.)
For any $c \in \Omega_C$, we denote $u_{sup}=\sup\{u:f_Y(x,m,c,u)\preceq y\}$. 
We consider the situations ``the function $f_Y(x,m,c,U_Y)$ is monotonic increasing on $U_Y$'' and ``the function $f_Y(x,m,c,U_Y)$ is monotonic decreasing on $U_Y$'', separately. \vspace{0.2cm}

\noindent {\bf (1).}
If the function $f_Y(x,m,c,U_Y)$ is {\bf monotonic increasing} on $U_Y$
for all $x \in \Omega_X$ almost surely w.r.t. $\mathbb{P}_{U_Y}$, we have
\begin{equation}
\begin{aligned}
&f_Y(x,m,c,u_{sup}) \preceq f_Y(x,m,c,u)\\
&\text{ and }f_Y(x',m',c,u_{sup}) \preceq f_Y(x',m',c,u)
\end{aligned}
\end{equation}
for $\mathbb{P}_{U_Y}$-almost every $u \in \Omega_{U_Y}$ such that $u\succeq u_{sup}$.
We have the following statements:
\begin{enumerate}
    \item Supposing $f_Y(x,m,c,u_{sup})\succ f_Y(x',m',c,u_{sup})$, 
we have $y= f_Y(x,m,c,u_{sup}) \succ f_Y(x',m',c,u_{sup})\succeq f_Y(x',m',c,u)=Y_{x',m'}(c,u)$ for $\mathbb{P}_{U_Y}$-almost every $u \in \Omega_{U_Y}$ such that $f_Y(x,m,c,u)\prec y$.
It means $Y_{x,m}(c,u)\prec y \Rightarrow Y_{x',m'}(c,u) \prec y$ for $\mathbb{P}_{U_Y}$-almost every $u \in \Omega_{U_Y}$ and $\mathbb{P}(Y_{x,m}\prec y \preceq Y_{x',m'}|C=c)=0$.
\item Supposing $f_Y(x,m,c,u_{sup})\preceq f_Y(x',m',c,u_{sup})$, we have $f_Y(x',m',c,u) \succeq f_Y(x',m',c,u_{sup}) \succeq f_Y(x,m,c,u_{sup}) =y$ for $\mathbb{P}_{U_Y}$-almost every $u \in \Omega_{U_Y}$ such that $f_Y(x,m,c,u)\succeq y$.
It means $Y_{x,m}(c,u)\succeq y \Rightarrow Y_{x',m'}(c,u) \succeq y$ for $\mathbb{P}_{U_Y}$-almost every $u \in \Omega_{U_Y}$ and  $\mathbb{P}(Y_{x',m'}\prec y \preceq Y_{x,m}|C=c)=0$.
\end{enumerate}
Then, these results imply Assumption 4'. \vspace{0.2cm}

\noindent {\bf (2).}
If the function $f_Y(x,m,c,U_Y)$ is {\bf monotonic decreasing} on $U_Y$
for all $x \in \Omega_X$ almost surely w.r.t. $\mathbb{P}_{U_Y}$, we have
\begin{equation}
\begin{aligned}
&f_Y(x,m,c,u_{sup}) \succeq f_Y(x,m,c,u)\\
&\text{ and }f_Y(x',m',c,u_{sup}) \succeq f_Y(x',m',c,u)
\end{aligned}
\end{equation}
for $\mathbb{P}_{U_Y}$-almost every $u \in \Omega_{U_Y}$ such that $u\succeq u_{sup}$.
We have the following statements:
\begin{enumerate}
    \item Supposing $f_Y(x,m,c,u_{sup})\preceq f_Y(x',m',c,u_{sup})$, 
we have $y= f_Y(x,m,c,u_{sup}) \preceq f_Y(x',m',c,u_{sup})\preceq f_Y(x',m',c,u)=Y_{x',m'}(c,u)$ for $\mathbb{P}_{U_Y}$-almost every $u \in \Omega_{U_Y}$ such that $f_Y(x,m,c,u)\succeq y$.
It means $Y_{x,m}(c,u)\succeq y \Rightarrow Y_{x',m'}(u) \succeq y$ for $\mathbb{P}_{U_Y}$-almost every $u \in \Omega_{U_Y}$ and  $\mathbb{P}(Y_{x',m'}\prec y \preceq Y_{x,m}|C=c)=0$.
\item Supposing $f_Y(x,m,c,u_{sup})\succ f_Y(x',m',c,u_{sup})$, we have $f_Y(x',m',c,u) \prec f_Y(x',m',c,u_{sup}) \preceq f_Y(x,m,c,u_{sup}) =y$ for $\mathbb{P}_{U_Y}$-almost every $u \in \Omega_{U_Y}$ such that $f_Y(x,m,c,u)\prec y$.
It means $Y_{x,m}(c,u)\prec y \Rightarrow Y_{x',m'}(c,u) \prec y$ for $\mathbb{P}_{U_Y}$-almost every $u \in \Omega_{U_Y}$ and  $\mathbb{P}(Y_{x,m}\prec y \preceq Y_{x',m'}|C=c)=0$.
\end{enumerate}
Then, these results imply Assumption 4'.
In conclusion, Assumption 4' implies Assumption 4 under Assumption 2.

\end{proof}

\vspace{0.2cm}
\noindent {\bf Proof of Statement (ii).}
We prove the statement (ii).

\begin{proof}
We show the proof of equivalence of Assumptions 5 and 5’ under Assumption 3. \vspace{0.2cm}

\noindent (Assumption 5 $\Rightarrow$ Assumption 5'.)
For any $c \in \Omega_C$, from Assumption 3, if we have the negation of Assumption 5'
\begin{center}
there exists a set ${\cal U} \subset \Omega_{\tilde{U}}$ such that $0<\mathbb{P}({\cal U})<1$, and 
\begin{equation}
\begin{aligned}
&(f_Y \circ f_M)(x',x,c,\tilde{u}_0)\succeq y\succ (f_Y \circ f_M)(x',x,c,\tilde{u}_1)\\
&\land (f_Y \circ f_M)(x''',x'',c,\tilde{u}_0)\prec y \preceq (f_Y \circ f_M)(x''',x'',c,\tilde{u}_1)        
\end{aligned}
\end{equation} for some $x,x', x'',x''' \in \Omega_X$ and $y \in \Omega_Y$ and for any $\tilde{u}_0,\tilde{u}_1 \in {\cal U}$ such that $\tilde{u}_0\preceq \tilde{u}_1$,
\end{center}
then we have
\begin{center}
$(f_Y \circ f_M)(x',x,c,\tilde{u}_0)\succeq y \succ (f_Y \circ f_M)(x''',x'',\tilde{u}_0)$ and $(f_Y \circ f_M)(x',x,c,\tilde{u}_1)\prec y \preceq (f_Y \circ f_M)(x''',x'',c,\tilde{u}_1)$ for some $x,x', x'',x''' \in \Omega_X$ and $y \in \Omega_Y$ and\\ for any $\tilde{u}_0, \tilde{u}_1 \in {\cal U}$ such that $\tilde{u}_0 \preceq \tilde{u}_1$,
\end{center}
and we also have 
\begin{center}
$(f_Y \circ f_M)(x',x,c,\tilde{u})\succeq y \succ (f_Y \circ f_M)(x''',x'',c,\tilde{u})$ and $(f_Y \circ f_M)(x',x,c,\tilde{u})\prec y \preceq (f_Y \circ f_M)(x''',x'',c,\tilde{u})$ for some $x,x', x'',x''' \in \Omega_X$ and $y \in \Omega_Y$ and  for any $\tilde{u} \in {\cal U}$.
\end{center}
This implies the negation of Assumption 5 $\mathbb{P}(Y_{x',M_{x}}\prec y \preceq Y_{x''',M_{x''}}|C=c)\ne 0$ and $\mathbb{P}(Y_{x''',M_{x''}}\prec y \preceq Y_{x',M_{x}}|C=c)\ne 0$ for some $x,x', x'',x'''  \in \Omega_Y$ and $y \in \Omega_Y$ since $(f_Y \circ f_M)(x',x,c,\tilde{u})\succeq y \succ (f_Y \circ f_M)(x''',x'',c,\tilde{u}) \Leftrightarrow Y_{x',M_{x}}(c,\tilde{u})\succeq y \succ Y_{x''',M_{x''}}(c,\tilde{u})$ and $(f_Y \circ f_M)(x',x,c,\tilde{u})\prec y \preceq (f_Y \circ f_M)(x''',x'',c,\tilde{u}) \Leftrightarrow Y_{x''',M_{x''}}(c,\tilde{u})\succeq y \succ Y_{x',M_{x}}(c,\tilde{u})$. \vspace{0.2cm}

\noindent (Assumption 5' $\Rightarrow$ Assumption 5.)
For any $c \in \Omega_C$, we denote $\tilde{u}_{sup}=\sup\{\tilde{u}:(f_Y \circ f_M)(x',x,c,\tilde{u})\preceq y\}$. 
We consider the situations ``the function $(f_Y \circ f_M)(x',x,c,\tilde{U})$ is monotonic increasing on $\tilde{U}$'' and ``the function $(f_Y \circ f_M)(x',x,c,\tilde{U})$ is monotonic decreasing on $\tilde{U}$'', separately. \vspace{0.2cm}

\noindent {\bf (1).}
If the function $(f_Y \circ f_M)(x',x,c,\tilde{U})$ is {\bf monotonic increasing} on $\tilde{U}$
for all $x \in \Omega_X$ almost surely w.r.t. $\mathbb{P}_{\tilde{U}}$, we have
\begin{equation}
\begin{aligned}
&(f_Y \circ f_M)(x',x,c,\tilde{u}_{sup}) \preceq (f_Y \circ f_M)(x',x,c,\tilde{u})\\
&\text{ and }(f_Y \circ f_M)(x''',x'',c,\tilde{u}_{sup}) \preceq (f_Y \circ f_M)(x''',x'',c,\tilde{u})
\end{aligned}
\end{equation}
for $\mathbb{P}_{\tilde{U}}$-almost every $\tilde{u} \in \Omega_{\tilde{U}}$ such that $\tilde{u}\succeq \tilde{u}_{sup}$.
We have the following statements:
\begin{enumerate}
\item Supposing $(f_Y \circ f_M)(x',x,c,\tilde{u}_{sup})\succ (f_Y \circ f_M)(x''',x'',c,\tilde{u}_{sup})$, 
we have $y= (f_Y \circ f_M)(x',x,c,\tilde{u}_{sup}) \succ (f_Y \circ f_M)(x''',x'',c,\tilde{u}_{sup})\succeq (f_Y \circ f_M)(x''',x'',c,\tilde{u})=Y_{x''',M_{x''}}(c,\tilde{u})$ for $\mathbb{P}_{\tilde{U}}$-almost every $\tilde{u} \in \Omega_{\tilde{U}}$ such that $(f_Y \circ f_M)(x',x,c,\tilde{u})\prec y$.
It means $Y_{x',M_{x}}(c,\tilde{u})\prec y \Rightarrow Y_{x''',M_{x''}}(c,\tilde{u}) \prec y$ for $\mathbb{P}_{\tilde{U}}$-almost every $\tilde{u} \in \Omega_{\tilde{U}}$ and  $\mathbb{P}(Y_{x',M_{x}}\prec y \preceq Y_{x''',M_{x''}}|C=c)=0$.
\item Supposing $(f_Y \circ f_M)(x',x,c,\tilde{u}_{sup})\preceq (f_Y \circ f_M)(x''',x'',c,\tilde{u}_{sup})$, we have $(f_Y \circ f_M)(x''',x'',c,\tilde{u}) \succeq (f_Y \circ f_M)(x''',x'',c,\tilde{u}_{sup}) \succeq (f_Y \circ f_M)(x',x,c,\tilde{u}_{sup}) =y$ for $\mathbb{P}_{\tilde{U}}$-almost every $\tilde{u} \in \Omega_{\tilde{U}}$ such that $(f_Y \circ f_M)(x',x,c,\tilde{u})\succeq y$.
It means $Y_{x',M_{x}}(c,\tilde{u})\succeq y \Rightarrow Y_{x''',M_{x''}}(c,\tilde{u}) \succeq y$ for $\mathbb{P}_{\tilde{U}}$-almost every $\tilde{u} \in \Omega_{\tilde{U}}$ and  $\mathbb{P}(Y_{x''',M_{x''}}\prec y \preceq Y_{x',M_{x}}|C=c)=0$.
\end{enumerate}
Then, these results imply Assumption 5'. \vspace{0.2cm}

\noindent {\bf (2).}
If the function $(f_Y \circ f_M)(x',x,c,\tilde{U})$ is {\bf monotonic decreasing} on $\tilde{U}$
for all $x \in \Omega_X$ almost surely w.r.t. $\mathbb{P}_{\tilde{U}}$, we have
\begin{equation}
\begin{aligned}
&(f_Y \circ f_M)(x',x,c,\tilde{u}_{sup}) \succeq (f_Y \circ f_M)(x',x,c,\tilde{u})\\
&\text{ and }(f_Y \circ f_M)(x''',x'',c,\tilde{u}_{sup}) \succeq (f_Y \circ f_M)(x''',x'',c,\tilde{u})
\end{aligned}
\end{equation}
for $\mathbb{P}_{\tilde{U}}$-almost every $\tilde{u} \in \Omega_{\tilde{U}}$ such that $\tilde{u}\succeq \tilde{u}_{sup}$.
We have the following statements:
\begin{enumerate}
    \item Supposing $(f_Y \circ f_M)(x',x,c,\tilde{u}_{sup})\preceq (f_Y \circ f_M)(x''',x'',c,\tilde{u}_{sup})$, 
we have $y= (f_Y \circ f_M)(x',x,c,\tilde{u}_{sup}) \preceq (f_Y \circ f_M)(x''',x'',c,\tilde{u}_{sup})\preceq (f_Y \circ f_M)(x''',x'',c,\tilde{u})=Y_{x''',M_{x''}}(c,\tilde{u})$ for $\mathbb{P}_{\tilde{U}}$-almost every $\tilde{u} \in \Omega_{\tilde{U}}$ such that $(f_Y \circ f_M)(x',x,c,\tilde{u})\succeq y$.
It means $Y_{x',M_{x}}(c,\tilde{u})\succeq y \Rightarrow Y_{x''',M_{x''}}(\tilde{u}) \succeq y$ for $\mathbb{P}_{\tilde{U}}$-almost every $\tilde{u} \in \Omega_{\tilde{U}}$ and  $\mathbb{P}(Y_{x''',M_{x''}}\prec y \preceq Y_{x',M_{x}}|C=c)=0$.
\item Supposing $(f_Y \circ f_M)(x',x,c,\tilde{u}_{sup})\succ (f_Y \circ f_M)(x''',x'',c,\tilde{u}_{sup})$, we have $(f_Y \circ f_M)(x''',x'',c,\tilde{u}) \prec (f_Y \circ f_M)(x''',x'',c,\tilde{u}_{sup}) \preceq (f_Y \circ f_M)(x',x,c,\tilde{u}_{sup}) =y$ for $\mathbb{P}_{\tilde{U}}$-almost every $\tilde{u} \in \Omega_{\tilde{U}}$ such that $(f_Y \circ f_M)(x',x,c,\tilde{u})\prec y$.
It means $Y_{x',M_{x}}(c,\tilde{u})\prec y \Rightarrow Y_{x''',M_{x''}}(c,\tilde{u}) \prec y$ for $\mathbb{P}_{\tilde{U}}$-almost every $\tilde{u} \in \Omega_{\tilde{U}}$ and  $\mathbb{P}(Y_{x',M_{x}}\prec y \preceq Y_{x''',M_{x''}}|C=c)=0$.
\end{enumerate}
Then, these resluts imply Assumption 5'.
In conclusion, Assumption 5' implies Assumption 5 under Assumption 3.

\end{proof}

\section*{Appendix C: Additional Theorem}

In this section, we provide the identification theorems of T-PNS, ND-PNS, and NI-PNS with evidence ${\cal E}''=(X=x^*, M\in {\cal I}_M, Y\in {\cal I}_Y)$, which differs from ${\cal E}$ or ${\cal E}'$, with additional assumption, where ${\cal I}_M$ is a half-open interval $[m^l,m^u)$ w.r.t. the total order on $\Omega_M$.
We make the following additional assumption:\\

\noindent{\bf Assumption A1.}
(Monotonicity over $f_M$)
{\it
The function $f_M(x,c,U_M)$ is either monotonic increasing on $\tilde{U}$ for all $x \in \Omega_X$ and $c \in \Omega_C$, or monotonic decreasing on $\tilde{U}$ for all $x \in \Omega_X$ and $c \in \Omega_C$ almost surely w.r.t. $\mathbb{P}_{\tilde{U}}$.
}\\

\noindent Supposing $\Omega_{\tilde{U}}=\Omega_{U_Y}\times \Omega_{U_M}$ is totally order by the lexicographical order \citep{Harzheim2005} with $(U_M,U_Y)$ and the function $f_M(x,c,U_M)$ is either monotonic increasing on $U_M$ for all $x \in \Omega_X$ and $c \in \Omega_C$, or monotonic decreasing on $U_M$ for all $x \in \Omega_X$ and $c \in \Omega_C$ almost surely w.r.t. $\mathbb{P}_{U_M}$, then Assumption 6 holds.
SCMs ${\cal M}^L$ or ${\cal M}^N$ do not satisfies this assumption.\\

\noindent{\bf Theorem A1.}
{\it
Let ${\cal I}_Y$ and ${\cal I}_M$ be half-open intervals $[y^l,y^u)$ and $[m^l,m^u)$ in evidence ${\cal E}'$.
Under SCM ${\cal M}$, and Assumptions 1, 3, 5, and A1, for each $x',x \in \Omega_X$, $y \in \Omega_Y$, and $c \in \Omega_C$, we have

\noindent {\normalfont (A).} If $\mathbb{P}(Y \prec y^u,M \prec m^u|X=x^*,C=c)\ne\mathbb{P}(Y \prec y^l, M \prec m^l|X=x^*,C=c)$, then
\begin{equation}
\begin{gathered}
\text{\normalfont T-PNS}(y;x',x,{\cal E}'',c)=\max\left\{{\gamma^T}/{\delta},0\right\},\\
\text{\normalfont ND-PNS}(y;x',x,{\cal E}'',c)=\max\left\{{\gamma^D}/{\delta},0\right\},\\
\text{\normalfont NI-PNS}(y;x',x,{\cal E}'',c)=\max\left\{{\gamma^I}/{\delta},0\right\},
\end{gathered}
\end{equation}
where 
\begin{equation}
\begin{aligned}
&\gamma^T=\min\Big\{\mathbb{P}(Y \prec y|X=x',C=c),\\
&\hspace{1.5cm}\mathbb{P}(Y \prec y^u,M \prec m^u|X=x^*,C=c)\}\\
&\hspace{1cm}-\max\{\mathbb{P}(Y \prec y|X=x,C=c),\\
&\hspace{2cm}\mathbb{P}(Y \prec y^l,M \prec m^l|X=x^*,C=c)\Big\},
\end{aligned}
\end{equation}
\begin{equation}
\begin{aligned}
&\gamma^D=\min\Big\{\mathbb{P}(Y \prec y|X=x',C=c),\\
&\hspace{1cm}\mathbb{P}(Y \prec y^u,M \prec m^u|X=x^*,C=c),\rho(y;x',x,c)\}\\
&\hspace{1cm}-\max\{\mathbb{P}(Y \prec y|X=x,C=c),\\
&\hspace{2cm}\mathbb{P}(Y \prec y^l,M \prec m^l|X=x^*,C=c)\Big\},
\end{aligned}
\end{equation}
\begin{equation}
\begin{aligned}
&\gamma^I=\min\Big\{\mathbb{P}(Y \prec y|X=x',C=c),\\
&\hspace{2cm}\mathbb{P}(Y \prec y^u,M \prec m^u|X=x^*,C=c)\}\\
&\hspace{1cm}-\max\{\mathbb{P}(Y \prec y|X=x,C=c),\\
&\hspace{1cm}\mathbb{P}(Y \prec y^l,M \prec m^l|X=x^*,C=c),\rho(y;x',x,c)\Big\},
\end{aligned}
\end{equation}
and
\begin{equation}
\begin{aligned}
&\delta=\mathbb{P}(Y \prec y^u,M \prec m^u|X=x^*,C=c)\\
&\hspace{1cm}-\mathbb{P}(Y \prec y^l,M \prec m^l|X=x^*,C=c).
\end{aligned}
\end{equation}

\noindent {\normalfont (B).}  If $\mathbb{P}(Y \prec y^u,M \prec m^u|X=x^*,C=c)=\mathbb{P}(Y \prec y^l,M \prec m^l|X=x^*,C=c)$, then
\begin{equation}
\begin{aligned}
&\text{\normalfont T-PNS}(y;x',x,{\cal E}'',c)=\mathbb{I}\Big(\mathbb{P}(Y \prec y|X=x',C=c)\leq\\
&\hspace{1cm}\mathbb{P}(Y \prec y^l,M \prec m^l|X=x^*,C=c) \\
&\hspace{2cm}< \mathbb{P}(Y \prec y|X=x,C=c)\Big),
\end{aligned}
\end{equation}
\begin{equation}
\begin{aligned}
&\text{\normalfont ND-PNS}(y;x',x,{\cal E}'',c)=\mathbb{I}\Big(\mathbb{P}(Y \prec y|X=x',C=c)\leq\\
&\hspace{0.5cm}\mathbb{P}(Y \prec y^l,M \prec m^l|X=x^*,C=c) \\
&\hspace{1cm}< \mathbb{P}(Y \prec y|X=x,C=c),\\
&\hspace{0.5cm}\rho(y;x',x,c)\leq \mathbb{P}(Y \prec y^l,M \prec m^l|X=x^*,C=c)\Big),
\end{aligned}
\end{equation}
\begin{equation}
\begin{aligned}
&\text{\normalfont NI-PNS}(y;x',x,{\cal E}'',c)=\mathbb{I}\Big(\mathbb{P}(Y \prec y|X=x',C=c)\leq\\
&\hspace{1cm}\mathbb{P}(Y \prec y^l,M \prec m^l|X=x^*,C=c)\\
&\hspace{2cm} < \mathbb{P}(Y \prec y|X=x,C=c),\\
&\hspace{0.5cm} \mathbb{P}(Y \prec y^l,M \prec m^l|X=x^*,C=c)<\rho(y;x',x,c)\Big).
\end{aligned}
\end{equation}
}
\noindent When ${\cal I}_Y$ is an closed intervel $[y^l,y^u]$,
we have the identification results by changing ``$Y \prec y^u$" to ``$Y \preceq y^u$" in Theorem A1.
Also, when ${\cal I}_M$ is an closed intervel $[m^l,m^u]$,
we have the identification results by changing ``$M \prec m^u$" to ``$M \preceq m^u$" in Theorem A1.

\begin{proof}
Let $\tilde{u}_{x*,y^u,m^u}=\sup\{\tilde{u} \in \Omega_{\tilde{U}}; Y_{x^*}(\tilde{u}) \prec y^u,M_{x^*}(\tilde{u}) \prec m^u\}$ and  $\tilde{u}_{x^*,y^l,m^l}=\sup\{u \in \Omega_{\tilde{U}}; Y_{x^*}(\tilde{u}) \prec y^l,M_{x^*}(\tilde{u}) \prec m^l\}$ given $C=c$.

\noindent {\normalfont (A).} 
From Assumptions 1, 3, 5, and A1, when $\mathbb{P}(Y \prec y^u, M \prec m^u|X=x^*,C=c)\ne\mathbb{P}(Y \prec y^l, M \prec m^l|X=x^*,C=c)$,
then we have
\begin{equation}
\begin{aligned}
&\text{\normalfont T-PNS}(y;x',x,{\cal E}'',c)\\
&=\mathbb{P}(Y_{x'} \prec y \preceq Y_{x}|X=x^*,Y\in {\cal I}_Y,C=c)\\
&={\mathbb{P}(Y_{x'} \prec y \preceq Y_{x},M_{x^*}\in {\cal I}_M, Y_{x^*}\in {\cal I}_Y|X=x^*,C=c)}\\
&\hspace{2cm}\Big/{\mathbb{P}(M_{x^*}\in {\cal I}_M, Y_{x^*}\in {\cal I}_Y|X=x^*,C=c)}\\
&=\mathbb{P}_{\tilde{U}}(\tilde{u}_{x',y} \preceq \tilde{U} \prec \tilde{u}_{x,y},\\
&\hspace{2cm}\tilde{u}_{x*,y^l,m^l} \preceq \tilde{U} \prec \tilde{u}_{x*,y^u,m^u}|X=x^*,C=c)\\
&\hspace{2cm}\Big/{\mathbb{P}(\tilde{u}_{x*,y^l,m^l} \preceq \tilde{U} \prec \tilde{u}_{x*,y^u,m^u}|X=x^*,C=c)}\\
&=\max\left\{{{\gamma}^T}/{\delta},0\right\},
\end{aligned}
\end{equation}
where $\gamma^T=\min\{\mathbb{P}(Y \prec y|X=x',C=c),\mathbb{P}(Y \prec y^u, M \prec m^u|X=x^*,C=c)\}-\max\{\mathbb{P}(Y \prec y|X=x,C=c),\mathbb{P}(Y \prec y^l, M \prec m^l|X=x^*,C=c)\}$ and $\delta=\mathbb{P}(Y \prec y^u, M \prec m^u|X=x^*,C=c)-\mathbb{P}(Y \prec y^l, M \prec m^l|X=x^*,C=c)$ for each $x',x \in \Omega_X$, $y \in \Omega_Y$, and $c \in \Omega_C$, and
\begin{equation}
\begin{aligned}
&\text{\normalfont ND-PNS}(y;x',x,{\cal E}'',c)\\
&=\mathbb{P}(Y_{x'} \prec y \preceq Y_{x},  Y_{x',M_{x}} \prec y|X=x^*,Y\in {\cal I}_Y,C=c)\\
&=\mathbb{P}(Y_{x'} \prec y \preceq Y_{x},  Y_{x',M_{x}} \prec y,\\
&\hspace{2cm}M_{x^*}\in {\cal I}_M, Y_{x^*}\in {\cal I}_Y|X=x^*,C=c)\\
&\hspace{2cm}\Big/{\mathbb{P}(M_{x^*}\in {\cal I}_M, Y_{x^*}\in {\cal I}_Y|X=x^*,C=c)}\\
&=\mathbb{P}(Y_{x'} \prec y \preceq Y_{x},  Y_{x',M_{x}} \prec y,\\
&=\mathbb{P}_{\tilde{U}}(\tilde{u}_{x',y} \preceq \tilde{U} \prec \tilde{u}_{x,y},  u_{x',M_x,y} \prec \tilde{U},\\
&\hspace{2cm}\tilde{u}_{x*,y^l,m^l} \preceq \tilde{U} \prec \tilde{u}_{x*,y^u,m^u}|X=x^*,C=c)\\
&\hspace{1.5cm}\Big/{\mathbb{P}(\tilde{u}_{x*,y^l,m^l} \preceq \tilde{U} \prec \tilde{u}_{x*,y^u,m^u}|X=x^*,C=c)}\\
&=\max\left\{{{\gamma}^D}/{\delta},0\right\},
\end{aligned}
\end{equation}
where ${\gamma}^D=\min\{\mathbb{P}(Y \prec y|X=x',C=c),\mathbb{P}(Y \prec y^u, M \prec m^u|X=x^*,C=c),\rho(y;x',x,c)\}-\max\{\mathbb{P}(Y \prec y|X=x,C=c),\mathbb{P}(Y \prec y^l, M \prec m^l|X=x^*,C=c)\}$ and $\delta=\mathbb{P}(Y \prec y^u, M \prec m^u|X=x^*,C=c)-\mathbb{P}(Y \prec y^l, M \prec m^l|X=x^*,C=c)$ for each $x',x \in \Omega_X$, $y \in \Omega_Y$, and $c \in \Omega_C$,
and
\begin{equation}
\begin{aligned}
&\text{\normalfont NI-PNS}(y;x',x,{\cal E}'',c)\\
&=\mathbb{P}(Y_{x'} \prec y \preceq Y_{x}, y\preceq  Y_{x',M_{x}}|X=x^*,Y\in {\cal I}_Y,C=c)\\
&=\mathbb{P}(Y_{x'} \prec y \preceq Y_{x},  Y_{x',M_{x}} \prec y,\\
&\hspace{2cm}M_{x^*}\in {\cal I}_M, Y_{x^*}\in {\cal I}_Y|X=x^*,C=c)\\
&\hspace{2.5cm}\Big/{\mathbb{P}(M_{x^*}\in {\cal I}_M, Y_{x^*}\in {\cal I}_Y|X=x^*,C=c)}\\
&=\mathbb{P}_{\tilde{U}}(\tilde{u}_{x',y} \preceq \tilde{U} \prec \tilde{u}_{x,y}, \tilde{U} \preceq u_{x',M_x,y},\\
&\hspace{2cm}\tilde{u}_{x*,y^l,m^l} \preceq \tilde{U} \prec \tilde{u}_{x*,y^u,m^u}|X=x^*,C=c)\\
&\hspace{1.5cm}\Big/{\mathbb{P}(\tilde{u}_{x*,y^l,m^l} \preceq \tilde{U} \prec \tilde{u}_{x*,y^u,m^u}|X=x^*,C=c)}\\
&=\max\left\{{{\gamma}^I}/{\delta},0\right\},
\end{aligned}
\end{equation}
where $\gamma^I=\min\{\mathbb{P}(Y \prec y|X=x',C=c),\mathbb{P}(Y \prec y^u, M \prec m^u|X=x^*,C=c)\}-\max\{\mathbb{P}(Y \prec y|X=x,C=c),\mathbb{P}(Y \prec y^l, M \prec m^l|X=x^*,C=c),\rho(y;x',x,c)\}$ and $\delta=\mathbb{P}(Y \prec y^u, M \prec m^u|X=x^*,C=c)-\mathbb{P}(Y \prec y^l, M \prec m^l|X=x^*,C=c)$
for each $x',x \in \Omega_X$, $y \in \Omega_Y$, and $c \in \Omega_C$.

\noindent {\normalfont (B).} 
When $\mathbb{P}(Y \prec y^u, M \prec m^u|X=x^*,C=c)\ne\mathbb{P}(Y \prec y^l, M \prec m^l|X=x^*,C=c)$, then we have
\begin{equation}
\begin{aligned}
&\text{\normalfont T-PNS}(y;x',x,{\cal E}'',c)\\
&=\mathbb{P}(Y_{x'} \prec y \preceq Y_{x}|X=x^*,Y\in {\cal I}_Y,C=c)\\
&=\mathbb{P}_{\tilde{U}}(\tilde{u}_{x',y} \preceq \tilde{U} \prec \tilde{u}_{x,y}|\tilde{U}=\tilde{u}_{x*,y^l,m^l},\\
&\hspace{2cm}X=x^*,Y\in {\cal I}_Y,C=c)\\
&=\mathbb{I}\Big(\mathbb{P}(Y \prec y|X=x',C=c)\leq\\
&\hspace{1cm}\mathbb{P}(Y \prec y^l, M \prec m^l|X=x^*,C=c) \\
&\hspace{2cm}< \mathbb{P}(Y \prec y|X=x,C=c)\Big),
\end{aligned}
\end{equation}
\begin{equation}
\begin{aligned}
&\text{\normalfont ND-PNS}(y;x',x,{\cal E}'',c)\\
&=\mathbb{P}(Y_{x'} \prec y \preceq Y_{x},  Y_{x',M_{x}} \prec y|X=x^*,Y\in {\cal I}_Y,C=c)\\
&=\mathbb{P}_{\tilde{U}}(\tilde{u}_{x',y} \preceq \tilde{U} \prec \tilde{u}_{x,y},  u_{x',M_x,y} \prec \tilde{U}|\tilde{U}=\tilde{u}_{x*,y^l,m^l},\\
&\hspace{2cm}X=x^*,Y\in {\cal I}_Y,C=c)\\
&=\mathbb{I}\Big(\mathbb{P}(Y \prec y|X=x',C=c)\leq\\
&\hspace{1cm}\mathbb{P}(Y \prec y^l, M \prec m^l|X=x^*,C=c)\\
&\hspace{2cm} < \mathbb{P}(Y \prec y|X=x,C=c),\\
&\hspace{0.5cm}\rho(y;x',x,c)\leq \mathbb{P}(Y \prec y^l, M \prec m^l|X=x^*,C=c)\Big),
\end{aligned}
\end{equation}
and
\begin{equation}
\begin{aligned}
&\text{\normalfont NI-PNS}(y;x',x,{\cal E}'',c)\\
&=\mathbb{P}(Y_{x'} \prec y \preceq Y_{x},  Y_{x',M_{x}} \prec y|X=x^*,Y\in {\cal I}_Y,C=c)\\
&=\mathbb{P}_{\tilde{U}}(\tilde{u}_{x',y} \preceq \tilde{U} \prec \tilde{u}_{x,y},  \tilde{U} \prec u_{x',M_x,y}|\tilde{U}=\tilde{u}_{x*,y^l,m^l},\\
&\hspace{2cm}X=x^*,Y\in {\cal I}_Y,C=c)\\
&=\mathbb{I}\Big(\mathbb{P}(Y \prec y|X=x',C=c)\leq\\
&\hspace{1cm}\mathbb{P}(Y \prec y^l, M \prec m^l|X=x^*,C=c)\\
&\hspace{2cm} < \mathbb{P}(Y \prec y|X=x,C=c),\\
&\hspace{0cm} \mathbb{P}(Y \prec y^l, M \prec m^l|X=x^*,C=c)<\rho(y;x',x,c)\Big)
\end{aligned}
\end{equation}
for each $x',x \in \Omega_X$, $y \in \Omega_Y$, and $c \in \Omega_C$.
\end{proof}

\section*{Appendix D: Binary Treatment and Outcome}
In this section, we illustrate how the results in the body of the paper reduce to binary variables.
All measures in this section are identifiable by Theorems 1-4 in the body of the paper.


\subsubsection{T-PNS, CD-PNS, ND-PNS, and NI-PNS.}
{Supposing binary treatment, outcome, and mediator,
T-PNS, CD-PNS, ND-PNS, and NI-PNS reduce to
\begin{equation}
\text{\normalfont T-PNS}(c)\defeq\mathbb{P}(Y_{0}=0,Y_{1}=1|C=c),
\end{equation} 
\begin{equation}
\text{\normalfont CD-PNS}(m,c)\defeq\mathbb{P}(Y_{0,m}=0,Y_{1,m}=1|C=c),
\end{equation} 
\begin{equation}
\text{\normalfont ND-PNS}(c)\defeq\mathbb{P}(Y_{0}=0, Y_{1}=1,Y_{0,M_{1}}=0|C=c),
\end{equation} 
\begin{equation}
\text{\normalfont NI-PNS}(c)\defeq\mathbb{P}(Y_{0}=0, Y_{1}=1,Y_{0,M_{1}}=1|C=c)
\end{equation}
for any $c \in \Omega_C$, and the proportions of ND-PNS and NI-PNS are
${\text{\normalfont ND-PNS}(c)}/{\text{\normalfont T-PNS}(c)}=\mathbb{P}(Y_{0,M_{1}}=0|Y_{0}=0, Y_{1}=1,C=c)$ and 
${\text{\normalfont NI-PNS}(c)}/{\text{\normalfont T-PNS}(c)}=\mathbb{P}(Y_{0,M_{1}}=1|Y_{0}=0, Y_{1}=1,C=c)$ for any $c \in \Omega_C$, respectively.} 


\subsubsection{CD-PNS, ND-PNS, and NI-PNS with evidence.}
{Supposing binary treatment, outcome, and mediator, CD-PNS with evidence $(X=1,M=1,Y=1)$, and ND-PNS and NI-PNS with evidence $(X=1,Y=1)$ are
\begin{equation}
\begin{aligned}
&\text{\normalfont CD-PNS}(m,c)\\
&\hspace{0.2cm}\defeq\mathbb{P}(Y_{0,m}=0,Y_{1,m}=1|X=1,M=1,Y=1,C=c),
\end{aligned}
\end{equation} 
\begin{equation} 
\begin{aligned}
&\text{\normalfont T-PNS}(c)\\
&\hspace{0.2cm}\defeq\mathbb{P}(Y_{0}=0, Y_{1}=1|X=1,Y=1,C=c),
\end{aligned}
\end{equation} 
\begin{equation} 
\begin{aligned}
&\text{\normalfont ND-PNS}(c)\\
&\hspace{0.2cm}\defeq\mathbb{P}(Y_{0}=0, Y_{1}=1,  Y_{0,M_{1}}=0|X=1,Y=1,C=c),
\end{aligned}
\end{equation} 
\begin{equation} 
\begin{aligned}
&\text{\normalfont NI-PNS}(c)\\
&\hspace{0.2cm}\defeq\mathbb{P}(Y_{0}=0, Y_{1}=1,  Y_{0,M_{1}}=1|X=1,Y=1,C=c)
\end{aligned}
\end{equation} 
for any $c \in \Omega_C$.}



\section{Appendix E: Comparison with \citet{Rubinstein2024}}
In this section, we discuss the difference with the work by \citet{Rubinstein2024} vs. ours.
\citet{Rubinstein2024} provided two types of definitions of mediated PoC.
First, $\delta(c)$ means ``given that an individual would experience a negative outcome and a negative mediator under an exposure (i.e. $Y_1=1$ and $M_1=1$), when his/her covariate is $C=c$, what is the probability that the negative outcome would not have occurred absent the exposure?"
Second, $\tilde{\delta}(c)$ expresses the probability that ``an individual who received treatment and experienced a negative mediator and outcome under exposure, when his/her covariate is $C=c$, would not have experienced the negative outcome had they not been exposed."
They provided corresponding direct and indirect measures for them, respectively.
We compare the measures  by \citet{Rubinstein2024} and ours for binary treatment and outcome in Table \ref{tab:ap1}, which are all different.

\begin{table*}[t]
\centering
\renewcommand{\arraystretch}{1.4}
\begin{tabular}{l|l}
\hline
\hline
\citet{Rubinstein2024}& Definitions \\
\hline
The total mediated PoC & $\delta(c)\defeq\mathbb{P}(Y_0=0|Y_1=1,M_1=1,C=c)$ \\
The direct mediated PoC & $\psi(c)\defeq\mathbb{P}(Y_{1,M_0}=0,Y_{0,M_0}=0|Y_{1,M_1}=1,M_1=1,C=c)$  \\
The indirect mediated PoC &
$\zeta(c)\defeq\mathbb{P}(Y_{1,M_0}=1,Y_{0,M_0}=0|Y_{1,M_1}=1,M_1=1,C=c)$\\
\hline
The total mediated PoC & $\tilde{\delta}(c)\defeq\mathbb{P}(Y_0=0|Y=1,M=1,X=1,C=c)$ \\
The direct mediated PoC & $\tilde{\psi}(c)\defeq\mathbb{P}(Y_{1,M_0}=0,Y_{0,M_0}=0|Y=1,M=1,X=1,C=c)$  \\
The indirect mediated PoC &
$\tilde{\zeta}(c)\defeq\mathbb{P}(Y_{1,M_0}=1,Y_{0,M_0}=0|Y=1,M=1,X=1,C=c)$\\
\hline
\hline
Our measures for binary treatment and outcome & Definitions \\
\hline
T-PNS & $\text{\normalfont T-PNS}(c)\defeq\mathbb{P}(Y_{0}=0,Y_{1}=1|C=c)$\\
CD-PNS & $\text{\normalfont CD-PNS}(m,c)\defeq\mathbb{P}(Y_{0,m}=0,Y_{1,m}=1|C=c)$\\
ND-PNS & $\text{\normalfont ND-PNS}(c)\defeq\mathbb{P}(Y_{0}=0, Y_{1}=1,Y_{0,M_{1}}=0|C=c)$\\
NI-PNS & $\text{\normalfont NI-PNS}(c)\defeq\mathbb{P}(Y_{0}=0, Y_{1}=1,Y_{0,M_{1}}=1|C=c)$\\
\hline
T-PN &$\text{\normalfont T-PN}(c)\defeq\mathbb{P}(Y_{0}=0|Y=1,X=1,C=c)$\\
ND-PN & $\text{\normalfont ND-PN}(c)\defeq\mathbb{P}(Y_{0}=0,Y_{0,M_{1}}=0|Y=1,X=1,C=c)$\\
NI-PN & $\text{\normalfont NI-PN}(c)\defeq\mathbb{P}(Y_{0}=0,Y_{0,M_{1}}=1|Y=1,X=1,C=c)$\\
\hline
T-PS &$\text{\normalfont T-PS}(c)\defeq\mathbb{P}(Y_{1}=1|Y=0,X=0,C=c)$\\
ND-PS & $\text{\normalfont ND-PS}(c)\defeq\mathbb{P}(Y_{1}=1,Y_{0,M_{1}}=0|Y=0,X=0,C=c)$\\
NI-PS & $\text{\normalfont NI-PS}(c)\defeq\mathbb{P}(Y_{1}=1,Y_{0,M_{1}}=1|Y=0,X=0,C=c)$\\
\hline
\end{tabular}
\caption{Comparison of the definitions of PoC measures in causal mediation analysis}
\label{tab:ap1}
\end{table*}

\section{Appendix F: Simulated Experiments}

\subsection{Estimation from finite sample size}

We performe additional finite sample experiments for T-PN, ND-PN, NI-PN, T-PS, ND-PS, and NI-PS.
The ground truths of T-PN, ND-PN, and NI-PN are $0.088$, $0.076$, and $0.011$.
When $N=100$, the estimates of T-PN, ND-PN, and NI-PN are
\begin{center}
T-PN: $0.088$ ($95\%$CI:[$0.000,0.241$]),\\\vspace{0.2cm}
ND-PN: $0.079$ ($95\%$CI:[$0.000,0.232$]),\\\vspace{0.2cm}
NI-PN: $0.009$ ($95\%$CI:[$0.000,0.052$]).
\end{center}
When $N=1000$, the estimates of T-PN, ND-PN, and NI-PN are
\begin{center}
T-PN: $0.087$ ($95\%$CI:[$0.034,0.137$]),\\\vspace{0.2cm}
ND-PN: $0.077$ ($95\%$CI:[$0.023,0.127$]),\\\vspace{0.2cm}
NI-PN: $0.009$ ($95\%$CI:[$0.000,0.020$]).
\end{center}
When $N=10000$, the estimates of T-PN, ND-PN, and NI-PN are
\begin{center}
T-PN: $0.086$ ($95\%$CI:[$0.070,0.102$]),\\\vspace{0.2cm}
ND-PN: $0.077$ ($95\%$CI:[$0.062,0.093$]),\\\vspace{0.2cm}
NI-PN: $0.009$ ($95\%$CI:[$0.004,0.014$]).
\end{center}

Next, the ground truths of T-PS, ND-PS, and NI-PS are $0.097$, $0.084$, and $0.012$.
When $N=100$, the estimates of T-PS, ND-PS, and NI-PS are
\begin{center}
T-PS: $0.103$ ($95\%$CI:[$0.000,0.318$]),\\\vspace{0.2cm}
ND-PS: $0.093$ ($95\%$CI:[$0.000,0.307$]),\\\vspace{0.2cm}
NI-PS: $0.010$ ($95\%$CI:[$0.000,0.060$]).
\end{center}
When $N=1000$, the estimates of T-PS, ND-PS, and NI-PS are
\begin{center}
T-PS: $0.094$ ($95\%$CI:[$0.036,0.158$]),\\\vspace{0.2cm}
ND-PS: $0.085$ ($95\%$CI:[$0.024,0.147$]),\\\vspace{0.2cm}
NI-PS: $0.010$ ($95\%$CI:[$0.000,0.023$]).
\end{center}
When $N=10000$, the estimates of T-PS, ND-PS, and NI-PS are
\begin{center}
T-PS: $0.094$ ($95\%$CI:[$0.076,0.113$]),\\\vspace{0.2cm}
ND-PS: $0.085$ ($95\%$CI:[$0.067,0.103$]),\\\vspace{0.2cm}
NI-PS: $0.009$ ($95\%$CI:[$0.006,0.013$]).
\end{center}
When the sample size is small ($N=100$), the estimators have relatively wide 95$\%$ CIs and are slightly biased due to the $\max\{\cdot,0\}$ function in the estimators.
When the sample size is large enough ($N=1000$ or $N=10000$), the estimators are close to the ground truths and have relatively narrow 95$\%$ CIs.
All of our estimators are reliable when the sample size is large.

\subsection{Illustration of the proposed measures}
To illustrate the behavior of the proposed direct and indirect PoC measures, we simulate data from an SCM and plot the measures against the covariate.

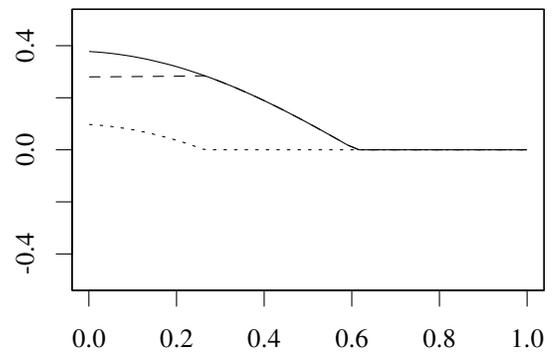
\begin{figure}
\vspace{0cm}
    \centering
\begin{tikzpicture}[x=1pt,y=1pt]
\definecolor{fillColor}{RGB}{255,255,255}
\path[use as bounding box,fill=fillColor,fill opacity=0.00] (0,0) rectangle (252.94,216.81);
\begin{scope}
\path[clip] ( 49.20, 61.20) rectangle (227.75,167.61);
\definecolor{drawColor}{RGB}{0,0,0}

\path[draw=drawColor,line width= 0.4pt,line join=round,line cap=round] ( 55.81,106.50) --
	( 56.34,106.50) --
	( 57.64,106.50) --
	( 58.96,106.50) --
	( 64.93,106.50) --
	( 65.83,106.50) --
	( 66.74,106.50) --
	( 66.92,106.50) --
	( 70.91,106.50) --
	( 75.80,106.48) --
	( 76.43,106.48) --
	( 76.50,106.48) --
	( 76.78,106.48) --
	( 77.81,106.47) --
	( 79.64,106.46) --
	( 81.74,106.45) --
	( 83.88,106.43) --
	( 87.13,106.41) --
	( 90.20,106.38) --
	( 91.49,106.36) --
	( 91.74,106.36) --
	( 92.23,106.35) --
	( 94.22,106.33) --
	( 96.10,106.30) --
	( 96.10,106.30) --
	( 97.57,106.27) --
	( 99.01,106.25) --
	(100.58,106.22) --
	(101.19,106.21) --
	(101.44,106.21) --
	(103.80,106.16) --
	(105.94,106.11) --
	(107.33,106.08) --
	(109.04,106.04) --
	(109.57,106.02) --
	(109.71,106.02) --
	(109.96,106.01) --
	(110.46,106.00) --
	(112.30,105.95) --
	(113.39,105.91) --
	(113.65,105.91) --
	(114.66,105.87) --
	(115.78,105.84) --
	(115.82,105.84) --
	(116.10,105.83) --
	(119.83,105.69) --
	(120.42,105.67) --
	(121.02,105.65) --
	(122.23,105.60) --
	(123.13,105.56) --
	(125.11,105.48) --
	(125.50,105.46) --
	(126.39,105.42) --
	(126.77,105.40) --
	(135.23,104.95) --
	(135.88,104.91) --
	(136.44,104.88) --
	(143.94,104.37) --
	(144.05,104.36) --
	(145.34,104.27) --
	(147.93,104.06) --
	(152.76,103.64) --
	(153.72,103.55) --
	(153.77,103.55) --
	(156.37,103.30) --
	(163.18,102.57) --
	(163.83,102.50) --
	(164.68,102.40) --
	(167.18,102.10) --
	(168.73,101.90) --
	(171.61,101.52) --
	(173.46,101.27) --
	(173.66,101.24) --
	(179.06,100.44) --
	(179.31,100.40) --
	(181.75,100.01) --
	(182.03, 99.97) --
	(183.61, 99.71) --
	(184.36, 99.58) --
	(186.64, 99.18) --
	(187.60, 99.01) --
	(194.07, 97.78) --
	(195.24, 97.55) --
	(195.51, 97.49) --
	(196.97, 97.19) --
	(200.27, 96.48) --
	(201.22, 96.27) --
	(201.72, 96.16) --
	(203.01, 95.87) --
	(204.37, 95.55) --
	(205.87, 95.20) --
	(210.78, 93.99) --
	(211.65, 93.77) --
	(211.87, 93.71) --
	(213.28, 93.35) --
	(215.19, 92.84) --
	(216.30, 92.53) --
	(218.17, 92.02) --
	(219.79, 91.56) --
	(221.13, 91.17);
\end{scope}
\begin{scope}
\path[clip] (  0.00,  0.00) rectangle (252.94,216.81);
\definecolor{drawColor}{RGB}{0,0,0}

\path[draw=drawColor,line width= 0.4pt,line join=round,line cap=round] ( 49.20, 61.20) --
	(227.75, 61.20) --
	(227.75,167.61) --
	( 49.20,167.61) --
	cycle;
\end{scope}
\begin{scope}
\path[clip] ( 49.20, 61.20) rectangle (227.75,167.61);
\definecolor{drawColor}{RGB}{0,0,0}

\path[draw=drawColor,line width= 0.4pt,dash pattern=on 4pt off 4pt ,line join=round,line cap=round] ( 55.81, 95.83) --
	( 56.34, 95.83) --
	( 57.64, 95.83) --
	( 58.96, 95.83) --
	( 64.93, 95.83) --
	( 65.83, 95.83) --
	( 66.74, 95.83) --
	( 66.92, 95.83) --
	( 70.91, 95.81) --
	( 75.80, 95.78) --
	( 76.43, 95.78) --
	( 76.50, 95.78) --
	( 76.78, 95.78) --
	( 77.81, 95.77) --
	( 79.64, 95.75) --
	( 81.74, 95.73) --
	( 83.88, 95.70) --
	( 87.13, 95.66) --
	( 90.20, 95.61) --
	( 91.49, 95.59) --
	( 91.74, 95.58) --
	( 92.23, 95.57) --
	( 94.22, 95.54) --
	( 96.10, 95.50) --
	( 96.10, 95.50) --
	( 97.57, 95.46) --
	( 99.01, 95.43) --
	(100.58, 95.39) --
	(101.19, 95.37) --
	(101.44, 95.36) --
	(103.80, 95.30) --
	(105.94, 95.23) --
	(107.33, 95.19) --
	(109.04, 95.13) --
	(109.57, 95.11) --
	(109.71, 95.11) --
	(109.96, 95.10) --
	(110.46, 95.08) --
	(112.30, 95.01) --
	(113.39, 94.97) --
	(113.65, 94.96) --
	(114.66, 94.92) --
	(115.78, 94.87) --
	(115.82, 94.87) --
	(116.10, 94.86) --
	(119.83, 94.69) --
	(120.42, 94.66) --
	(121.02, 94.64) --
	(122.23, 94.58) --
	(123.13, 94.53) --
	(125.11, 94.43) --
	(125.50, 94.40) --
	(126.39, 94.35) --
	(126.77, 94.33) --
	(135.23, 93.80) --
	(135.88, 93.76) --
	(136.44, 93.72) --
	(143.94, 93.14) --
	(144.05, 93.13) --
	(145.34, 93.02) --
	(147.93, 92.80) --
	(152.76, 92.34) --
	(153.72, 92.24) --
	(153.77, 92.24) --
	(156.37, 91.97) --
	(163.18, 91.20) --
	(163.83, 91.12) --
	(164.68, 91.01) --
	(167.18, 90.70) --
	(168.73, 90.49) --
	(171.61, 90.10) --
	(173.46, 89.84) --
	(173.66, 89.82) --
	(179.06, 89.01) --
	(179.31, 88.97) --
	(181.75, 88.58) --
	(182.03, 88.53) --
	(183.61, 88.27) --
	(184.36, 88.15) --
	(186.64, 87.76) --
	(187.60, 87.59) --
	(194.07, 86.39) --
	(195.24, 86.16) --
	(195.51, 86.11) --
	(196.97, 85.82) --
	(200.27, 85.15) --
	(201.22, 84.95) --
	(201.72, 84.84) --
	(203.01, 84.57) --
	(204.37, 84.27) --
	(205.87, 83.94) --
	(210.78, 82.81) --
	(211.65, 82.60) --
	(211.87, 82.55) --
	(213.28, 82.21) --
	(215.19, 81.75) --
	(216.30, 81.47) --
	(218.17, 81.00) --
	(219.79, 80.58) --
	(221.13, 80.23);
\end{scope}
\begin{scope}
\path[clip] (  0.00,  0.00) rectangle (252.94,216.81);
\definecolor{drawColor}{RGB}{0,0,0}

\path[draw=drawColor,line width= 0.4pt,line join=round,line cap=round] ( 49.20, 61.20) --
	(227.75, 61.20) --
	(227.75,167.61) --
	( 49.20,167.61) --
	cycle;
\end{scope}
\begin{scope}
\path[clip] ( 49.20, 61.20) rectangle (227.75,167.61);
\definecolor{drawColor}{RGB}{0,0,0}

\path[draw=drawColor,line width= 0.4pt,dash pattern=on 1pt off 3pt ,line join=round,line cap=round] ( 55.81, 75.81) --
	( 56.34, 75.81) --
	( 57.64, 75.81) --
	( 58.96, 75.81) --
	( 64.93, 75.81) --
	( 65.83, 75.81) --
	( 66.74, 75.82) --
	( 66.92, 75.82) --
	( 70.91, 75.82) --
	( 75.80, 75.84) --
	( 76.43, 75.84) --
	( 76.50, 75.84) --
	( 76.78, 75.84) --
	( 77.81, 75.84) --
	( 79.64, 75.85) --
	( 81.74, 75.86) --
	( 83.88, 75.87) --
	( 87.13, 75.89) --
	( 90.20, 75.91) --
	( 91.49, 75.91) --
	( 91.74, 75.92) --
	( 92.23, 75.92) --
	( 94.22, 75.93) --
	( 96.10, 75.94) --
	( 96.10, 75.94) --
	( 97.57, 75.95) --
	( 99.01, 75.96) --
	(100.58, 75.98) --
	(101.19, 75.98) --
	(101.44, 75.98) --
	(103.80, 76.00) --
	(105.94, 76.02) --
	(107.33, 76.03) --
	(109.04, 76.04) --
	(109.57, 76.05) --
	(109.71, 76.05) --
	(109.96, 76.05) --
	(110.46, 76.06) --
	(112.30, 76.07) --
	(113.39, 76.08) --
	(113.65, 76.09) --
	(114.66, 76.09) --
	(115.78, 76.10) --
	(115.82, 76.11) --
	(116.10, 76.11) --
	(119.83, 76.14) --
	(120.42, 76.15) --
	(121.02, 76.15) --
	(122.23, 76.16) --
	(123.13, 76.17) --
	(125.11, 76.19) --
	(125.50, 76.20) --
	(126.39, 76.20) --
	(126.77, 76.21) --
	(135.23, 76.29) --
	(135.88, 76.30) --
	(136.44, 76.30) --
	(143.94, 76.37) --
	(144.05, 76.37) --
	(145.34, 76.38) --
	(147.93, 76.41) --
	(152.76, 76.45) --
	(153.72, 76.45) --
	(153.77, 76.45) --
	(156.37, 76.47) --
	(163.18, 76.52) --
	(163.83, 76.52) --
	(164.68, 76.53) --
	(167.18, 76.54) --
	(168.73, 76.55) --
	(171.61, 76.56) --
	(173.46, 76.56) --
	(173.66, 76.57) --
	(179.06, 76.57) --
	(179.31, 76.57) --
	(181.75, 76.57) --
	(182.03, 76.57) --
	(183.61, 76.57) --
	(184.36, 76.57) --
	(186.64, 76.57) --
	(187.60, 76.56) --
	(194.07, 76.53) --
	(195.24, 76.52) --
	(195.51, 76.52) --
	(196.97, 76.51) --
	(200.27, 76.48) --
	(201.22, 76.47) --
	(201.72, 76.46) --
	(203.01, 76.44) --
	(204.37, 76.43) --
	(205.87, 76.41) --
	(210.78, 76.32) --
	(211.65, 76.31) --
	(211.87, 76.30) --
	(213.28, 76.27) --
	(215.19, 76.23) --
	(216.30, 76.21) --
	(218.17, 76.16) --
	(219.79, 76.12) --
	(221.13, 76.08);
\end{scope}
\begin{scope}
\path[clip] (  0.00,  0.00) rectangle (252.94,216.81);
\definecolor{drawColor}{RGB}{0,0,0}

\path[draw=drawColor,line width= 0.4pt,line join=round,line cap=round] ( 54.54, 61.20) -- (221.76, 61.20);

\path[draw=drawColor,line width= 0.4pt,line join=round,line cap=round] ( 54.54, 61.20) -- ( 54.54, 55.20);

\path[draw=drawColor,line width= 0.4pt,line join=round,line cap=round] ( 87.99, 61.20) -- ( 87.99, 55.20);

\path[draw=drawColor,line width= 0.4pt,line join=round,line cap=round] (121.43, 61.20) -- (121.43, 55.20);

\path[draw=drawColor,line width= 0.4pt,line join=round,line cap=round] (154.88, 61.20) -- (154.88, 55.20);

\path[draw=drawColor,line width= 0.4pt,line join=round,line cap=round] (188.32, 61.20) -- (188.32, 55.20);

\path[draw=drawColor,line width= 0.4pt,line join=round,line cap=round] (221.76, 61.20) -- (221.76, 55.20);

\node[text=drawColor,anchor=base,inner sep=0pt, outer sep=0pt, scale=  1.00] at ( 54.54, 39.60) {0.0};

\node[text=drawColor,anchor=base,inner sep=0pt, outer sep=0pt, scale=  1.00] at ( 87.99, 39.60) {0.2};

\node[text=drawColor,anchor=base,inner sep=0pt, outer sep=0pt, scale=  1.00] at (121.43, 39.60) {0.4};

\node[text=drawColor,anchor=base,inner sep=0pt, outer sep=0pt, scale=  1.00] at (154.88, 39.60) {0.6};

\node[text=drawColor,anchor=base,inner sep=0pt, outer sep=0pt, scale=  1.00] at (188.32, 39.60) {0.8};

\node[text=drawColor,anchor=base,inner sep=0pt, outer sep=0pt, scale=  1.00] at (221.76, 39.60) {1.0};

\path[draw=drawColor,line width= 0.4pt,line join=round,line cap=round] ( 49.20, 65.14) -- ( 49.20,152.72);

\path[draw=drawColor,line width= 0.4pt,line join=round,line cap=round] ( 49.20, 65.14) -- ( 43.20, 65.14);

\path[draw=drawColor,line width= 0.4pt,line join=round,line cap=round] ( 49.20, 87.04) -- ( 43.20, 87.04);

\path[draw=drawColor,line width= 0.4pt,line join=round,line cap=round] ( 49.20,108.93) -- ( 43.20,108.93);

\path[draw=drawColor,line width= 0.4pt,line join=round,line cap=round] ( 49.20,130.83) -- ( 43.20,130.83);

\path[draw=drawColor,line width= 0.4pt,line join=round,line cap=round] ( 49.20,152.72) -- ( 43.20,152.72);

\node[text=drawColor,rotate= 90.00,anchor=base,inner sep=0pt, outer sep=0pt, scale=  1.00] at ( 34.80, 65.14) {0.0};

\node[text=drawColor,rotate= 90.00,anchor=base,inner sep=0pt, outer sep=0pt, scale=  1.00] at ( 34.80, 87.04) {0.2};

\node[text=drawColor,rotate= 90.00,anchor=base,inner sep=0pt, outer sep=0pt, scale=  1.00] at ( 34.80,108.93) {0.4};

\node[text=drawColor,rotate= 90.00,anchor=base,inner sep=0pt, outer sep=0pt, scale=  1.00] at ( 34.80,130.83) {0.6};

\node[text=drawColor,rotate= 90.00,anchor=base,inner sep=0pt, outer sep=0pt, scale=  1.00] at ( 34.80,152.72) {0.8};

\path[draw=drawColor,line width= 0.4pt,line join=round,line cap=round] ( 49.20, 61.20) --
	(227.75, 61.20) --
	(227.75,167.61) --
	( 49.20,167.61) --
	cycle;
\end{scope}
\end{tikzpicture}
\vspace{-1.5cm}
\caption{Plots of T-PNS, ND-PNS, and NI-PNS. The solid line is T-PNS, the dashed line is NI-PNS, and the dotted line is NI-PNS. The x-axis represents the value of $C$, and the y-axis represents the values of T-PNS, ND-PNS, and NI-PNS, respectively.}
\label{fig:f1}
\end{figure}

We use the SCM setting in \cite[Appendix D]{Rubinstein2024}.
We plot T-PNS, ND-PNS, and NI-PNS in Figure \ref{fig:f1}, T-PN, ND-PN, and NI-PN in Figure \ref{fig:f2}, and T-PS, ND-PS, and NI-PS in Figure \ref{fig:f3}. 
The line of T-PNS is the same as the purple line in Figure 3 in \cite{Rubinstein2024}, while 
T-PNS, T-PN, and T-PS are all decomposed into direct and indirect influence differently. 
The total mediated PoC and indirect mediated PoC increase as the covariate value increases, and the direct mediated PoC decrease as the covariate value increases.
On the other hand, T-PNS and NI-PNS decrease as the covariate value increases, and ND-PNS is nearly constant.
The results imply that the direct influence is greater than the indirect influence around $C=0$, and the direct influence is almost the same as the indirect influence around $C=1$.

\begin{figure}
\vspace{0cm}
    \centering
\begin{tikzpicture}[x=1pt,y=1pt]
\definecolor{fillColor}{RGB}{255,255,255}
\path[use as bounding box,fill=fillColor,fill opacity=0.00] (0,0) rectangle (252.94,216.81);
\begin{scope}
\path[clip] ( 49.20, 61.20) rectangle (227.75,167.61);
\definecolor{drawColor}{RGB}{0,0,0}

\path[draw=drawColor,line width= 0.4pt,line join=round,line cap=round] ( 55.81,115.16) --
	( 56.34,115.19) --
	( 57.64,115.25) --
	( 58.96,115.31) --
	( 64.93,115.61) --
	( 65.83,115.66) --
	( 66.74,115.70) --
	( 66.92,115.71) --
	( 70.91,115.94) --
	( 75.80,116.23) --
	( 76.43,116.27) --
	( 76.50,116.27) --
	( 76.78,116.29) --
	( 77.81,116.35) --
	( 79.64,116.47) --
	( 81.74,116.61) --
	( 83.88,116.75) --
	( 87.13,116.98) --
	( 90.20,117.20) --
	( 91.49,117.29) --
	( 91.74,117.31) --
	( 92.23,117.35) --
	( 94.22,117.50) --
	( 96.10,117.64) --
	( 96.10,117.64) --
	( 97.57,117.76) --
	( 99.01,117.87) --
	(100.58,118.00) --
	(101.19,118.05) --
	(101.44,118.07) --
	(103.80,118.27) --
	(105.94,118.45) --
	(107.33,118.57) --
	(109.04,118.72) --
	(109.57,118.77) --
	(109.71,118.78) --
	(109.96,118.80) --
	(110.46,118.85) --
	(112.30,119.01) --
	(113.39,119.11) --
	(113.65,119.13) --
	(114.66,119.23) --
	(115.78,119.33) --
	(115.82,119.34) --
	(116.10,119.36) --
	(119.83,119.72) --
	(120.42,119.78) --
	(121.02,119.83) --
	(122.23,119.95) --
	(123.13,120.04) --
	(125.11,120.24) --
	(125.50,120.28) --
	(126.39,120.37) --
	(126.77,120.41) --
	(135.23,121.32) --
	(135.88,121.39) --
	(136.44,121.45) --
	(143.94,122.32) --
	(144.05,122.33) --
	(145.34,122.49) --
	(147.93,122.80) --
	(152.76,123.41) --
	(153.72,123.54) --
	(153.77,123.54) --
	(156.37,123.89) --
	(163.18,124.83) --
	(163.83,124.92) --
	(164.68,125.04) --
	(167.18,125.41) --
	(168.73,125.65) --
	(171.61,126.09) --
	(173.46,126.38) --
	(173.66,126.42) --
	(179.06,127.31) --
	(179.31,127.36) --
	(181.75,127.79) --
	(182.03,127.84) --
	(183.61,128.12) --
	(184.36,128.26) --
	(186.64,128.69) --
	(187.60,128.88) --
	(194.07,130.21) --
	(195.24,130.47) --
	(195.51,130.53) --
	(196.97,130.86) --
	(200.27,131.65) --
	(201.22,131.89) --
	(201.72,132.01) --
	(203.01,132.35) --
	(204.37,132.71) --
	(205.87,133.12) --
	(210.78,134.59) --
	(211.65,134.87) --
	(211.87,134.94) --
	(213.28,135.42) --
	(215.19,136.09) --
	(216.30,136.50) --
	(218.17,137.23) --
	(219.79,137.90) --
	(221.13,138.48);
\end{scope}
\begin{scope}
\path[clip] (  0.00,  0.00) rectangle (252.94,216.81);
\definecolor{drawColor}{RGB}{0,0,0}

\path[draw=drawColor,line width= 0.4pt,line join=round,line cap=round] ( 49.20, 61.20) --
	(227.75, 61.20) --
	(227.75,167.61) --
	( 49.20,167.61) --
	cycle;
\end{scope}
\begin{scope}
\path[clip] ( 49.20, 61.20) rectangle (227.75,167.61);
\definecolor{drawColor}{RGB}{0,0,0}

\path[draw=drawColor,line width= 0.4pt,dash pattern=on 4pt off 4pt ,line join=round,line cap=round] ( 55.81,102.25) --
	( 56.34,102.27) --
	( 57.64,102.32) --
	( 58.96,102.37) --
	( 64.93,102.59) --
	( 65.83,102.62) --
	( 66.74,102.65) --
	( 66.92,102.66) --
	( 70.91,102.82) --
	( 75.80,103.01) --
	( 76.43,103.04) --
	( 76.50,103.04) --
	( 76.78,103.05) --
	( 77.81,103.09) --
	( 79.64,103.17) --
	( 81.74,103.25) --
	( 83.88,103.34) --
	( 87.13,103.48) --
	( 90.20,103.61) --
	( 91.49,103.66) --
	( 91.74,103.67) --
	( 92.23,103.70) --
	( 94.22,103.78) --
	( 96.10,103.86) --
	( 96.10,103.86) --
	( 97.57,103.93) --
	( 99.01,103.99) --
	(100.58,104.06) --
	(101.19,104.09) --
	(101.44,104.10) --
	(103.80,104.20) --
	(105.94,104.30) --
	(107.33,104.36) --
	(109.04,104.43) --
	(109.57,104.46) --
	(109.71,104.46) --
	(109.96,104.47) --
	(110.46,104.50) --
	(112.30,104.58) --
	(113.39,104.63) --
	(113.65,104.64) --
	(114.66,104.68) --
	(115.78,104.73) --
	(115.82,104.73) --
	(116.10,104.75) --
	(119.83,104.91) --
	(120.42,104.94) --
	(121.02,104.97) --
	(122.23,105.02) --
	(123.13,105.06) --
	(125.11,105.15) --
	(125.50,105.16) --
	(126.39,105.20) --
	(126.77,105.22) --
	(135.23,105.59) --
	(135.88,105.61) --
	(136.44,105.64) --
	(143.94,105.95) --
	(144.05,105.96) --
	(145.34,106.01) --
	(147.93,106.11) --
	(152.76,106.30) --
	(153.72,106.34) --
	(153.77,106.34) --
	(156.37,106.44) --
	(163.18,106.69) --
	(163.83,106.71) --
	(164.68,106.74) --
	(167.18,106.82) --
	(168.73,106.87) --
	(171.61,106.96) --
	(173.46,107.02) --
	(173.66,107.02) --
	(179.06,107.18) --
	(179.31,107.18) --
	(181.75,107.25) --
	(182.03,107.25) --
	(183.61,107.29) --
	(184.36,107.31) --
	(186.64,107.36) --
	(187.60,107.38) --
	(194.07,107.51) --
	(195.24,107.53) --
	(195.51,107.53) --
	(196.97,107.55) --
	(200.27,107.60) --
	(201.22,107.61) --
	(201.72,107.61) --
	(203.01,107.62) --
	(204.37,107.64) --
	(205.87,107.65) --
	(210.78,107.67) --
	(211.65,107.68) --
	(211.87,107.68) --
	(213.28,107.68) --
	(215.19,107.68) --
	(216.30,107.68) --
	(218.17,107.67) --
	(219.79,107.67) --
	(221.13,107.66);
\end{scope}
\begin{scope}
\path[clip] (  0.00,  0.00) rectangle (252.94,216.81);
\definecolor{drawColor}{RGB}{0,0,0}

\path[draw=drawColor,line width= 0.4pt,line join=round,line cap=round] ( 49.20, 61.20) --
	(227.75, 61.20) --
	(227.75,167.61) --
	( 49.20,167.61) --
	cycle;
\end{scope}
\begin{scope}
\path[clip] ( 49.20, 61.20) rectangle (227.75,167.61);
\definecolor{drawColor}{RGB}{0,0,0}

\path[draw=drawColor,line width= 0.4pt,dash pattern=on 1pt off 3pt ,line join=round,line cap=round] ( 55.81, 78.05) --
	( 56.34, 78.05) --
	( 57.64, 78.07) --
	( 58.96, 78.08) --
	( 64.93, 78.16) --
	( 65.83, 78.18) --
	( 66.74, 78.19) --
	( 66.92, 78.19) --
	( 70.91, 78.26) --
	( 75.80, 78.36) --
	( 76.43, 78.37) --
	( 76.50, 78.38) --
	( 76.78, 78.38) --
	( 77.81, 78.40) --
	( 79.64, 78.45) --
	( 81.74, 78.50) --
	( 83.88, 78.55) --
	( 87.13, 78.64) --
	( 90.20, 78.73) --
	( 91.49, 78.77) --
	( 91.74, 78.78) --
	( 92.23, 78.79) --
	( 94.22, 78.86) --
	( 96.10, 78.92) --
	( 96.10, 78.92) --
	( 97.57, 78.97) --
	( 99.01, 79.03) --
	(100.58, 79.08) --
	(101.19, 79.11) --
	(101.44, 79.12) --
	(103.80, 79.21) --
	(105.94, 79.29) --
	(107.33, 79.35) --
	(109.04, 79.43) --
	(109.57, 79.45) --
	(109.71, 79.46) --
	(109.96, 79.47) --
	(110.46, 79.49) --
	(112.30, 79.57) --
	(113.39, 79.62) --
	(113.65, 79.64) --
	(114.66, 79.69) --
	(115.78, 79.74) --
	(115.82, 79.74) --
	(116.10, 79.76) --
	(119.83, 79.95) --
	(120.42, 79.98) --
	(121.02, 80.01) --
	(122.23, 80.08) --
	(123.13, 80.13) --
	(125.11, 80.24) --
	(125.50, 80.26) --
	(126.39, 80.31) --
	(126.77, 80.34) --
	(135.23, 80.87) --
	(135.88, 80.92) --
	(136.44, 80.96) --
	(143.94, 81.51) --
	(144.05, 81.52) --
	(145.34, 81.62) --
	(147.93, 81.83) --
	(152.76, 82.25) --
	(153.72, 82.34) --
	(153.77, 82.34) --
	(156.37, 82.59) --
	(163.18, 83.28) --
	(163.83, 83.36) --
	(164.68, 83.45) --
	(167.18, 83.73) --
	(168.73, 83.92) --
	(171.61, 84.27) --
	(173.46, 84.51) --
	(173.66, 84.53) --
	(179.06, 85.28) --
	(179.31, 85.32) --
	(181.75, 85.68) --
	(182.03, 85.72) --
	(183.61, 85.97) --
	(184.36, 86.09) --
	(186.64, 86.47) --
	(187.60, 86.64) --
	(194.07, 87.85) --
	(195.24, 88.09) --
	(195.51, 88.14) --
	(196.97, 88.45) --
	(200.27, 89.20) --
	(201.22, 89.42) --
	(201.72, 89.54) --
	(203.01, 89.86) --
	(204.37, 90.21) --
	(205.87, 90.62) --
	(210.78, 92.06) --
	(211.65, 92.34) --
	(211.87, 92.41) --
	(213.28, 92.88) --
	(215.19, 93.55) --
	(216.30, 93.96) --
	(218.17, 94.69) --
	(219.79, 95.37) --
	(221.13, 95.96);
\end{scope}
\begin{scope}
\path[clip] (  0.00,  0.00) rectangle (252.94,216.81);
\definecolor{drawColor}{RGB}{0,0,0}

\path[draw=drawColor,line width= 0.4pt,line join=round,line cap=round] ( 54.54, 61.20) -- (221.76, 61.20);

\path[draw=drawColor,line width= 0.4pt,line join=round,line cap=round] ( 54.54, 61.20) -- ( 54.54, 55.20);

\path[draw=drawColor,line width= 0.4pt,line join=round,line cap=round] ( 87.99, 61.20) -- ( 87.99, 55.20);

\path[draw=drawColor,line width= 0.4pt,line join=round,line cap=round] (121.43, 61.20) -- (121.43, 55.20);

\path[draw=drawColor,line width= 0.4pt,line join=round,line cap=round] (154.88, 61.20) -- (154.88, 55.20);

\path[draw=drawColor,line width= 0.4pt,line join=round,line cap=round] (188.32, 61.20) -- (188.32, 55.20);

\path[draw=drawColor,line width= 0.4pt,line join=round,line cap=round] (221.76, 61.20) -- (221.76, 55.20);

\node[text=drawColor,anchor=base,inner sep=0pt, outer sep=0pt, scale=  1.00] at ( 54.54, 39.60) {0.0};

\node[text=drawColor,anchor=base,inner sep=0pt, outer sep=0pt, scale=  1.00] at ( 87.99, 39.60) {0.2};

\node[text=drawColor,anchor=base,inner sep=0pt, outer sep=0pt, scale=  1.00] at (121.43, 39.60) {0.4};

\node[text=drawColor,anchor=base,inner sep=0pt, outer sep=0pt, scale=  1.00] at (154.88, 39.60) {0.6};

\node[text=drawColor,anchor=base,inner sep=0pt, outer sep=0pt, scale=  1.00] at (188.32, 39.60) {0.8};

\node[text=drawColor,anchor=base,inner sep=0pt, outer sep=0pt, scale=  1.00] at (221.76, 39.60) {1.0};

\path[draw=drawColor,line width= 0.4pt,line join=round,line cap=round] ( 49.20, 65.14) -- ( 49.20,152.72);

\path[draw=drawColor,line width= 0.4pt,line join=round,line cap=round] ( 49.20, 65.14) -- ( 43.20, 65.14);

\path[draw=drawColor,line width= 0.4pt,line join=round,line cap=round] ( 49.20, 87.04) -- ( 43.20, 87.04);

\path[draw=drawColor,line width= 0.4pt,line join=round,line cap=round] ( 49.20,108.93) -- ( 43.20,108.93);

\path[draw=drawColor,line width= 0.4pt,line join=round,line cap=round] ( 49.20,130.83) -- ( 43.20,130.83);

\path[draw=drawColor,line width= 0.4pt,line join=round,line cap=round] ( 49.20,152.72) -- ( 43.20,152.72);

\node[text=drawColor,rotate= 90.00,anchor=base,inner sep=0pt, outer sep=0pt, scale=  1.00] at ( 34.80, 65.14) {0.0};

\node[text=drawColor,rotate= 90.00,anchor=base,inner sep=0pt, outer sep=0pt, scale=  1.00] at ( 34.80, 87.04) {0.2};

\node[text=drawColor,rotate= 90.00,anchor=base,inner sep=0pt, outer sep=0pt, scale=  1.00] at ( 34.80,108.93) {0.4};

\node[text=drawColor,rotate= 90.00,anchor=base,inner sep=0pt, outer sep=0pt, scale=  1.00] at ( 34.80,130.83) {0.6};

\node[text=drawColor,rotate= 90.00,anchor=base,inner sep=0pt, outer sep=0pt, scale=  1.00] at ( 34.80,152.72) {0.8};

\path[draw=drawColor,line width= 0.4pt,line join=round,line cap=round] ( 49.20, 61.20) --
	(227.75, 61.20) --
	(227.75,167.61) --
	( 49.20,167.61) --
	cycle;
\end{scope}
\end{tikzpicture}
\vspace{-1.5cm}
\caption{Plots of T-PN, ND-PN, and NI-PN. The solid line is T-PN, the dashed line is NI-PN, and the dotted line is NI-PN. The x-axis represents the value of $C$, and the y-axis represents the values of T-PN, ND-PN, and NI-PN, respectively.}
\label{fig:f2}
\end{figure}

\begin{figure}
\vspace{0cm}
    \centering
\begin{tikzpicture}[x=1pt,y=1pt]
\definecolor{fillColor}{RGB}{255,255,255}
\path[use as bounding box,fill=fillColor,fill opacity=0.00] (0,0) rectangle (252.94,216.81);
\begin{scope}
\path[clip] ( 49.20, 61.20) rectangle (227.75,167.61);
\definecolor{drawColor}{RGB}{0,0,0}

\path[draw=drawColor,line width= 0.4pt,line join=round,line cap=round] ( 55.81,140.22) --
	( 56.34,140.17) --
	( 57.64,140.03) --
	( 58.96,139.90) --
	( 64.93,139.24) --
	( 65.83,139.14) --
	( 66.74,139.03) --
	( 66.92,139.01) --
	( 70.91,138.51) --
	( 75.80,137.87) --
	( 76.43,137.78) --
	( 76.50,137.77) --
	( 76.78,137.73) --
	( 77.81,137.59) --
	( 79.64,137.33) --
	( 81.74,137.02) --
	( 83.88,136.69) --
	( 87.13,136.18) --
	( 90.20,135.68) --
	( 91.49,135.46) --
	( 91.74,135.42) --
	( 92.23,135.34) --
	( 94.22,134.99) --
	( 96.10,134.66) --
	( 96.10,134.66) --
	( 97.57,134.39) --
	( 99.01,134.12) --
	(100.58,133.83) --
	(101.19,133.71) --
	(101.44,133.67) --
	(103.80,133.21) --
	(105.94,132.78) --
	(107.33,132.50) --
	(109.04,132.15) --
	(109.57,132.04) --
	(109.71,132.01) --
	(109.96,131.96) --
	(110.46,131.85) --
	(112.30,131.46) --
	(113.39,131.23) --
	(113.65,131.17) --
	(114.66,130.95) --
	(115.78,130.70) --
	(115.82,130.69) --
	(116.10,130.63) --
	(119.83,129.78) --
	(120.42,129.64) --
	(121.02,129.50) --
	(122.23,129.21) --
	(123.13,129.00) --
	(125.11,128.52) --
	(125.50,128.42) --
	(126.39,128.20) --
	(126.77,128.11) --
	(135.23,125.93) --
	(135.88,125.75) --
	(136.44,125.60) --
	(143.94,123.50) --
	(144.05,123.47) --
	(145.34,123.09) --
	(147.93,122.33) --
	(152.76,120.86) --
	(153.72,120.56) --
	(153.77,120.55) --
	(156.37,119.73) --
	(163.18,117.51) --
	(163.83,117.29) --
	(164.68,117.01) --
	(167.18,116.16) --
	(168.73,115.62) --
	(171.61,114.61) --
	(173.46,113.95) --
	(173.66,113.88) --
	(179.06,111.92) --
	(179.31,111.82) --
	(181.75,110.91) --
	(182.03,110.81) --
	(183.61,110.21) --
	(184.36,109.92) --
	(186.64,109.05) --
	(187.60,108.67) --
	(194.07,106.12) --
	(195.24,105.65) --
	(195.51,105.54) --
	(196.97,104.95) --
	(200.27,103.60) --
	(201.22,103.20) --
	(201.72,102.99) --
	(203.01,102.46) --
	(204.37,101.89) --
	(205.87,101.25) --
	(210.78, 99.16) --
	(211.65, 98.78) --
	(211.87, 98.69) --
	(213.28, 98.08) --
	(215.19, 97.25) --
	(216.30, 96.76) --
	(218.17, 95.94) --
	(219.79, 95.22) --
	(221.13, 94.63);
\end{scope}
\begin{scope}
\path[clip] (  0.00,  0.00) rectangle (252.94,216.81);
\definecolor{drawColor}{RGB}{0,0,0}

\path[draw=drawColor,line width= 0.4pt,line join=round,line cap=round] ( 49.20, 61.20) --
	(227.75, 61.20) --
	(227.75,167.61) --
	( 49.20,167.61) --
	cycle;
\end{scope}
\begin{scope}
\path[clip] ( 49.20, 61.20) rectangle (227.75,167.61);
\definecolor{drawColor}{RGB}{0,0,0}

\path[draw=drawColor,line width= 0.4pt,dash pattern=on 4pt off 4pt ,line join=round,line cap=round] ( 55.81,120.85) --
	( 56.34,120.81) --
	( 57.64,120.71) --
	( 58.96,120.61) --
	( 64.93,120.12) --
	( 65.83,120.04) --
	( 66.74,119.96) --
	( 66.92,119.94) --
	( 70.91,119.56) --
	( 75.80,119.05) --
	( 76.43,118.98) --
	( 76.50,118.97) --
	( 76.78,118.94) --
	( 77.81,118.82) --
	( 79.64,118.61) --
	( 81.74,118.36) --
	( 83.88,118.10) --
	( 87.13,117.68) --
	( 90.20,117.26) --
	( 91.49,117.08) --
	( 91.74,117.05) --
	( 92.23,116.98) --
	( 94.22,116.69) --
	( 96.10,116.41) --
	( 96.10,116.41) --
	( 97.57,116.19) --
	( 99.01,115.96) --
	(100.58,115.71) --
	(101.19,115.62) --
	(101.44,115.58) --
	(103.80,115.19) --
	(105.94,114.83) --
	(107.33,114.59) --
	(109.04,114.28) --
	(109.57,114.19) --
	(109.71,114.17) --
	(109.96,114.12) --
	(110.46,114.03) --
	(112.30,113.69) --
	(113.39,113.49) --
	(113.65,113.44) --
	(114.66,113.25) --
	(115.78,113.04) --
	(115.82,113.03) --
	(116.10,112.98) --
	(119.83,112.24) --
	(120.42,112.12) --
	(121.02,112.00) --
	(122.23,111.75) --
	(123.13,111.57) --
	(125.11,111.15) --
	(125.50,111.07) --
	(126.39,110.88) --
	(126.77,110.80) --
	(135.23,108.90) --
	(135.88,108.75) --
	(136.44,108.62) --
	(143.94,106.80) --
	(144.05,106.77) --
	(145.34,106.44) --
	(147.93,105.78) --
	(152.76,104.50) --
	(153.72,104.24) --
	(153.77,104.23) --
	(156.37,103.52) --
	(163.18,101.59) --
	(163.83,101.40) --
	(164.68,101.16) --
	(167.18,100.42) --
	(168.73, 99.96) --
	(171.61, 99.09) --
	(173.46, 98.52) --
	(173.66, 98.46) --
	(179.06, 96.77) --
	(179.31, 96.69) --
	(181.75, 95.90) --
	(182.03, 95.82) --
	(183.61, 95.30) --
	(184.36, 95.06) --
	(186.64, 94.31) --
	(187.60, 93.99) --
	(194.07, 91.82) --
	(195.24, 91.42) --
	(195.51, 91.33) --
	(196.97, 90.83) --
	(200.27, 89.69) --
	(201.22, 89.36) --
	(201.72, 89.18) --
	(203.01, 88.73) --
	(204.37, 88.25) --
	(205.87, 87.72) --
	(210.78, 85.98) --
	(211.65, 85.66) --
	(211.87, 85.58) --
	(213.28, 85.08) --
	(215.19, 84.39) --
	(216.30, 83.99) --
	(218.17, 83.31) --
	(219.79, 82.72) --
	(221.13, 82.24);
\end{scope}
\begin{scope}
\path[clip] (  0.00,  0.00) rectangle (252.94,216.81);
\definecolor{drawColor}{RGB}{0,0,0}

\path[draw=drawColor,line width= 0.4pt,line join=round,line cap=round] ( 49.20, 61.20) --
	(227.75, 61.20) --
	(227.75,167.61) --
	( 49.20,167.61) --
	cycle;
\end{scope}
\begin{scope}
\path[clip] ( 49.20, 61.20) rectangle (227.75,167.61);
\definecolor{drawColor}{RGB}{0,0,0}

\path[draw=drawColor,line width= 0.4pt,dash pattern=on 1pt off 3pt ,line join=round,line cap=round] ( 55.81, 84.51) --
	( 56.34, 84.50) --
	( 57.64, 84.46) --
	( 58.96, 84.43) --
	( 64.93, 84.26) --
	( 65.83, 84.24) --
	( 66.74, 84.21) --
	( 66.92, 84.21) --
	( 70.91, 84.09) --
	( 75.80, 83.96) --
	( 76.43, 83.94) --
	( 76.50, 83.94) --
	( 76.78, 83.93) --
	( 77.81, 83.90) --
	( 79.64, 83.85) --
	( 81.74, 83.79) --
	( 83.88, 83.73) --
	( 87.13, 83.64) --
	( 90.20, 83.56) --
	( 91.49, 83.52) --
	( 91.74, 83.51) --
	( 92.23, 83.50) --
	( 94.22, 83.44) --
	( 96.10, 83.39) --
	( 96.10, 83.39) --
	( 97.57, 83.35) --
	( 99.01, 83.30) --
	(100.58, 83.26) --
	(101.19, 83.24) --
	(101.44, 83.23) --
	(103.80, 83.16) --
	(105.94, 83.10) --
	(107.33, 83.06) --
	(109.04, 83.01) --
	(109.57, 82.99) --
	(109.71, 82.99) --
	(109.96, 82.98) --
	(110.46, 82.97) --
	(112.30, 82.91) --
	(113.39, 82.88) --
	(113.65, 82.87) --
	(114.66, 82.84) --
	(115.78, 82.80) --
	(115.82, 82.80) --
	(116.10, 82.79) --
	(119.83, 82.68) --
	(120.42, 82.66) --
	(121.02, 82.64) --
	(122.23, 82.60) --
	(123.13, 82.57) --
	(125.11, 82.51) --
	(125.50, 82.49) --
	(126.39, 82.46) --
	(126.77, 82.45) --
	(135.23, 82.16) --
	(135.88, 82.14) --
	(136.44, 82.12) --
	(143.94, 81.85) --
	(144.05, 81.84) --
	(145.34, 81.79) --
	(147.93, 81.69) --
	(152.76, 81.50) --
	(153.72, 81.46) --
	(153.77, 81.46) --
	(156.37, 81.35) --
	(163.18, 81.06) --
	(163.83, 81.03) --
	(164.68, 80.99) --
	(167.18, 80.88) --
	(168.73, 80.80) --
	(171.61, 80.67) --
	(173.46, 80.58) --
	(173.66, 80.57) --
	(179.06, 80.29) --
	(179.31, 80.28) --
	(181.75, 80.15) --
	(182.03, 80.13) --
	(183.61, 80.05) --
	(184.36, 80.01) --
	(186.64, 79.88) --
	(187.60, 79.82) --
	(194.07, 79.44) --
	(195.24, 79.37) --
	(195.51, 79.35) --
	(196.97, 79.26) --
	(200.27, 79.05) --
	(201.22, 78.99) --
	(201.72, 78.95) --
	(203.01, 78.87) --
	(204.37, 78.78) --
	(205.87, 78.67) --
	(210.78, 78.33) --
	(211.65, 78.26) --
	(211.87, 78.25) --
	(213.28, 78.14) --
	(215.19, 78.00) --
	(216.30, 77.91) --
	(218.17, 77.77) --
	(219.79, 77.64) --
	(221.13, 77.53);
\end{scope}
\begin{scope}
\path[clip] (  0.00,  0.00) rectangle (252.94,216.81);
\definecolor{drawColor}{RGB}{0,0,0}

\path[draw=drawColor,line width= 0.4pt,line join=round,line cap=round] ( 54.54, 61.20) -- (221.76, 61.20);

\path[draw=drawColor,line width= 0.4pt,line join=round,line cap=round] ( 54.54, 61.20) -- ( 54.54, 55.20);

\path[draw=drawColor,line width= 0.4pt,line join=round,line cap=round] ( 87.99, 61.20) -- ( 87.99, 55.20);

\path[draw=drawColor,line width= 0.4pt,line join=round,line cap=round] (121.43, 61.20) -- (121.43, 55.20);

\path[draw=drawColor,line width= 0.4pt,line join=round,line cap=round] (154.88, 61.20) -- (154.88, 55.20);

\path[draw=drawColor,line width= 0.4pt,line join=round,line cap=round] (188.32, 61.20) -- (188.32, 55.20);

\path[draw=drawColor,line width= 0.4pt,line join=round,line cap=round] (221.76, 61.20) -- (221.76, 55.20);

\node[text=drawColor,anchor=base,inner sep=0pt, outer sep=0pt, scale=  1.00] at ( 54.54, 39.60) {0.0};

\node[text=drawColor,anchor=base,inner sep=0pt, outer sep=0pt, scale=  1.00] at ( 87.99, 39.60) {0.2};

\node[text=drawColor,anchor=base,inner sep=0pt, outer sep=0pt, scale=  1.00] at (121.43, 39.60) {0.4};

\node[text=drawColor,anchor=base,inner sep=0pt, outer sep=0pt, scale=  1.00] at (154.88, 39.60) {0.6};

\node[text=drawColor,anchor=base,inner sep=0pt, outer sep=0pt, scale=  1.00] at (188.32, 39.60) {0.8};

\node[text=drawColor,anchor=base,inner sep=0pt, outer sep=0pt, scale=  1.00] at (221.76, 39.60) {1.0};

\path[draw=drawColor,line width= 0.4pt,line join=round,line cap=round] ( 49.20, 65.14) -- ( 49.20,152.72);

\path[draw=drawColor,line width= 0.4pt,line join=round,line cap=round] ( 49.20, 65.14) -- ( 43.20, 65.14);

\path[draw=drawColor,line width= 0.4pt,line join=round,line cap=round] ( 49.20, 87.04) -- ( 43.20, 87.04);

\path[draw=drawColor,line width= 0.4pt,line join=round,line cap=round] ( 49.20,108.93) -- ( 43.20,108.93);

\path[draw=drawColor,line width= 0.4pt,line join=round,line cap=round] ( 49.20,130.83) -- ( 43.20,130.83);

\path[draw=drawColor,line width= 0.4pt,line join=round,line cap=round] ( 49.20,152.72) -- ( 43.20,152.72);

\node[text=drawColor,rotate= 90.00,anchor=base,inner sep=0pt, outer sep=0pt, scale=  1.00] at ( 34.80, 65.14) {0.0};

\node[text=drawColor,rotate= 90.00,anchor=base,inner sep=0pt, outer sep=0pt, scale=  1.00] at ( 34.80, 87.04) {0.2};

\node[text=drawColor,rotate= 90.00,anchor=base,inner sep=0pt, outer sep=0pt, scale=  1.00] at ( 34.80,108.93) {0.4};

\node[text=drawColor,rotate= 90.00,anchor=base,inner sep=0pt, outer sep=0pt, scale=  1.00] at ( 34.80,130.83) {0.6};

\node[text=drawColor,rotate= 90.00,anchor=base,inner sep=0pt, outer sep=0pt, scale=  1.00] at ( 34.80,152.72) {0.8};

\path[draw=drawColor,line width= 0.4pt,line join=round,line cap=round] ( 49.20, 61.20) --
	(227.75, 61.20) --
	(227.75,167.61) --
	( 49.20,167.61) --
	cycle;
\end{scope}
\end{tikzpicture}
\vspace{-1.5cm}
\caption{Plots of T-PS, ND-PS, and NI-PS. The solid line is T-PS, the dashed line is NI-PS, and the dotted line is NI-PS. The x-axis represents the value of $C$, and the y-axis represents the values of T-PS, ND-PS, and NI-PS, respectively.}
\label{fig:f3}
\end{figure}

Next, we plot TE, NDE, and NIE  in Figure \ref{fig:a1} and T-PNS, ND-PNS, and NI-PNS  in Figure \ref{fig:a2}.
The settings are the same as that in \cite[Appendix D]{Rubinstein2024} except changing from $\tilde{\mu}_{11}(C)=0.4-0.1C-0.2C^2$ to $\tilde{\mu}_{11}(C)=0.4+0.1C+2C^2$. 
T-PNS decreases as the covariate value increases, ND-PNS and NI-PNS increase as the covariate value increases, and TE, NDE, and NIE take negative values.
The results show there  exist considerable direct and indirect influences for most value of the covariate.
On the other hand, T-PNS, ND-PNS, and NI-PNS never take negative values because they are probabilities.
T-PNS, ND-PNS, and NI-PNS decrease as the covariate value increases.
ND-PNS and NI-PNS are always under T-PNS, and ND-PNS and NI-PNS are 0 when T-PNS is equal to 0.
The results show that there exist direct and indirect influences for $C<0.2$, and there exists only indirect influence around $C=0.2$.

\begin{figure}
\vspace{0cm}
    \centering
\begin{tikzpicture}[x=1pt,y=1pt]
\definecolor{fillColor}{RGB}{255,255,255}
\path[use as bounding box,fill=fillColor,fill opacity=0.00] (0,0) rectangle (252.94,216.81);
\begin{scope}
\path[clip] ( 49.20, 61.20) rectangle (227.75,167.61);
\definecolor{drawColor}{RGB}{0,0,0}

\path[draw=drawColor,line width= 0.4pt,line join=round,line cap=round] ( 55.81,151.62) --
	( 58.64,151.43) --
	( 58.91,151.41) --
	( 59.31,151.38) --
	( 59.37,151.37) --
	( 59.47,151.36) --
	( 61.28,151.20) --
	( 62.75,151.05) --
	( 63.00,151.03) --
	( 63.00,151.03) --
	( 64.45,150.86) --
	( 65.78,150.70) --
	( 66.74,150.57) --
	( 70.42,150.02) --
	( 70.81,149.96) --
	( 72.15,149.73) --
	( 73.12,149.56) --
	( 73.27,149.53) --
	( 77.06,148.79) --
	( 78.69,148.44) --
	( 85.33,146.83) --
	( 85.71,146.73) --
	( 87.74,146.17) --
	( 88.01,146.10) --
	( 88.14,146.06) --
	( 88.79,145.87) --
	( 89.33,145.72) --
	( 90.87,145.26) --
	( 93.02,144.59) --
	( 98.26,142.85) --
	( 98.69,142.70) --
	( 98.83,142.65) --
	(100.69,141.99) --
	(101.43,141.72) --
	(105.02,140.35) --
	(108.77,138.85) --
	(108.82,138.83) --
	(110.45,138.16) --
	(111.81,137.58) --
	(112.13,137.44) --
	(116.04,135.72) --
	(116.37,135.58) --
	(118.79,134.47) --
	(119.44,134.17) --
	(120.40,133.72) --
	(120.56,133.64) --
	(121.41,133.24) --
	(123.65,132.17) --
	(126.87,130.58) --
	(130.33,128.83) --
	(131.36,128.30) --
	(134.23,126.81) --
	(135.77,125.99) --
	(138.67,124.44) --
	(140.17,123.62) --
	(140.71,123.33) --
	(142.84,122.16) --
	(143.23,121.94) --
	(143.44,121.83) --
	(144.86,121.04) --
	(149.04,118.70) --
	(149.77,118.29) --
	(150.59,117.82) --
	(153.45,116.20) --
	(157.74,113.75) --
	(159.33,112.84) --
	(159.72,112.61) --
	(160.19,112.35) --
	(163.40,110.50) --
	(164.15,110.08) --
	(164.48,109.89) --
	(164.63,109.80) --
	(165.22,109.46) --
	(165.22,109.46) --
	(167.37,108.24) --
	(169.35,107.11) --
	(171.84,105.70) --
	(171.98,105.63) --
	(172.31,105.44) --
	(174.28,104.34) --
	(174.56,104.18) --
	(178.99,101.73) --
	(179.47,101.47) --
	(183.27, 99.43) --
	(189.04, 96.42) --
	(189.82, 96.03) --
	(191.30, 95.29) --
	(197.00, 92.56) --
	(200.34, 91.04) --
	(202.17, 90.24) --
	(203.46, 89.70) --
	(203.59, 89.64) --
	(207.04, 88.24) --
	(209.38, 87.34) --
	(210.20, 87.04) --
	(211.67, 86.51) --
	(216.23, 84.99) --
	(217.25, 84.67) --
	(219.14, 84.12) --
	(221.13, 83.57);
\end{scope}
\begin{scope}
\path[clip] (  0.00,  0.00) rectangle (252.94,216.81);
\definecolor{drawColor}{RGB}{0,0,0}

\path[draw=drawColor,line width= 0.4pt,line join=round,line cap=round] ( 49.20, 61.20) --
	(227.75, 61.20) --
	(227.75,167.61) --
	( 49.20,167.61) --
	cycle;
\end{scope}
\begin{scope}
\path[clip] ( 49.20, 61.20) rectangle (227.75,167.61);
\definecolor{drawColor}{RGB}{0,0,0}

\path[draw=drawColor,line width= 0.4pt,dash pattern=on 4pt off 4pt ,line join=round,line cap=round] ( 55.81,142.02) --
	( 58.64,142.04) --
	( 58.91,142.04) --
	( 59.31,142.04) --
	( 59.37,142.04) --
	( 59.47,142.04) --
	( 61.28,142.05) --
	( 62.75,142.06) --
	( 63.00,142.06) --
	( 63.00,142.06) --
	( 64.45,142.07) --
	( 65.78,142.08) --
	( 66.74,142.09) --
	( 70.42,142.11) --
	( 70.81,142.12) --
	( 72.15,142.13) --
	( 73.12,142.14) --
	( 73.27,142.14) --
	( 77.06,142.17) --
	( 78.69,142.18) --
	( 85.33,142.25) --
	( 85.71,142.25) --
	( 87.74,142.28) --
	( 88.01,142.28) --
	( 88.14,142.28) --
	( 88.79,142.29) --
	( 89.33,142.29) --
	( 90.87,142.31) --
	( 93.02,142.34) --
	( 98.26,142.42) --
	( 98.69,142.42) --
	( 98.83,142.43) --
	(100.69,142.46) --
	(101.43,142.47) --
	(105.02,142.53) --
	(108.77,142.61) --
	(108.82,142.61) --
	(110.45,142.64) --
	(111.81,142.67) --
	(112.13,142.68) --
	(116.04,142.78) --
	(116.37,142.78) --
	(118.79,142.85) --
	(119.44,142.87) --
	(120.40,142.89) --
	(120.56,142.90) --
	(121.41,142.92) --
	(123.65,142.99) --
	(126.87,143.10) --
	(130.33,143.22) --
	(131.36,143.26) --
	(134.23,143.37) --
	(135.77,143.44) --
	(138.67,143.57) --
	(140.17,143.64) --
	(140.71,143.67) --
	(142.84,143.77) --
	(143.23,143.79) --
	(143.44,143.80) --
	(144.86,143.88) --
	(149.04,144.11) --
	(149.77,144.16) --
	(150.59,144.21) --
	(153.45,144.39) --
	(157.74,144.68) --
	(159.33,144.80) --
	(159.72,144.83) --
	(160.19,144.86) --
	(163.40,145.11) --
	(164.15,145.18) --
	(164.48,145.20) --
	(164.63,145.21) --
	(165.22,145.26) --
	(165.22,145.26) --
	(167.37,145.45) --
	(169.35,145.62) --
	(171.84,145.86) --
	(171.98,145.87) --
	(172.31,145.90) --
	(174.28,146.09) --
	(174.56,146.12) --
	(178.99,146.58) --
	(179.47,146.63) --
	(183.27,147.05) --
	(189.04,147.75) --
	(189.82,147.84) --
	(191.30,148.03) --
	(197.00,148.78) --
	(200.34,149.25) --
	(202.17,149.50) --
	(203.46,149.69) --
	(203.59,149.71) --
	(207.04,150.21) --
	(209.38,150.55) --
	(210.20,150.67) --
	(211.67,150.89) --
	(216.23,151.57) --
	(217.25,151.72) --
	(219.14,152.00) --
	(221.13,152.28);
\end{scope}
\begin{scope}
\path[clip] (  0.00,  0.00) rectangle (252.94,216.81);
\definecolor{drawColor}{RGB}{0,0,0}

\path[draw=drawColor,line width= 0.4pt,line join=round,line cap=round] ( 49.20, 61.20) --
	(227.75, 61.20) --
	(227.75,167.61) --
	( 49.20,167.61) --
	cycle;
\end{scope}
\begin{scope}
\path[clip] ( 49.20, 61.20) rectangle (227.75,167.61);
\definecolor{drawColor}{RGB}{0,0,0}

\path[draw=drawColor,line width= 0.4pt,dash pattern=on 1pt off 3pt ,line join=round,line cap=round] ( 55.81,104.81) --
	( 58.64,105.01) --
	( 58.91,105.04) --
	( 59.31,105.07) --
	( 59.37,105.08) --
	( 59.47,105.09) --
	( 61.28,105.26) --
	( 62.75,105.42) --
	( 63.00,105.44) --
	( 63.00,105.44) --
	( 64.45,105.62) --
	( 65.78,105.79) --
	( 66.74,105.92) --
	( 70.42,106.50) --
	( 70.81,106.56) --
	( 72.15,106.80) --
	( 73.12,106.98) --
	( 73.27,107.01) --
	( 77.06,107.78) --
	( 78.69,108.14) --
	( 85.33,109.82) --
	( 85.71,109.93) --
	( 87.74,110.51) --
	( 88.01,110.59) --
	( 88.14,110.63) --
	( 88.79,110.82) --
	( 89.33,110.98) --
	( 90.87,111.46) --
	( 93.02,112.15) --
	( 98.26,113.97) --
	( 98.69,114.13) --
	( 98.83,114.18) --
	(100.69,114.88) --
	(101.43,115.16) --
	(105.02,116.58) --
	(108.77,118.16) --
	(108.82,118.18) --
	(110.45,118.89) --
	(111.81,119.50) --
	(112.13,119.64) --
	(116.04,121.46) --
	(116.37,121.61) --
	(118.79,122.78) --
	(119.44,123.10) --
	(120.40,123.58) --
	(120.56,123.66) --
	(121.41,124.09) --
	(123.65,125.23) --
	(126.87,126.92) --
	(130.33,128.79) --
	(131.36,129.36) --
	(134.23,130.97) --
	(135.77,131.85) --
	(138.67,133.53) --
	(140.17,134.42) --
	(140.71,134.74) --
	(142.84,136.02) --
	(143.23,136.25) --
	(143.44,136.38) --
	(144.86,137.24) --
	(149.04,139.82) --
	(149.77,140.27) --
	(150.59,140.79) --
	(153.45,142.59) --
	(157.74,145.34) --
	(159.33,146.37) --
	(159.72,146.62) --
	(160.19,146.92) --
	(163.40,149.01) --
	(164.15,149.50) --
	(164.48,149.72) --
	(164.63,149.82) --
	(165.22,150.20) --
	(165.22,150.21) --
	(167.37,151.61) --
	(169.35,152.91) --
	(171.84,154.56) --
	(171.98,154.65) --
	(172.31,154.86) --
	(174.28,156.16) --
	(174.56,156.34) --
	(178.99,159.25) --
	(179.47,159.56) --
	(183.27,162.03) --
	(189.04,165.73) --
	(189.82,166.22) --
	(191.30,167.14) --
	(197.00,170.63) --
	(200.34,172.61) --
	(202.17,173.67) --
	(203.46,174.40) --
	(203.59,174.47) --
	(207.04,176.37) --
	(209.38,177.61) --
	(210.20,178.04) --
	(211.67,178.79) --
	(216.23,180.99) --
	(217.25,181.45) --
	(219.14,182.29) --
	(221.13,183.12);
\end{scope}
\begin{scope}
\path[clip] (  0.00,  0.00) rectangle (252.94,216.81);
\definecolor{drawColor}{RGB}{0,0,0}

\path[draw=drawColor,line width= 0.4pt,line join=round,line cap=round] ( 55.57, 61.20) -- (221.32, 61.20);

\path[draw=drawColor,line width= 0.4pt,line join=round,line cap=round] ( 55.57, 61.20) -- ( 55.57, 55.20);

\path[draw=drawColor,line width= 0.4pt,line join=round,line cap=round] ( 88.72, 61.20) -- ( 88.72, 55.20);

\path[draw=drawColor,line width= 0.4pt,line join=round,line cap=round] (121.87, 61.20) -- (121.87, 55.20);

\path[draw=drawColor,line width= 0.4pt,line join=round,line cap=round] (155.02, 61.20) -- (155.02, 55.20);

\path[draw=drawColor,line width= 0.4pt,line join=round,line cap=round] (188.17, 61.20) -- (188.17, 55.20);

\path[draw=drawColor,line width= 0.4pt,line join=round,line cap=round] (221.32, 61.20) -- (221.32, 55.20);

\node[text=drawColor,anchor=base,inner sep=0pt, outer sep=0pt, scale=  1.00] at ( 55.57, 39.60) {0.0};

\node[text=drawColor,anchor=base,inner sep=0pt, outer sep=0pt, scale=  1.00] at ( 88.72, 39.60) {0.2};

\node[text=drawColor,anchor=base,inner sep=0pt, outer sep=0pt, scale=  1.00] at (121.87, 39.60) {0.4};

\node[text=drawColor,anchor=base,inner sep=0pt, outer sep=0pt, scale=  1.00] at (155.02, 39.60) {0.6};

\node[text=drawColor,anchor=base,inner sep=0pt, outer sep=0pt, scale=  1.00] at (188.17, 39.60) {0.8};

\node[text=drawColor,anchor=base,inner sep=0pt, outer sep=0pt, scale=  1.00] at (221.32, 39.60) {1.0};

\path[draw=drawColor,line width= 0.4pt,line join=round,line cap=round] ( 49.20, 74.99) -- ( 49.20,153.82);

\path[draw=drawColor,line width= 0.4pt,line join=round,line cap=round] ( 49.20, 74.99) -- ( 43.20, 74.99);

\path[draw=drawColor,line width= 0.4pt,line join=round,line cap=round] ( 49.20, 94.70) -- ( 43.20, 94.70);

\path[draw=drawColor,line width= 0.4pt,line join=round,line cap=round] ( 49.20,114.41) -- ( 43.20,114.41);

\path[draw=drawColor,line width= 0.4pt,line join=round,line cap=round] ( 49.20,134.11) -- ( 43.20,134.11);

\path[draw=drawColor,line width= 0.4pt,line join=round,line cap=round] ( 49.20,153.82) -- ( 43.20,153.82);

\node[text=drawColor,rotate= 90.00,anchor=base,inner sep=0pt, outer sep=0pt, scale=  1.00] at ( 34.80, 74.99) {-0.4};

\node[text=drawColor,rotate= 90.00,anchor=base,inner sep=0pt, outer sep=0pt, scale=  1.00] at ( 34.80,114.41) {0.0};

\node[text=drawColor,rotate= 90.00,anchor=base,inner sep=0pt, outer sep=0pt, scale=  1.00] at ( 34.80,153.82) {0.4};

\path[draw=drawColor,line width= 0.4pt,line join=round,line cap=round] ( 49.20, 61.20) --
	(227.75, 61.20) --
	(227.75,167.61) --
	( 49.20,167.61) --
	cycle;
\end{scope}
\end{tikzpicture}
\vspace{-1.5cm}
\caption{Plots of TE, NDE, and NIE. The solid line is TE, the dashed line is NIE, and the dotted line is NIE. The x-axis represents the value of $C$, and the y-axis represents the values of TE, NDE, and NIE, respectively.}
\label{fig:a1}
\end{figure}

\begin{figure}
\vspace{0cm}
    \centering
\begin{tikzpicture}[x=1pt,y=1pt]
\definecolor{fillColor}{RGB}{255,255,255}
\path[use as bounding box,fill=fillColor,fill opacity=0.00] (0,0) rectangle (252.94,216.81);
\begin{scope}
\path[clip] ( 49.20, 61.20) rectangle (227.75,167.61);
\definecolor{drawColor}{RGB}{0,0,0}

\path[draw=drawColor,line width= 0.4pt,line join=round,line cap=round] ( 55.81,151.62) --
	( 58.64,151.43) --
	( 58.91,151.41) --
	( 59.31,151.38) --
	( 59.37,151.37) --
	( 59.47,151.36) --
	( 61.28,151.20) --
	( 62.75,151.05) --
	( 63.00,151.03) --
	( 63.00,151.03) --
	( 64.45,150.86) --
	( 65.78,150.70) --
	( 66.74,150.57) --
	( 70.42,150.02) --
	( 70.81,149.96) --
	( 72.15,149.73) --
	( 73.12,149.56) --
	( 73.27,149.53) --
	( 77.06,148.79) --
	( 78.69,148.44) --
	( 85.33,146.83) --
	( 85.71,146.73) --
	( 87.74,146.17) --
	( 88.01,146.10) --
	( 88.14,146.06) --
	( 88.79,145.87) --
	( 89.33,145.72) --
	( 90.87,145.26) --
	( 93.02,144.59) --
	( 98.26,142.85) --
	( 98.69,142.70) --
	( 98.83,142.65) --
	(100.69,141.99) --
	(101.43,141.72) --
	(105.02,140.35) --
	(108.77,138.85) --
	(108.82,138.83) --
	(110.45,138.16) --
	(111.81,137.58) --
	(112.13,137.44) --
	(116.04,135.72) --
	(116.37,135.58) --
	(118.79,134.47) --
	(119.44,134.17) --
	(120.40,133.72) --
	(120.56,133.64) --
	(121.41,133.24) --
	(123.65,132.17) --
	(126.87,130.58) --
	(130.33,128.83) --
	(131.36,128.30) --
	(134.23,126.81) --
	(135.77,125.99) --
	(138.67,124.44) --
	(140.17,123.62) --
	(140.71,123.33) --
	(142.84,122.16) --
	(143.23,121.94) --
	(143.44,121.83) --
	(144.86,121.04) --
	(149.04,118.70) --
	(149.77,118.29) --
	(150.59,117.82) --
	(153.45,116.20) --
	(157.74,114.41) --
	(159.33,114.41) --
	(159.72,114.41) --
	(160.19,114.41) --
	(163.40,114.41) --
	(164.15,114.41) --
	(164.48,114.41) --
	(164.63,114.41) --
	(165.22,114.41) --
	(165.22,114.41) --
	(167.37,114.41) --
	(169.35,114.41) --
	(171.84,114.41) --
	(171.98,114.41) --
	(172.31,114.41) --
	(174.28,114.41) --
	(174.56,114.41) --
	(178.99,114.41) --
	(179.47,114.41) --
	(183.27,114.41) --
	(189.04,114.41) --
	(189.82,114.41) --
	(191.30,114.41) --
	(197.00,114.41) --
	(200.34,114.41) --
	(202.17,114.41) --
	(203.46,114.41) --
	(203.59,114.41) --
	(207.04,114.41) --
	(209.38,114.41) --
	(210.20,114.41) --
	(211.67,114.41) --
	(216.23,114.41) --
	(217.25,114.41) --
	(219.14,114.41) --
	(221.13,114.41);
\end{scope}
\begin{scope}
\path[clip] (  0.00,  0.00) rectangle (252.94,216.81);
\definecolor{drawColor}{RGB}{0,0,0}

\path[draw=drawColor,line width= 0.4pt,line join=round,line cap=round] ( 49.20, 61.20) --
	(227.75, 61.20) --
	(227.75,167.61) --
	( 49.20,167.61) --
	cycle;
\end{scope}
\begin{scope}
\path[clip] ( 49.20, 61.20) rectangle (227.75,167.61);
\definecolor{drawColor}{RGB}{0,0,0}

\path[draw=drawColor,line width= 0.4pt,dash pattern=on 4pt off 4pt ,line join=round,line cap=round] ( 55.81,142.02) --
	( 58.64,142.04) --
	( 58.91,142.04) --
	( 59.31,142.04) --
	( 59.37,142.04) --
	( 59.47,142.04) --
	( 61.28,142.05) --
	( 62.75,142.06) --
	( 63.00,142.06) --
	( 63.00,142.06) --
	( 64.45,142.07) --
	( 65.78,142.08) --
	( 66.74,142.09) --
	( 70.42,142.11) --
	( 70.81,142.12) --
	( 72.15,142.13) --
	( 73.12,142.14) --
	( 73.27,142.14) --
	( 77.06,142.17) --
	( 78.69,142.18) --
	( 85.33,142.25) --
	( 85.71,142.25) --
	( 87.74,142.28) --
	( 88.01,142.28) --
	( 88.14,142.28) --
	( 88.79,142.29) --
	( 89.33,142.29) --
	( 90.87,142.31) --
	( 93.02,142.34) --
	( 98.26,142.42) --
	( 98.69,142.42) --
	( 98.83,142.43) --
	(100.69,141.99) --
	(101.43,141.72) --
	(105.02,140.35) --
	(108.77,138.85) --
	(108.82,138.83) --
	(110.45,138.16) --
	(111.81,137.58) --
	(112.13,137.44) --
	(116.04,135.72) --
	(116.37,135.58) --
	(118.79,134.47) --
	(119.44,134.17) --
	(120.40,133.72) --
	(120.56,133.64) --
	(121.41,133.24) --
	(123.65,132.17) --
	(126.87,130.58) --
	(130.33,128.83) --
	(131.36,128.30) --
	(134.23,126.81) --
	(135.77,125.99) --
	(138.67,124.44) --
	(140.17,123.62) --
	(140.71,123.33) --
	(142.84,122.16) --
	(143.23,121.94) --
	(143.44,121.83) --
	(144.86,121.04) --
	(149.04,118.70) --
	(149.77,118.29) --
	(150.59,117.82) --
	(153.45,116.20) --
	(157.74,114.41) --
	(159.33,114.41) --
	(159.72,114.41) --
	(160.19,114.41) --
	(163.40,114.41) --
	(164.15,114.41) --
	(164.48,114.41) --
	(164.63,114.41) --
	(165.22,114.41) --
	(165.22,114.41) --
	(167.37,114.41) --
	(169.35,114.41) --
	(171.84,114.41) --
	(171.98,114.41) --
	(172.31,114.41) --
	(174.28,114.41) --
	(174.56,114.41) --
	(178.99,114.41) --
	(179.47,114.41) --
	(183.27,114.41) --
	(189.04,114.41) --
	(189.82,114.41) --
	(191.30,114.41) --
	(197.00,114.41) --
	(200.34,114.41) --
	(202.17,114.41) --
	(203.46,114.41) --
	(203.59,114.41) --
	(207.04,114.41) --
	(209.38,114.41) --
	(210.20,114.41) --
	(211.67,114.41) --
	(216.23,114.41) --
	(217.25,114.41) --
	(219.14,114.41) --
	(221.13,114.41);
\end{scope}
\begin{scope}
\path[clip] (  0.00,  0.00) rectangle (252.94,216.81);
\definecolor{drawColor}{RGB}{0,0,0}

\path[draw=drawColor,line width= 0.4pt,line join=round,line cap=round] ( 49.20, 61.20) --
	(227.75, 61.20) --
	(227.75,167.61) --
	( 49.20,167.61) --
	cycle;
\end{scope}
\begin{scope}
\path[clip] ( 49.20, 61.20) rectangle (227.75,167.61);
\definecolor{drawColor}{RGB}{0,0,0}

\path[draw=drawColor,line width= 0.4pt,dash pattern=on 1pt off 3pt ,line join=round,line cap=round] ( 55.81,124.00) --
	( 58.64,123.80) --
	( 58.91,123.77) --
	( 59.31,123.74) --
	( 59.37,123.73) --
	( 59.47,123.72) --
	( 61.28,123.55) --
	( 62.75,123.39) --
	( 63.00,123.37) --
	( 63.00,123.37) --
	( 64.45,123.19) --
	( 65.78,123.02) --
	( 66.74,122.89) --
	( 70.42,122.31) --
	( 70.81,122.25) --
	( 72.15,122.01) --
	( 73.12,121.83) --
	( 73.27,121.80) --
	( 77.06,121.03) --
	( 78.69,120.67) --
	( 85.33,118.99) --
	( 85.71,118.88) --
	( 87.74,118.30) --
	( 88.01,118.22) --
	( 88.14,118.18) --
	( 88.79,117.99) --
	( 89.33,117.83) --
	( 90.87,117.35) --
	( 93.02,116.66) --
	( 98.26,114.84) --
	( 98.69,114.68) --
	( 98.83,114.63) --
	(100.69,114.41) --
	(101.43,114.41) --
	(105.02,114.41) --
	(108.77,114.41) --
	(108.82,114.41) --
	(110.45,114.41) --
	(111.81,114.41) --
	(112.13,114.41) --
	(116.04,114.41) --
	(116.37,114.41) --
	(118.79,114.41) --
	(119.44,114.41) --
	(120.40,114.41) --
	(120.56,114.41) --
	(121.41,114.41) --
	(123.65,114.41) --
	(126.87,114.41) --
	(130.33,114.41) --
	(131.36,114.41) --
	(134.23,114.41) --
	(135.77,114.41) --
	(138.67,114.41) --
	(140.17,114.41) --
	(140.71,114.41) --
	(142.84,114.41) --
	(143.23,114.41) --
	(143.44,114.41) --
	(144.86,114.41) --
	(149.04,114.41) --
	(149.77,114.41) --
	(150.59,114.41) --
	(153.45,114.41) --
	(157.74,114.41) --
	(159.33,114.41) --
	(159.72,114.41) --
	(160.19,114.41) --
	(163.40,114.41) --
	(164.15,114.41) --
	(164.48,114.41) --
	(164.63,114.41) --
	(165.22,114.41) --
	(165.22,114.41) --
	(167.37,114.41) --
	(169.35,114.41) --
	(171.84,114.41) --
	(171.98,114.41) --
	(172.31,114.41) --
	(174.28,114.41) --
	(174.56,114.41) --
	(178.99,114.41) --
	(179.47,114.41) --
	(183.27,114.41) --
	(189.04,114.41) --
	(189.82,114.41) --
	(191.30,114.41) --
	(197.00,114.41) --
	(200.34,114.41) --
	(202.17,114.41) --
	(203.46,114.41) --
	(203.59,114.41) --
	(207.04,114.41) --
	(209.38,114.41) --
	(210.20,114.41) --
	(211.67,114.41) --
	(216.23,114.41) --
	(217.25,114.41) --
	(219.14,114.41) --
	(221.13,114.41);
\end{scope}
\begin{scope}
\path[clip] (  0.00,  0.00) rectangle (252.94,216.81);
\definecolor{drawColor}{RGB}{0,0,0}

\path[draw=drawColor,line width= 0.4pt,line join=round,line cap=round] ( 55.57, 61.20) -- (221.32, 61.20);

\path[draw=drawColor,line width= 0.4pt,line join=round,line cap=round] ( 55.57, 61.20) -- ( 55.57, 55.20);

\path[draw=drawColor,line width= 0.4pt,line join=round,line cap=round] ( 88.72, 61.20) -- ( 88.72, 55.20);

\path[draw=drawColor,line width= 0.4pt,line join=round,line cap=round] (121.87, 61.20) -- (121.87, 55.20);

\path[draw=drawColor,line width= 0.4pt,line join=round,line cap=round] (155.02, 61.20) -- (155.02, 55.20);

\path[draw=drawColor,line width= 0.4pt,line join=round,line cap=round] (188.17, 61.20) -- (188.17, 55.20);

\path[draw=drawColor,line width= 0.4pt,line join=round,line cap=round] (221.32, 61.20) -- (221.32, 55.20);

\node[text=drawColor,anchor=base,inner sep=0pt, outer sep=0pt, scale=  1.00] at ( 55.57, 39.60) {0.0};

\node[text=drawColor,anchor=base,inner sep=0pt, outer sep=0pt, scale=  1.00] at ( 88.72, 39.60) {0.2};

\node[text=drawColor,anchor=base,inner sep=0pt, outer sep=0pt, scale=  1.00] at (121.87, 39.60) {0.4};

\node[text=drawColor,anchor=base,inner sep=0pt, outer sep=0pt, scale=  1.00] at (155.02, 39.60) {0.6};

\node[text=drawColor,anchor=base,inner sep=0pt, outer sep=0pt, scale=  1.00] at (188.17, 39.60) {0.8};

\node[text=drawColor,anchor=base,inner sep=0pt, outer sep=0pt, scale=  1.00] at (221.32, 39.60) {1.0};

\path[draw=drawColor,line width= 0.4pt,line join=round,line cap=round] ( 49.20, 74.99) -- ( 49.20,153.82);

\path[draw=drawColor,line width= 0.4pt,line join=round,line cap=round] ( 49.20, 74.99) -- ( 43.20, 74.99);

\path[draw=drawColor,line width= 0.4pt,line join=round,line cap=round] ( 49.20, 94.70) -- ( 43.20, 94.70);

\path[draw=drawColor,line width= 0.4pt,line join=round,line cap=round] ( 49.20,114.41) -- ( 43.20,114.41);

\path[draw=drawColor,line width= 0.4pt,line join=round,line cap=round] ( 49.20,134.11) -- ( 43.20,134.11);

\path[draw=drawColor,line width= 0.4pt,line join=round,line cap=round] ( 49.20,153.82) -- ( 43.20,153.82);

\node[text=drawColor,rotate= 90.00,anchor=base,inner sep=0pt, outer sep=0pt, scale=  1.00] at ( 34.80, 74.99) {-0.4};

\node[text=drawColor,rotate= 90.00,anchor=base,inner sep=0pt, outer sep=0pt, scale=  1.00] at ( 34.80,114.41) {0.0};

\node[text=drawColor,rotate= 90.00,anchor=base,inner sep=0pt, outer sep=0pt, scale=  1.00] at ( 34.80,153.82) {0.4};

\path[draw=drawColor,line width= 0.4pt,line join=round,line cap=round] ( 49.20, 61.20) --
	(227.75, 61.20) --
	(227.75,167.61) --
	( 49.20,167.61) --
	cycle;
\end{scope}
\end{tikzpicture}
\vspace{-1.5cm}
\caption{Plots of T-PNS, ND-PNS, and NI-PNS. The solid line is T-PNS, the dashed line is NI-PNS, and the dotted line is NI-PNS. The x-axis represents the value of $C$, and the y-axis represents the values of T-PNS, ND-PNS, and NI-PNS, respectively.}
\label{fig:a2}
\end{figure}

\end{document}